%% file: digitcode_arxiv.tex
\documentclass[letterpaper]{article} % DO NOT CHANGE THIS
\usepackage[preprint]{aaai2027}  % arXiv preprint mode: shows authors, no anonymized slug
\usepackage[hyphens]{url}  % DO NOT CHANGE THIS
\usepackage{graphicx} % DO NOT CHANGE THIS
\usepackage{natbib}  % DO NOT CHANGE THIS AND DO NOT ADD ANY OPTIONS TO IT
\usepackage{caption} % DO NOT CHANGE THIS AND DO NOT ADD ANY OPTIONS TO IT
\usepackage{amsmath}
\usepackage{amssymb}
\usepackage{array}
\usepackage{booktabs}
\usepackage{multirow}
\usepackage{tikz}
\usetikzlibrary{arrows.meta,positioning,fit,backgrounds,shapes.geometric,decorations.pathreplacing,calc}
\input{Figures/fig_ic_cube.tex}
\input{Figures/fig_ic_sphere.tex}
\input{Figures/fig_ic_finger.tex}

\title{DigitCode: Symbolic Tokenization of Hand Motion by Anatomical Units}

\author{
    Haoyu Gu\textsuperscript{\rm 1}\equalcontrib,
    Haotian Lu\textsuperscript{\rm 2}\equalcontrib,
    Jingrun Du\textsuperscript{\rm 2},
    Xiao-Ping Zhang\textsuperscript{\rm 2}\corresponding
}
\affiliations{
    \textsuperscript{\rm 1}School of Future Technology, South China University of Technology\\
    \textsuperscript{\rm 2}Shenzhen Key Laboratory of Ubiquitous Data Enabling,\\
    Tsinghua Shenzhen International Graduate School, Tsinghua University\\
    ghy20050104@gmail.com, haotianlu666@gmail.com, xpzhang@ieee.org
}

\begin{document}

\maketitle

\begin{abstract}
Hand motion carries the finest-grained information in human activity, yet the representations behind hand generation, understanding, and robot learning are overwhelmingly continuous---joint angles or MANO parameters. These are accurate but unstructured: a finger cannot be indexed or edited as a symbol, and nothing marks a pose as anatomically valid. Discrete symbolic representations supply exactly this structure, and Hand Labanotation (HL) has shown they are feasible for the hand, writing motion as a $T\times 40$ grid of one fixed direction symbol per bone. Building on this grid, we ask the question underneath it: the anatomical \emph{unit} a symbol should span---bone, finger, or whole hand. DigitCode answers it by adapting, grouping, and layering HL's alphabet along the hand's unit hierarchy within one code, cutting the symbolic representation's quantization error by three quarters. The lever is the unit, not the quantizer family: at a fixed unit, training-free and learned \emph{strong} quantizers are interchangeable on reconstruction, while moving down the anatomical hierarchy is what shifts accuracy. The hierarchy also tracks what downstream tasks need. Because a finger is a genuine, enumerable unit, one per-finger token doubles as a training-free, editable handle for jobs a continuous representation cannot address---repairing malformed generated hands, and retargeting them onto robots. We release \textsc{HandTok}, a reproducible testbed, so hand tokenizers can be compared unit-for-unit. Project page: \url{https://digitcode-demo.github.io}.
\end{abstract}

\section{Introduction}

When a generative model draws a hand with a thumb bent backward through the palm, the usual fix redraws the entire hand, because nothing in its representation isolates \emph{the thumb} as a part one could correct in place. That failure traces back to how the hand was tokenized in the first place: every discrete representation decides two things---how finely to quantize, and what a token should \emph{span}. Tokenization research has overwhelmingly pursued the first---larger codebooks, learned and residual quantizers---while the second is usually inherited from the data's grid and left unexamined. For signals with natural compositional structure, we argue, the second decision is the one that matters. Hand motion is an unusually clean place to test this. Its parts are not a matter of taste---everyone agrees a hand has fingers and fingers have bones---so the candidate units come from anatomy rather than from the modeller. How those parts move together is known in advance, from tendons and joint limits, so \emph{which} grouping should pay is a physical prediction rather than a hyperparameter search. And a symbolic grid for the hand already exists, so the unit can be varied while the alphabet, frame rate, and extraction pipeline are held still. Few signals offer all three, so an answer here reads as evidence about the design decision rather than about one dataset.

Hand generation, understanding, and robot learning all turn on representing hand motion faithfully \citep{qin2022dexmv,shaw2023videodex}, yet the representations underneath are overwhelmingly continuous---joint angles, or the MANO parametric model \citep{romero2017mano} recovered by pose estimators \citep{moon2020interhand,pavlakos2024hamer}---accurate but unstructured for anything symbolic: a gesture cannot be indexed, one finger cannot be edited without re-optimizing parameters that spill across the hand, and nothing marks a pose as anatomically legal. A discrete symbolic representation supplies this structure by construction, and Hand Labanotation (HL) \citep{li2024hl} has shown it is feasible for the hand: a $T\times 40$ grid with one symbol per bone from a fixed $26$-direction alphabet (HL-26), deterministic and reproducible from public 3D joints. HL thus brings the hand into the symbolic world---but at a resolution the notation dictated, one symbol per bone, not one the hand's own anatomy suggests. Its flat bone grid ignores that the bones within a finger move together while fingers move apart (\S\ref{sec:method}); that unit underneath the symbol, not the symbol itself, is what we make the design question. HL's coarseness is not simply a deficit: we find it is the right unit for identity-level tasks under noise (\S\ref{sec:cross}). The question is which unit for which task, not that HL chose wrong.

Two findings organize the paper. First, the unit, not the quantizer, is where the leverage lies: at a fixed unit, training-free and learned \emph{strong} quantizers are interchangeable, whereas changing the unit reshapes the accuracy--rate trade-off. The reason is structural: a finger's four bones move as a coupled chain, so quantizing them independently spends rate on redundancy no quantizer can recover, while one finger token captures the coupling for free. Second, the hand offers a hierarchy---bone, finger, hand---that lines up with what tasks need.

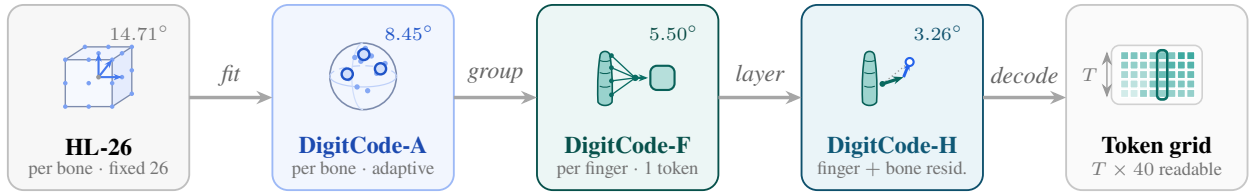
\begin{figure*}[!t]
\centering
\input{Figures/fig_overview.tex}
\caption{Overview of DigitCode. Fitting, grouping, and layering HL's alphabet change the anatomical \emph{unit} a token spans (DigitCode-A/F/H), taking held-out error from $14.71^\circ$ to $3.26^\circ$ at essentially HL-26's rate ($4.75$ vs.\ $4.70$ bits); colour marks the unit.}
\label{fig:overview}
\end{figure*}

DigitCode turns these findings into a code by three changes to HL, each driven by a property of hand motion. Bone directions concentrate anisotropically, so we \emph{fit} the alphabet to the data. That coupling says where to cut, so we \emph{group} a finger's bones and quantize them jointly. One finger token caps precision, so we \emph{layer} a per-bone residual beneath it, keeping both units at once; relative and event coding then compress the near-constant time axis. Every token stays a readable direction, and held-out error falls from $14.71^\circ$ to $3.26^\circ$ (Fig.~\ref{fig:overview}).

This paper makes three contributions.
\begin{enumerate}
\item \textbf{DigitCode}, a symbolic code whose tokens are anatomical units. By fitting, grouping, and layering HL's alphabet along the bone--finger--hand hierarchy, a single code spans three units and cuts held-out quantization error by three quarters, with every token still a readable direction (\S\ref{sec:method}).
\item \textbf{A controlled separation of the two decisions every discrete hand representation makes}---what unit a token spans, and how to quantize it. On a common harness that sweeps both axes at matched rate, moving the unit down the anatomy dominates any change of quantizer family: training-free and learned \emph{strong} quantizers are interchangeable at a fixed unit, while switching the unit reshapes the rate--distortion frontier (\S\ref{sec:rd}).
\item \textbf{A genuine anatomical unit that doubles as a task interface.} Because a finger is enumerable, one per-finger token is a training-free handle for detecting, editing, and repairing corrupted fingers and for compiling retargeting onto robot hands---operations no continuous parameterization supports. The useful unit tracks the task across six downstream settings: dynamics read the bone, interaction the finger, identity the whole hand. We release the evaluation harness as \textsc{HandTok} (\S\ref{sec:exp}--\ref{sec:embodied}).
\end{enumerate}

\section{Related Work}

\textbf{Continuous representations.} Continuous parameterizations are accurate but carry no symbolic structure, and this paper supplies what they lack. Hand motion is dominantly represented by joint angles or the MANO parametric model \citep{romero2017mano}, recovered from images by pose estimators \citep{moon2020interhand,pavlakos2024hamer} and consumed by generation and robot-learning pipelines alike---with no indexing, no part-level handle, no legal set. The gap shows at the two ends this paper reaches: in generated imagery, malformed hands are the most common artifact, and continuous-domain repair (HandRefiner \citep{lu2024handrefiner}, RealisHuman \citep{wang2025realishuman}) must re-paint the \emph{whole} hand, detector and repairer separate and neither part-addressable; in robotics, video-to-dexterity pipelines \citep{qin2022dexmv,shaw2023videodex} retarget estimated poses \citep{qin2023anyteleop} with no legality check or per-finger handle. \S\ref{sec:interface} and \S\ref{sec:embodied} show a symbolic per-finger unit supplies both.

\textbf{Symbolic notation.} Symbolic notation provides the structure continuous poses lack but fixes the unit at body-part resolution; we study how to move it. Labanotation writes movement as direction symbols on a grid of body parts, HamNoSys \citep{prillwitz1989hamnosys} does the same for sign language, trading precision for symbols that are readable, editable, and archivable. A computational lineage followed---automated transcription \citep{genlaban2014,laban_transformer_lstm}, \emph{bidirectional} conversion (DASKEL \citep{daskel}, the premise of our rate--distortion protocol), and modeling in symbol space \citep{hkbu_lbnmae,jiang2026lamogen}---with HL \citep{li2024hl} marking its arrival at the finger level. A parallel instinct appears in learned tokenizers that discretize \emph{along} physical structure: M3T \citep{m3t2026} gives body, hand, and face separate codebooks, and hand--object work \citep{huang2025hoigpt,luo2025beingh0,cha2024text2hoi,liu2024contactgen} factorizes along the interacting parts. All fix the symbol at body-part resolution; building on HL's grid, we ask how to \emph{adapt}, \emph{factorize}, and \emph{layer} the alphabet at finer anatomical units (\S\ref{sec:method}), keeping its interpretability.

\textbf{Discrete tokenization.} Learned tokenizers pour their design effort into the quantizer; we find the anatomical unit the greater lever. VQ-VAE \citep{vandenoord2017vqvae} launched a wave of discrete motion tokenizers \citep{zhang2023t2mgpt,guo2022tm2t,jiang2023motiongpt} with residual and masked variants \citep{guo2024momask,pinyoanuntapong2024mmm,yuan2024mogents,scalemogen2026,anymo2026}. Some already tokenize at a spatial unit finer than the whole body---MoGenTS \citep{yuan2024mogents} quantizes per joint---but there the unit is fixed once by the architecture; none varies it against the quantizer at matched rate, which is the comparison we run. Symbolic music makes the same move on the time axis \citep{qian2026beat}. Sign-language production followed into Pose-VQVAE token spaces \citep{saunders2020progressive,xie2024g2pddm,signsastokens2024}; CoMo \citep{huang2024como} makes such codes editable, though it cannot detect or localize what needs editing. Underneath sit the quantizers themselves---RVQ \citep{zeghidour2021soundstream}, FSQ \citep{mentzer2024fsq}, PQ \citep{jegou2011pq}, BSQ \citep{zhao2024bsq}---while our relative and event coding follow analogues in symbolic music \citep{hsiao2021cp}. For the hand this design space---quantizer family against anatomical unit, learned against training-free---has not been mapped; we characterize it on a common footing (\S\ref{sec:exp}).

\section{Method}
\label{sec:method}

DigitCode is a family of symbolic codes on HL's grid, built by changing HL-26's design decisions one at a time: we recall HL-26, measure where a better code should differ, then make the changes in turn.

For each hand, HL extracts $20$ bone-direction vectors from 3D joints and normalizes them to the unit sphere $S^2$. HL-26 quantizes a direction $\mathbf{v}\in S^2$ to
\(
q(\mathbf{v}) = \arg\max_{\mathbf{c}\in\mathcal{C}_{26}} \mathbf{v}^\top \mathbf{c},
\)
where $\mathcal{C}_{26}$ normalizes $\{-1,0,1\}^3\setminus\{0\}$ (the cube's $6$ faces, $12$ edges, $8$ corners). Stacking two hands over $T$ frames yields the $T\times 40$ symbolic grid we build on. The rule is deterministic and reproducible from any public source of 3D joints, so HL-26 needs none of HL's internal data.

Three empirical observations on InterHand2.6M ($36$ captures, $717$K vectors) motivate the design, each pointing to one departure from HL-26's fixed per-bone grid. \emph{(1) Anisotropy}: $68.1\%$ of all bone directions fall on the single $+Y$ face of the cube, so HL-26's uniform alphabet wastes most of its symbols and should instead be \emph{fit to the data} (App.~\ref{app:geom}). \emph{(2) Intra-finger coupling}: mutual information averages $39\%$ within a finger vs.\ $29\%$ across (cross-finger peak $51\%$, middle--ring), so the \emph{finger}, not the bone, is the natural token boundary---strong dependence inside to exploit, a weak seam to cut. \emph{(3) Temporal redundancy}: $79\%$ of tokens are unchanged from the previous frame, so the \emph{time axis} is compressible. None of these regularities is available to a fixed per-bone code on a flat grid, the design HL inherited from Labanotation's per-body-part columns; HL-26's own held-out error is $14.71^\circ$ (per-observation statistics in App.~\ref{app:repro}).

DigitCode makes these three changes in turn (Fig.~\ref{fig:overview}): \textbf{DigitCode-A} fits the codebook to the data. \textbf{DigitCode-F} moves the unit from the bone to the finger. \textbf{DigitCode-H} layers both into a coarse-plus-residual hierarchy. Relative and event coding then compress the time axis, and an anatomy-grouped schedule decodes it. Every step stays on HL's grid and keeps the codes geometric and readable. Throughout, we call a token span a \emph{genuine unit} when its useful properties follow from the physical structure it names rather than from a training objective---a finger is one, a learned latent over the whole hand is not, and \S\ref{sec:interface} makes the difference precise by turning it into operations. Bone, finger, and hand are all genuine units; which one a task should read is the separate question \S\ref{sec:downstream} answers.

\subsection{DigitCode-A: Data-Adaptive Codebook on $S^2$}
\label{sec:adaptive}
The smallest departure keeps the bone unit and replaces only the fixed alphabet (change 1). DigitCode-A runs spherical k-means \citep{lloyd1982kmeans} on training directions and quantizes by argmax-cosine exactly as HL-26; the direction space is low-dimensional, so no encoder training is needed (\S\ref{sec:discussion}). At the same codebook size ($K{=}26$, both costing $4.70$ fixed-width bits/bone), the adaptive code attains $8.45^\circ$, a $43\%$ reduction over HL-26's $14.71^\circ$ (low-rate operating points in App.~\ref{app:quant}).
\subsection{DigitCode-F: Per-Finger Joint Quantization}
\label{sec:perfinger}
The next change is the unit itself (change 2): from the bone to the finger. Rather than quantize each of a finger's four bones independently, DigitCode-F concatenates them into a single $12$-D vector and quantizes the finger \emph{jointly}; fingers stay separate, cutting along the weak cross-finger seam. The seam being \emph{anatomical} is what pays, not the larger block: at fixed $12$-D block, $K$, and rate, regrouping the same bones at random costs $2.4^\circ$ ($9.48$ vs.\ $7.10^\circ$, $K{=}64$), a half-anatomical split landing between the two---the classical block effect \citep{jegou2011pq} is secondary to \emph{which} bones share a code (App.~\ref{app:grouping}). One token now denotes an entire finger pose:
\begin{equation}
\mathbf{f}=[\mathbf{v}_1;\mathbf{v}_2;\mathbf{v}_3;\mathbf{v}_4]\in\mathbb{R}^{12},
\quad
q_F(\mathbf{f})=\arg\min_{k}\|\mathbf{f}-\mathbf{c}_k\|_2,
\end{equation}
decoded by splitting $\mathbf{c}_{q_F(\mathbf{f})}$ into four bone directions and renormalizing each. DigitCode-F reaches $5.50^\circ$ at $2.0$ bits/bone against $5.69^\circ$ at $6.0$ bits for independent per-bone k-means---the same fidelity at a third of the rate. The decoded grid stays $T\times 40$; the token stream is now $T\times 10$.
\subsection{DigitCode-H: Hierarchical Coarse + Residual}
\label{sec:hier}
One finger token must summarize four bones---an accuracy ceiling. The last change (change 3) keeps both units at once, layered: the finger code becomes a \emph{coarse} token capturing ``which finger pose,'' and a per-bone code on its \emph{residual} refines ``how much.'' The finger is the interface unit, the bone the precision unit, and one code carries both. Writing $\hat{\mathbf{v}}^{c}_b$ for the coarse reconstruction of bone $b$ and $R_{\mathbf{u}\to\mathbf{e}_y}$ for the minimal rotation taking a unit vector $\mathbf{u}$ to the canonical pole, the residual is itself a direction,
\begin{equation}
\mathbf{r}_b=R_{\hat{\mathbf{v}}^{c}_b\to\mathbf{e}_y}\,\mathbf{v}_b,
\qquad
\hat{\mathbf{v}}_b=R^{-1}_{\hat{\mathbf{v}}^{c}_b\to\mathbf{e}_y}\,\hat{\mathbf{r}}_b,
\end{equation}
quantized per bone with a single shared spherical $K_2$-entry codebook. The two layers carry near-independent information (coarse$\to$residual MI is only $10\%$), so the residual is cheap. DigitCode-H halves the error of a flat per-bone code at lower rate, continuing to $1.86^\circ$ at $6.75$ bits (Table~\ref{tab:progression}; the flat-code comparison is in App.~\ref{app:hier}). Residual quantization is standard \citep{zeghidour2021soundstream}; here the coarse layer is an anatomically grounded finger pose, so the split has physical meaning.

\subsection{Temporal and Decoding Axes}
\label{sec:relative}
Two further axes are orthogonal to the unit question and we treat them briefly. Coding each direction relative to its temporal or kinematic reference shifts the whole R--D curve left (App.~\ref{app:rel}); emitting tokens only at curvature-aware \emph{keyframes} \citep{douglas1973rdp} and interpolating the rest by slerp \citep{shoemake1985slerp} halves the sequence at matched fidelity, once a fair control credits the apparent speed gain to the decoder rather than the tokens (App.~\ref{app:event}).\label{sec:event} A frame's per-bone tokens can also be emitted in parallel, flat, or in an anatomy-grouped \emph{delay} order \citep{copet2023musicgen,wang2025timeshifted}; scored by leak-free generation---teacher-forced likelihood is not comparable, since a cross-frame delay leaks future frames---the delay wins on long continuous motion and reverses on short static clips, so we adopt it only there (App.~\ref{app:schedule}). None of these changes the unit, and the headline codes below use none of them.\label{sec:schedule}

\section{Experiments}
\label{sec:exp}

This section evaluates the code itself---reconstruction, downstream tasks, generalization; \S\ref{sec:interface}--\ref{sec:embodied} then read the genuine unit's four properties as an interface and a robot pipeline.

\subsection{Setup}
Four public capture regimes span the axes of variation: two-hand interaction (InterHand2.6M \citep{moon2020interhand}), a static single hand (FreiHAND \citep{zimmermann2019freihand}), free single-hand sequences (HanCo \citep{zimmermann2021hanco}), and noisy lifted sign language (ASL-Skeleton3D \citep{amorim2022aslskeleton3d}), all from public 3D joints. Splits are leakage-controlled (capture-disjoint; subject-disjoint for identity tasks, App.~\ref{app:repro}). Distortion is the mean angular error between reconstructed and ground-truth directions; rate is reported under two labeled conventions, fixed-width $\log_2 K$ for the headline tables and empirical entropy for the bake-offs. The quantizer harness fits a codebook at any unit and sweeps k-means/VQ/RVQ/FSQ/PQ/BSQ against HL-26 ($14.71^\circ$); we release it as \textsc{HandTok} so other tokenizers can be compared on the same footing.

Adaptive codebooks are fit on the training partition only; autoregressive tasks use a small LLaMA-style transformer \citep{touvron2023llama}. Headline comparisons carry sequence-level bootstrap CIs and paired permutation tests; generation and schedule models are trained at $3$ seeds and scored by a collapse-sensitive Fr\'echet distance in a motion-autoencoder latent (AE-FGD) and by the Jensen--Shannon divergence of speed distributions (App.~\ref{app:repro}). Learned baselines are tuned (EMA, dead-code revival); the quantizers we sweep are the cores of published motion tokenizers---VQ (T2M-GPT), RVQ (MoMask), FSQ---retrained rate-matched, since body-motion checkpoints do not transfer to hand directions. A full MoMask-style RVQ-VAE \emph{system} is aligned in Table~\ref{tab:progression}.

\subsection{Rate--Distortion: the Unit Sets the Frontier}
\label{sec:rd}
\begin{figure}[t]
\centering
\includegraphics[width=\linewidth]{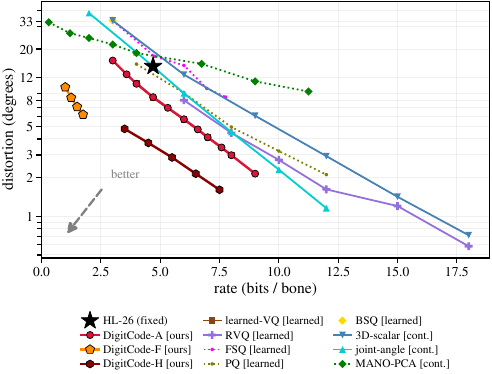}
\caption{Rate--distortion design space (held-out InterHand2.6M; log distortion, fixed-width rate); each family is a sweep over codebook size $K$.}
\label{fig:rd}
\end{figure}

Table~\ref{tab:progression} and Fig.~\ref{fig:rd} read as one move down the anatomy: fitting the bone unit to the data (DigitCode-A) reaches $8.45^\circ$, moving the unit to the finger (DigitCode-F) reaches $5.50^\circ$ at \emph{one-third} the rate of a matched-fidelity per-bone code (\S\ref{sec:perfinger}), and layering the two (DigitCode-H) reaches $3.26^\circ$ at essentially HL-26's rate ($4.75$ vs.\ $4.70$ bits), continuing to $1.86^\circ$ at $6.75$. Switching the unit lowers error and rate together, which no quantizer change at a fixed unit achieves, and it holds under either rate convention (entropy: $3.29$ bits for HL's $14.71^\circ$ vs.\ $1.94$ for F's $5.50^\circ$).
\begin{table}[t]
\centering
\small
\setlength{\tabcolsep}{3pt}
\begin{tabular}{lccc}
\toprule
Representation & fixed b. & entr.\ b. & Ang.\ ($^\circ$) \\
\midrule
HL-26 (fixed, $K{=}26$) & $4.70$ & $3.29$ & $14.71$ \\
\;+ DigitCode-A ($K{=}26$) & $4.70$ & $4.38$ & $8.45$ \\
\;+ DigitCode-F ($K{=}256$) & $\mathbf{2.0}$ & $1.94$ & $\mathbf{5.50}$ \\
\;+ DigitCode-H ($K_1{=}128$, $K_2{=}8$)  & $4.75$ & $4.37$ & $\mathbf{3.26}$ \\
\;+ DigitCode-H ($K_1{=}128$, $K_2{=}16$) & $5.75$ & $5.20$ & $2.55$ \\
\;+ DigitCode-H ($K_1{=}128$, $K_2{=}32$) & $6.75$ & $6.12$ & $1.86$ \\
\midrule
learned VQ ($K{=}26$)    & $4.70$ & $4.47$ & $8.36$ \\
learned VQ ($K{=}128$)   & $7.0$  & $6.70$ & $4.11$ \\
MoMask-style RVQ-VAE ($V{=}4$)  & $1.80$ & $1.67$ & $5.83$ \\
MoMask-style RVQ-VAE ($V{=}14$) & $6.30$ & $5.99$ & $3.90$ \\
joint-angle (continuous) & $6.0$  & $5.21$ & $9.06$ \\
\bottomrule
\end{tabular}
\caption{Three-step progression, held-out InterHand2.6M; below the rule: learned VQ, a MoMask-style sequence RVQ-VAE \emph{system} (App.~\ref{app:quant}), continuous joint-angle. Rates as in \S\ref{sec:rd}. At HL-26's rate DigitCode-A halves its error and DigitCode-H ($K_2{=}8$) quarters it; DigitCode-F (bold) beats it on \emph{both} axes at $43\%$ of the rate.}
\label{tab:progression}
\end{table}

The rest of the design space behaves the same way. At matched bits (Table~\ref{tab:quant}), training-free k-means ties a tuned learned VQ---equivalent to within a declared $0.5^\circ$ margin (\S\ref{sec:discussion})---while FSQ, PQ, and BSQ trail by up to $3.8^\circ$, so the null we claim is over \emph{strong} quantizers only. A uniform spherical grid is even \emph{worse} than HL-26 (App.~\ref{app:neg}). Residual VQ is rate-dependent: it trails k-means in every column, \emph{most} where rate is scarcest ($1.6^\circ$ at ${\sim}4.4$ bits, closing to $0.6$ and $0.4^\circ$), so depth repays only far above our operating points, and there at the finger unit, not the bone (App.~\ref{app:hier}). The remaining gaps are structural: FSQ's and BSQ's grids leave most cells empty under the $+Y$ concentration, saturating their entropy ${\sim}2.8$ and ${\sim}4.9$ bits below nominal. Even a full MoMask-style RVQ-VAE, with temporal context our frame-local codes lack, at best draws level with the finger code and stays ${\sim}2\times$ behind the layered one---training does not beat unit alignment. The three-step progression doubles as the cumulative ablation, and every family in Fig.~\ref{fig:rd}, where the three \textbf{ours} families hold the frontier, is a $K$-sweep---a sensitivity analysis by construction.

\begin{table}[t]
\centering
\small
\setlength{\tabcolsep}{4pt}
\begin{tabular}{lccc c}
\toprule
 & \multicolumn{3}{c}{3D, matched bits ($^\circ$)} & Sph.\ \\
\cmidrule(lr){2-4}
Method & ${\sim}4.4$b & ${\sim}5.6$b & ${\sim}6.6$b & $K{=}64$ \\
\midrule
k-means        & $8.43$  & $5.64$ & $4.07$ & $6.59$ \\
VQ (learned)   & $8.53$  & $5.63$ & $4.08$ & $6.49$ \\
FSQ            & $9.88$  & $8.47$ & --     & -- \\
PQ             & $12.58$ & $9.11$ & $6.25$ & $7.36$ \\
RVQ            & $10.02$ & $6.24$ & $4.45$ & $8.71$ \\
BSQ            & $11.65$ & $9.45$ & --     & $23.57$ \\
\bottomrule
\end{tabular}
\caption{Quantizer comparison at matched empirical-entropy rate (``--'': no operating point; full sweep in App.~\ref{app:quant}). FSQ/PQ/BSQ trail: the null is over \emph{strong} quantizers, not all of them.}
\label{tab:quant}
\end{table}

\subsection{Downstream Tasks: Granularity Selects the Unit}
\label{sec:downstream}
Across six uses---generation, forecasting, retrieval, editing, denoising, and classification (Table~\ref{tab:downstream})---the winning level is ordered: dynamics read the bone, interaction the finger, and identity the whole hand. Not every row is a win---forecasting only \emph{ties} continuous copy and retrieval's spread is modest---but the ordering is not arbitrary: a task's natural unit is the coarsest one that still carries the information it needs. Dynamics turn on tiny frame-to-frame velocities, so they demand the finest, bone-level detail; interaction turns on which finger is doing what, so it aligns with the finger; identity turns on the coarse silhouette of the whole hand, so a whole-hand code suffices: a \emph{single} whole-hand token reaches $46.8\%$ top-1, matching the $20$-token per-bone code's $46.2\%$ at $1/14.6$ of its rate (App.~\ref{app:toplevel}). The rest of this section reads each task at its natural unit.

\begin{table*}[t]
\centering
\small
\setlength{\tabcolsep}{6pt}
\begin{tabular}{@{}l>{\raggedright\arraybackslash}p{1.32in}ll>{\raggedright\arraybackslash}p{1.78in}@{}}
\toprule
Unit $\to$ regime & Task & DigitCode & Reference & Note \\
\midrule
bone $\to$ dynamics      & Single-step forecasting & $\mathbf{12.90^\circ}$     & HL-26 $18.85^\circ$      & $-31\%$; \emph{ties} cont.\ copy $12.74^\circ$ \\
\;(unresolved)           & Distribution generation & AE-FGD $\mathbf{3.01}$     & HL-26 $4.46$             & HL-26 gap only; speed-JS $2.2\times$ \\
finger $\to$ interaction & Retrieval               & P@1 $\mathbf{25.4\%}$      & random $1.1\%$           & $23\times$ random (bag matching) \\
finger $\to$ interaction & Denoising               & $\mathbf{+5.5^\circ}$      & MANO refit               & at $\sigma{=}20^\circ$; refit $440\times$ slower \\
finger $\to$ interaction & Single-finger editing   & off-fing.\ $\mathbf{0^\circ}$$^{\dagger}$ & MANO $0.90$ viol.\ & exact by construction \\
hand $\to$ identity      & Gesture classification  & top-1 $\mathbf{49.7\%}$    & HL-26 $43.0\%$           & $+6.7$pp, CI $[+4.3,+9.0]$ \\
\bottomrule
\end{tabular}
\caption{Downstream tasks, ordered by the unit their granularity calls for. ``DigitCode'' is the best arm at that unit, ``Reference'' the baseline it is read against. $^{\dagger}$Exact zero by construction (the decoding map factorizes), not a measured near-zero. Generation separates the adaptive codes from HL-26 but does not resolve bone against finger, so it carries no unit label. Classification is subject-disjoint. Protocols and per-tokenizer tables in App.~\ref{app:down}.}
\label{tab:downstream}
\end{table*}

Dynamics read the bone. These tasks turn on frame-to-frame velocity, and the bone-level code wins them. Forecasting shows it most cleanly, on a deterministic metric: the per-bone code predicts the next pose to $12.90^\circ$ against the per-finger code's $19.04^\circ$ and HL-26's $18.85^\circ$, leading on three of four model classes---coarser units fail exactly as the ordering predicts, their tokens moving too much per frame to be predictable. No tokenizer beats a ``predict no motion'' baseline ($12.74^\circ$), so this is evidence on \emph{which unit}, not a win over continuous poses (App.~\ref{app:down}).

Generation supplies the mechanism. HL-26 trails both adaptive codes at every seed on the collapse-sensitive AE-FGD ($4.46$ vs.\ $3.01$), its generated speed drifting toward stillness: coarse tokens quantize away small inter-frame changes, so a model trained on them \emph{mean-collapses}. The gap is a property of the code, not the decoder---a stronger masked decoder \citep{guo2024momask,pinyoanuntapong2024mmm} does not close it---and it survives a window-matched calibration against the metric's empirical null (App.~\ref{app:repro}). It does not separate the two adaptive codes from each other, which is why the ordering above rests on forecasting.

Interaction reads the finger. People grasp, point, and press with fingers, and retrieval, denoising, and editing are three views of one token, unified by a single operation on a finger's codeword: matching it against a query, against the codebook, or overwriting it (Table~\ref{tab:downstream}). \emph{Retrieval}: symbolic matching finds same-gesture clips at $23\times$ the random rate, and composed finger predicates hold zero-shot where a trained classifier compounds errors. \emph{Denoising}: snapping a noisy finger to its nearest codeword projects it back onto the manifold of real poses in one shot, beating an iterative MANO re-fit at every noise level and ${\sim}440\times$ faster. \emph{Editing}: rewriting one finger's code moves that finger and \emph{exactly} no other, exiting the real-hand flexion envelope on $0.02\%$ of edits against $90\%$ for a matched unconstrained MANO edit (\S\ref{sec:interface}; App.~\ref{app:bimanual}).

Identity reads the whole hand. A gesture is read off the coarse silhouette, not off any bone, and the rate that buys is the striking part: on subject-disjoint $137$-class classification a \emph{single} whole-hand token reaches $46.8\%$ top-1, matching the $20$-token per-bone code's $46.2\%$ at $1/14.6$ of its rate. DigitCode-F matches HL-26 at $45\%$ of its rate, and adding the whole-hand token reaches $49.7\%$ ($+6.7$pp, CI $[+4.3,+9.0]$)---the units are complementary. The tokens also beat MANO and joint-angle features under a fixed probe (App.~\ref{app:toplevel}).

\subsection{Cross-Dataset and Cross-Regime Generalization}
\label{sec:cross}
DigitCode-A transfers across datasets and regimes: fit on one dataset it still beats HL-26 on the other ($-61\%$ on FreiHAND), and an InterHand-fit codebook dominates on HanCo and lifted ASL alike ($-46$ to $-61\%$)---generalization, not overfitting (App.~\ref{app:cross}).

Noise makes the granularity ordering visible from the other side. Discrimination under noise is an identity task and rewards coarseness---HL-26's coarse alphabet quantizes the noise away and is most robust---while fidelity reads the finest unit, so the finger code that wins clean compression collapses on noisy ASL ($14.03^\circ$), its error dominated by a single worst bone. The bone-precision layer is the fix: DigitCode-H's per-bone residual decouples the corrupted bone, reaching $4.29^\circ$ on ASL ($-69\%$ vs.\ DigitCode-F; App.~\ref{app:regime}). Coarse for identity, fine for fidelity. This is not a second measurement of the ordering but a test of it: the mechanism of \S\ref{sec:downstream} predicts that noise should reverse which unit wins, and it does. A codebook, unlike a frozen parametric prior, can also be \emph{refit} to a target skeleton ($6.75^\circ$ vs.\ $19$--$25^\circ$; App.~\ref{app:regime}).

\section{Detecting and Repairing Malformed Hands}
\label{sec:interface}

Malformed hands are the most common artifact of generative image and video models, yet the standard fix redraws the whole hand and never localizes which finger is wrong. A symbolic per-finger code supplies the missing interface, and here the \emph{genuine unit} of \S\ref{sec:method} earns its name: a finger token is \textbf{(i)} \emph{grounded}, naming a referent outside the model, \textbf{(ii)} \emph{enumerable}, so a finite codebook lists the legal set, \textbf{(iii)} \emph{closed}, since decoding lands on a prototype fit to real hands, and \textbf{(iv)} \emph{locally editable}, because the code factorizes along the structure---a property of the decoding map, not statistical independence. A learned latent over the whole hand meets only the second. Each property becomes an operation with no added machinery: the codeword residual detects and localizes a corrupted finger, the token is the editing handle, and closure keeps every output inside the real-hand set. We test all four on known-ground-truth corruptions of held-out InterHand2.6M poses (DigitCode-F, $K{=}128$) in two regimes, \emph{off-manifold} ($\sigma{=}45^\circ$) and \emph{anatomically illegal} (hyperextension), scored by the metrics of Table~\ref{tab:repair} (App.~\ref{app:interface}).

Detection needs no training (Table~\ref{tab:detect}). The per-finger \emph{codeword residual}---angular distance to the nearest codeword---is the only detector strong on both regimes: a \emph{trained} Mahalanobis detector edges it off-manifold but collapses on illegal poses, and a MANO-refit residual has no detection power at all. Distance to a closed valid set is the structural signal continuous parameterizations lack (App.~\ref{app:interface}).

\begin{table}[t]
\centering
\small
\setlength{\tabcolsep}{4pt}
\begin{tabular}{lcc}
\toprule
Detector (AUC$\uparrow$) & off-manifold & illegal \\
\midrule
codeword residual (ours, zero-training) & $0.953$ & $\mathbf{0.823}$ \\
Mahalanobis (trained)                   & $\mathbf{0.961}$ & $0.684$ \\
intra-finger incoherence                & $0.737$ & $0.615$ \\
MANO-refit residual                     & $0.428$ & $0.534$ \\
\bottomrule
\end{tabular}
\caption{Corrupted-finger detection AUC by regime.}
\label{tab:detect}
\end{table}

\begin{table}[t]
\centering
\small
\setlength{\tabcolsep}{3pt}
\begin{tabular}{lcccc}
\toprule
Repair & corr.$^\circ\downarrow$ & innoc.$^\circ\downarrow$ & plaus.$^\circ\downarrow$ & OOL$\downarrow$ \\
\midrule
none                      & $36.2$ & $0$    & $25.3$ & --- \\
snap (nearest code)       & $28.4$ & $2.6$  & $7.1$  & $0.00$ \\
token-infill              & $23.8$ & $5.3$  & $5.6$ & $0.00$ \\
whole-hand VQ             & $17.0$ & $15.8$ & $5.2$  & $0.00$ \\
MANO-PCA refit            & $52.3$ & $25.6$ & $39.0$ & $0.94$ \\
cont.\ inpaint (learned)  & $12.4$ & $2.6$ & $7.8$ & $0.00$ \\
\;\;+ codebook projection & $12.5$ & $2.7$  & $7.6$  & $0.00^{\dagger}$ \\
\bottomrule
\end{tabular}
\caption{Repair arms on off-manifold corruptions ($^{\dagger}$zero \emph{by construction}; protocol, illegal regime, and $\tau$ sweeps in App.~\ref{app:interface}).}
\label{tab:repair}
\end{table}

The detected finger is then repaired in place (Table~\ref{tab:repair}). A purely symbolic \emph{token-infill}---a small masked transformer that rewrites only the flagged finger conditioned on the other four---beats snapping on both regimes and yields the most plausible fingers of any \emph{addressable} repair. MANO-based repair, by contrast, cannot stay local: MANO's parameters are not decomposed per finger, so refitting to correct one finger perturbs the rest. Every MANO arm thus fails catastrophically---infilling one finger by fitting MANO to the four clean ones lands far off, because \emph{the parameterization carries no cross-finger conditional prior}. That prior must therefore be \emph{learned}, on an addressable substrate: a continuous inpainter running \emph{behind} the same symbolic detector---not a standalone baseline---reaches the lowest raw error, though raw error-to-GT rewards predicting the conditional mean (its fingers score $7.8^\circ$ plausibility, \emph{below} real fingers' $8.3^\circ$). Projecting its output onto the codebook costs almost nothing and buys closure \emph{by construction}. We recommend the hybrid: the symbolic layer supplies detection, closure, and addressability; a learned model supplies accuracy. Editing is the same operation from the other entry point---rewriting one finger's token hits the target exactly, moves no other finger, and lands on a codebook prototype.

A real estimator shifts the boundary. Behind MediaPipe on FreiHAND the detector still transfers, but the noise is now \emph{diffuse}---many small errors rather than a few grossly wrong fingers---and there infill-style repairs fall below no-repair while snap denoising wins. This gives a deployment rule: route by noise regime---structural corruption to context infill, diffuse estimator noise to snap---with one detector serving both. The loop closes on generative content: on hands lifted from CogVideoX frames the residual flags high-residual fingers and snap returns them to the real-pose manifold, with no ground truth (App.~\ref{app:interface}).

\section{Retargeting and Screening for Robot Hands}
\label{sec:embodied}
Learning dexterity from human video \citep{qin2022dexmv,shaw2023videodex,luo2025beingh0} chains video $\to$ hand estimate $\to$ retarget $\to$ action label, and every stage today runs on continuous poses: inverse kinematics is re-solved each frame, nothing checks that an estimated pose is reachable, and a label cannot be edited or screened. The finger unit plugs into three of these steps from one representation, with no task-specific training, on the same properties (i)--(iv) of \S\ref{sec:interface}. We retarget with a standard position optimizer \citep{qin2023anyteleop} onto six robot hands spanning independent-finger, shared-wrist, and underactuated designs (App.~\ref{app:embodied}).

The finger is the right unit here, and a compositional test says why. Fit only on FreiHAND and tested on ASL handshapes---novel finger combinations---a per-finger encoding degrades \emph{less} than a $6.4\times$ larger whole-hand codebook ($31.2^\circ$ vs.\ $34.9^\circ$ encoding residual), even though that whole-hand code is the more accurate fit in-distribution ($1.75^\circ$ vs.\ $3.98^\circ$). The coarser unit models a seen distribution better; the factored one covers combinations it has never seen---the granularity ordering of \S\ref{sec:downstream} once more, now on a robot-facing task.

The first step, retargeting, can be compiled away. Because a per-finger codebook is a finite set, the map from code to robot joints is solved \emph{once}---$640$ inverse-kinematics solves ($5$ fingers $\times$ $128$ codes), $1.6$\,s---into a per-finger lookup table, after which streaming retargeting is $O(1)$ table assembly: $480$--$1500\times$ faster than per-frame optimization on \emph{all six} hands, at $0.4$--$5.0$\,mm of accuracy cost, the low end being Allegro (Table~\ref{tab:robot}), though a whole-hand table compiles more accurately on the other five (App.~\ref{app:embodied}). The finger's contribution is addressability and amortization, not peak accuracy: editing one token re-looks-up one finger, and the table exists only because a finger is a complete IK target---a per-bone code is not, so it enters Table~\ref{tab:robot} as a timing control only ($2.5$ vs.\ $2.4$\,ms).

\begin{table}[t]
\centering
\small
\setlength{\tabcolsep}{5pt}
\begin{tabular}{lcc}
\toprule
Arm (Allegro) & task (mm) & ms/frame \\
\midrule
per-frame IK (frozen base)     & $35.1$ & $2.4$ \\
per-bone code $+$ per-frame IK & --     & $2.5$ \\
DigitCode-F lookup (compiled)  & $35.5$ & $0.005$ \\
\bottomrule
\end{tabular}
\caption{Compiled retargeting on Allegro (``--'': no value produced; whole-hand baselines and all six hands in App.~\ref{app:embodied}).}
\label{tab:robot}
\end{table}

The same residual then screens the pipeline's action labels, which otherwise have no ground-truth-free check. Illegal symbols map to targets outside the reachable set, raising the forward-kinematics residual across four hands (every CI excluding zero), so the closed codebook doubles as a reachability filter \emph{before} IK. A screen that only rejects leaves the pipeline short of data, so the same detector localizes the offending finger zero-shot and routes repair by the regime rule of \S\ref{sec:interface}---lowering label error $25$--$27\%$ for structural corruption and $13$--$15\%$ for real estimator noise (all $4/4$ hands). For labels beyond repair, detect-and-drop curation collapses a mixed stream's error to near zero (App.~\ref{app:embodied}). The interface of \S\ref{sec:interface} thus replicates on a third domain, across four hands including underactuated, coupled designs. One representation thus carries all six operations, though some of them (infill, the offline IK solve) are themselves learned or optimized. We claim no manipulation-task result; these are representation- and data-layer properties.

\section{Analysis and Discussion}
\label{sec:discussion}

Two mechanisms explain the results. The quantizer is interchangeable because the direction space is low-dimensional (participation ratio $2.8$--$4.2$), leaving a learned encoder little to add over Lloyd's algorithm---a parity that also holds downstream, where VQ-VAE tokens classify no better than DigitCode-A. At matched rate the unit outmoves the quantizer rule by more than an order of magnitude (effect decomposition, App.~\ref{app:quant}): swapping the tuned VQ for training-free k-means moves reconstruction by at most $0.10^\circ$, and holding mechanism, depth and code fixed while moving only the coarse stage's unit from bone to finger shifts it $1.63^\circ$, with the per-bone control given \emph{more} rate, not less---a per-bone coarse stage never matching a per-finger one anywhere in a $6$--$9$ bit sweep (App.~\ref{app:hier}). Since a null test cannot establish ``no difference'', we state the quantizer claim as an \emph{equivalence} against a declared margin $\delta=0.5^\circ$---under a third of the smallest \emph{positive} effect we claim ($1.63^\circ$ for the unit). The observed gap ($\le0.10^\circ$, either sign) and the paired per-sequence deltas over three VQ seeds ($\le0.15^\circ$) both sit inside $\delta$ and above the VQ's own $\pm0.04^\circ$ run-to-run floor: equivalent to within a margin too small to matter here, not merely non-significant. Depth is the one knob that goes the wrong way, costing $0.38^\circ$ at a fixed bone unit. One trade-off runs through DigitCode: a per-finger codeword is a \emph{prototype pose from real hands}---a soft prior that pays on clean data and costs off-manifold, where one corrupted bone dominates the shared code. Any finite codebook exchanges coverage for precision, and the coarser the unit the sooner it meets that limit---which is why the design carries both units at once, the per-bone residual (\S\ref{sec:hier}) covering it passively and the infill (\S\ref{sec:interface}) actively.

Our claims are scoped in three ways. The evidence is strongest on reconstruction, robustness, and embodied screening, where every headline effect carries a significance test or paired bootstrap CI (App.~\ref{app:signif}); the generative studies are more exploratory. The ASL results are best read as relative comparisons, their ground truth being lifted and noisy, so those degrees index improvement rather than absolute calibration. And the robot results sit at the representation and data layer, establishing that the finger unit \emph{compiles and screens}; whether it also improves downstream control is for the next system built on it. A full system-level comparison also needs a common grid the field does not yet share---what \textsc{HandTok} ships the harness for.

\section{Conclusion}
We started from HL's observation that hand motion can be written as symbols, and asked what those symbols should cover. The answer is not one unit but a hierarchy: a finger token for the interface, a bone residual for precision, both carried in one code. The design is simple---spherical k-means, concatenation, one rotation---yet at matched rate it holds the frontier against learned alternatives, because the gain is structural: a finger's bones are coupled, and no quantizer can undo redundancy that was baked in by the wrong unit. The more portable finding is methodological. Tokenization research routinely varies the quantizer at a fixed, inherited unit; our harness varies both, and the unit is the larger lever---swapping a tuned learned quantizer for training-free k-means at a fixed unit moves nothing measurable, while moving the unit reshapes the frontier. Aligning a token to a genuine anatomical part also turns a code into an interface: one per-finger codeword detects corruption with no training, repairs it in place, and compiles retargeting into a lookup table---operations no whole-hand parameterization can perform on a single finger without disturbing the rest. Whether the same unit-first principle governs other compositional signals is open, but \textsc{HandTok} makes it testable: the same harness accepts any skeleton topology and any quantizer family. The finger is not the only unit worth designing for; it is the proof that the unit is worth designing for at all.

\clearpage
\bibliography{hlref}

\clearpage
\appendix
\section*{Technical Appendix}
\noindent This appendix is the technical supplement reviewed with the paper, inlined verbatim:  reproducibility details, full protocols and tables for every deferred study, and a consolidated table of negative and boundary results (Sec.~\ref{app:neg}). All numbers are on held-out, capture-disjoint splits; the HL-26 baseline is $14.7$--$14.85^\circ$ on InterHand2.6M across two independent encoders, and $14.4$--$14.9^\circ$ on the other three regimes.

\noindent\textbf{Where to check what.} After an opening block on protocol (Secs.~\ref{app:repro}--\ref{app:signif}), the appendix sections follow the order of the main text above: the three method decisions, the time axis, the quantizer sweep, the downstream tasks, generalization, and the two applications, closing with the results that did not work. This index maps each load-bearing claim to the section carrying its full protocol, table and controls.

\begin{table}[h]
\centering
\small
\setlength{\tabcolsep}{4pt}
\begin{tabular}{@{}>{\raggedright\arraybackslash}p{2.02in}cl@{}}
\toprule
Claim & Main & Here \\
\midrule
The unit moves the R--D frontier; the quantizer rule does not & \S4.2 & \ref{app:quant}, \ref{app:signif} \\
\;\;where to cut a finger (grouping control) & \S3.2 & \ref{app:grouping} \\
\;\;the hierarchy, under a single-variable control & \S3.3, \S7 & \ref{app:hier} \\
\;\;a fixed geometry cannot substitute & \S3 & \ref{app:geom} \\
Task granularity selects the unit & \S4.3 & \ref{app:down}, \ref{app:toplevel}, \ref{app:bimanual} \\
Noise regime reverses which code wins & \S4.4 & \ref{app:regime}, \ref{app:cross} \\
A finger token is an editable interface & \S5 & \ref{app:interface}, \ref{app:bimanual} \\
The finger unit reaches the robot & \S6 & \ref{app:embodied} \\
Time-axis coding (relative, event, schedule) & \S3.4 & \ref{app:rel}, \ref{app:event}, \ref{app:schedule} \\
\midrule
Splits, metrics, seeds, compute & \S4.1 & \ref{app:repro} \\
What did \emph{not} work, and why (appendix-only) & --- & \ref{app:neg} \\
\bottomrule
\end{tabular}
\end{table}

\section{Reproducibility and Implementation Details}
\label{app:repro}

\begin{figure*}[t]
\centering
\includegraphics[width=0.98\linewidth]{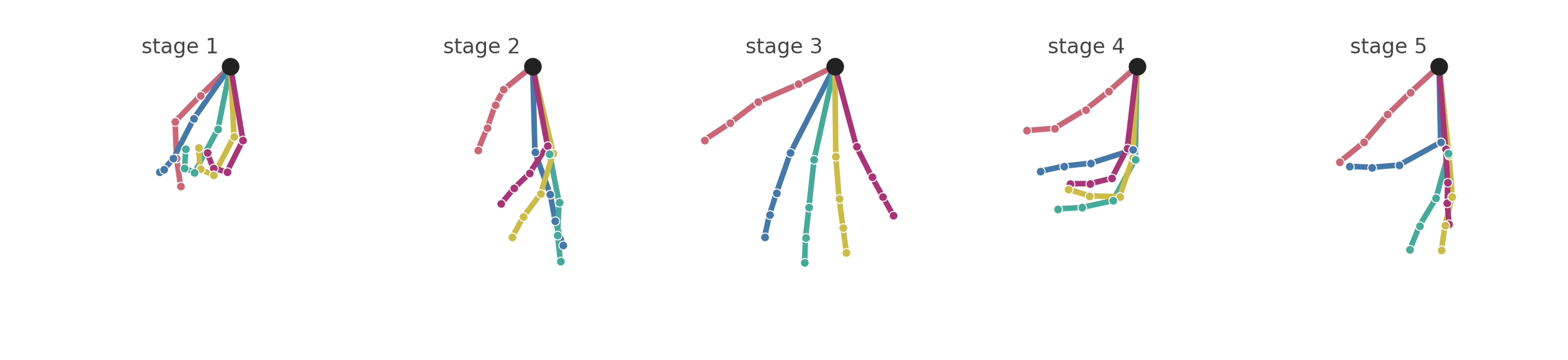}
\caption{A hand gesture from InterHand2.6M as a 3D joint skeleton at five stages (per-finger colors; the wrist root in black). HL extracts $20$ bone-direction vectors from these $21$ joints per hand and quantizes each to a direction token.}
\label{fig:handstages}
\end{figure*}

\paragraph{Data.}
We use the full InterHand2.6M \citep{moon2020interhand} ($36$ captures, $717$K bone-direction vectors from two-hand sequences) and the full FreiHAND \citep{zimmermann2019freihand} ($32{,}560$ single-hand poses). For each hand we extract $20$ regional bone-direction vectors following HL \citep{li2024hl} and normalize them to the unit sphere $S^2$. Stacking two hands per frame yields the $T\times 40$ grid. Figure~\ref{fig:handstages} shows a sample gesture as the underlying $21$-joint 3D skeleton.

\paragraph{Splits.}
All evaluations are held-out and disjoint by capture. For InterHand2.6M we fit on even-indexed captures and test on odd-indexed captures (and the reverse for symmetry checks); for sequence tasks we use \texttt{seqs[::2]} for training and \texttt{seqs[1::2]} for testing. Adaptive codebooks are fit on the training partition only and evaluated on the held-out partition; we never use a random per-frame split, which would leak near-duplicate adjacent frames. Gesture classification uses a stricter split: the official subject annotations show that subjects span multiple captures (e.g.\ subject~$0$ appears in Captures $0$, $4$, and $17$; subject~$1$ in Captures $1$ and $2$), so an even/odd capture split leaks signer identity for classification; we therefore assign $14$ subjects to train and $13$ to test with zero overlap and quote only this subject-disjoint split.

\paragraph{Metrics.}
\emph{Distortion} is the mean angular error (degrees) between the reconstructed and ground-truth unit direction, $\arccos(\hat{\mathbf{v}}^\top\mathbf{v})$, averaged over all bones and frames. For \emph{rate} we follow the main paper: headline rate--distortion figures and operating-point tables use the deployable \emph{fixed-width} cost $\log_2 K$ bits per covered bone, while quantizer bake-offs and compressibility studies (relative/event coding, gzip, usage skew) use empirical token \emph{entropy} per covered bone; each table states which rate it reports. Distortion comparisons are made at matched rate. Unless noted, reconstruction means are frame-pooled; the significance study (Sec.~\ref{app:signif}) uses sequence-weighted means, its resampling unit.

\paragraph{What the HL-26 reference is.}
HL-26 is the reference in nearly every table, and it was not designed to minimize angular error: like the Labanotation tradition it comes from, its alphabet was chosen to be small, fixed, and human-transcribable. Rate--distortion is nonetheless the right axis, for two reasons. It is the axis the notation line itself adopted---the grid's value rests on \emph{bidirectional} conversion \citep{daskel}, and a code that cannot reconstruct cannot be converted back. And our claim does not rest on it: ``at matched rate the unit moves error and the quantizer rule does not'' is settled by controlled contrasts among our \emph{own} variants (Tables~\ref{tab:grouping},~\ref{tab:effect}), with HL-26 entering as one operating point rather than as the thing refuted. Where HL-26's own objective is the relevant one---a coarse, noise-tolerant, identity-bearing symbol---it wins, and we report that (Tables~\ref{tab:signif},~\ref{tab:retr}; Sec.~\ref{app:regime}).

\paragraph{Coupling statistics.}
Mutual information between bone tokens rises along a finger's kinematic chain, reaching $59\%$ toward the fingertip; \emph{cross}-finger MI peaks at $51\%$ (middle--ring) with the thumb most independent (${\sim}27\%$), and averages $39\%$ within a finger vs.\ $29\%$ across fingers. Temporally, $79\%$ of HL-26 tokens and $66\%$ of DigitCode-A ($K{=}64$) tokens are unchanged from the previous frame---strong dependence inside a finger to exploit, a weak seam between fingers to cut, and heavy frame-to-frame redundancy to compress. The pairwise numbers are corroborated by a \emph{partition}-level statistic (Sec.~\ref{app:schedule}): among candidate ways of cutting the $20$ bones into groups, grouping by finger carries the most within-group MI ($0.39$), above grouping by joint level ($0.33$) and above a cross-finger baseline ($0.29$), so the anatomical partition is also the MI-preferred one. This is the statistic the token boundary rests on; the $51\%$ middle--ring peak is a single \emph{pair} and does not tile into five disjoint groups.

\paragraph{Generation metrics (AE-FGD and speed-JS).}
Our primary generation metric is a \emph{motion-AE latent FGD} (a Fr\'echet gesture/motion distance computed in a learned motion-autoencoder latent space; AE-FGD): a small motion autoencoder (window $W{=}24$) is trained once per dataset on real training motion (seed $0$, $1{,}500$ steps), and the Fr\'echet distance is computed between Gaussians fit to its latent codes on generated vs.\ held-out real clips. All arms and seeds of a study share the same feature space; CIs are clip-level bootstraps ($B{=}500$). Unlike a fixed statistic-feature distance, the learned latent is \emph{collapse-sensitive}: a near-still generator scores poorly. We report it alongside the speed-distribution Jensen--Shannon divergence. An earlier hand-crafted $45$-D motion-statistics FGD is retained only as a recorded negative control: it is insensitive to mean-collapse and its cross-seed std ($\pm283$--$424$ on the AR study) flips rankings between seeds, so no conclusion rests on it; it appears only in the exploratory single-budget schedule sweep (Sec.~\ref{app:schedule}), where only separations far larger than that std are read. AE-FGD values remain comparable only \emph{within} a table at matched setup: the event-compression (Sec.~\ref{app:event}), decoding-schedule (Sec.~\ref{app:schedule}), and generation (Table~\ref{tab:gen}) studies use different pipelines and windows, so values must not be compared across tables. Neither metric is comparable to body-motion FGD numbers in the literature.

\paragraph{Finite-sample calibration of AE-FGD.}
A plug-in Fr\'echet distance is upward biased at small sample size, and our latent is $64$-D, so we calibrate rather than assume. Two facts matter. First, the Gaussian is fit over sliding \emph{windows}, not clips, so the covariance sample size is much larger than the clip count ($120$--$341$ windows for the $16$-clip AR study, $63$ for the $9$-clip masked study, $291$--$521$ for the schedule study). Second, arms that generate clips of different lengths contribute different window counts, so each arm carries a different bias. We therefore recomputed every AE-FGD after subsampling all arms \emph{and} the real reference to a common window count ($200$ draws), and measured the empirical null---the distance between two disjoint halves of the \emph{real} windows at that same size, i.e.\ the score a perfect generator would post.

The null scales as ${\approx}50/n$, the $O(d/n)$ form finite-sample theory predicts: $0.81$ at $n{=}63$, $1.09$ at $n{=}69$, and $0.15$--$0.18$ at $n{\approx}300$. Against it, two conclusions harden and one does not. The HL-26 gap survives matching at every seed and both decoders (AR $5.40$ vs.\ $3.87$/$3.96$; masked $7.15$ vs.\ $5.92$/$6.17$), a margin $1.3$--$1.9\times$ the null. The decoding-schedule ordering also survives at every seed (delay $1.94$ vs.\ parallel $2.34$ vs.\ flat $2.53$), a margin $2$--$6\times$ its null, which is small there because that study has the most windows. But the DigitCode-A vs.\ DigitCode-F margin shrinks from $0.76$ to $0.09$ under matching and inverts at one of three AR seeds, placing it \emph{below} the null floor: AE-FGD separates HL-26 from the adaptive codes, and does not separate the two adaptive codes from each other at this sample size. This is why the main paper draws only the HL-26 comparison from the generation study without qualification, and it is also the quantitative form of the caution that AE-FGD values are comparable only within a table. Recomputation is offline---per-clip generated motions are dumped once---so it retrains nothing.

\paragraph{One operating point, two conventions.}
DigitCode-A at $K{=}26$ appears as two consistent numbers, differing only in pipeline and rate convention: $8.45^\circ$ (frame-pooled, fixed-width rate; the R--D tables and the significance study, whose resampling weights each sequence by its vector count) and $9.08^\circ$ (from the relative-coding pipeline at entropy rate; Table~\ref{tab:rel}). These are convention choices, not disagreements. The same applies to per-bone $K{=}64$ k-means, which is fit independently in four studies: $5.69^\circ$ under the R--D sweep pipeline (the progression tables and Method), $5.68^\circ$ under the robustness pipeline (Table~\ref{tab:robustfull}), $5.64^\circ$ under the quantizer bake-off pipeline (Table~\ref{tab:quantfull} and the seed study), and $5.58^\circ$ under the granularity-sweep pipeline (Table~\ref{tab:bound}). The spread ($\le 0.11^\circ$) reflects fitting subsets and initialization.

\paragraph{Why the quantizer null is measured within a pipeline.}
That $\le 0.11^\circ$ spread must not be confused with the $0.15^\circ$ paired k-means-vs-VQ deltas, and the learned VQ varies the same way: at $K{=}26$ it is $8.36^\circ$ under the R--D sweep pipeline (the main paper's progression table) and $8.53^\circ$ under the bake-off pipeline (Table~\ref{tab:quantfull}). These are near but not identical operating points---the achieved entropies are $4.47$ and $4.53$ bits---yet the $0.17^\circ$ spread still exceeds the ${\le}0.10^\circ$ k-means-vs-VQ gap we call null. That is why the comparison is made within a pipeline and never across. Within each pipeline the gap stays ${\le}0.10^\circ$ and flips sign (R--D sweep: k-means $8.45$ vs.\ VQ $8.36$; bake-off: k-means $8.43$ vs.\ VQ $8.53$), and the tie holds in all of them. This ${\le}0.11^\circ$ statement covers the per-bone code only; the per-finger unit shows a much larger unreconciled cross-pipeline gap, which we flag explicitly in the caption of Table~\ref{tab:bound} rather than fold into this convention note.

\paragraph{Compute.}
Runs use NVIDIA A800 ($80$\,GB) GPUs (driver 535) under Linux, with Python 3.11, PyTorch 2.5.1 (CUDA 12.1), and HuggingFace Transformers; the generative-video study additionally uses Diffusers 0.39 (CogVideoX-2b) and MediaPipe's hand landmarker. The geometric k-means/k-medoids codebooks are lightweight and CPU-computable, as are the repair-interface models; the released \textsc{HandTok} environment pins exact versions. The generation, masked-generation, event, and decoding-schedule studies are trained at $3$ seeds and reported as mean$\pm$std with clip-level bootstrap CIs; the significance study (Sec.~\ref{app:signif}) uses $10{,}000$ bootstrap/permutation replicates; remaining tables are single-run.

\paragraph{Model hyperparameters and seeds.}
All sequence models share one configuration, fixed a priori and not tuned per task: a LLaMA-style decoder with $6$ layers, hidden size $320$, $10$ attention heads, intermediate size $4\times$, and RoPE $\theta=10^4$, trained with AdamW (lr $3\times10^{-4}$, weight decay $0.01$), a cosine schedule with $3\%$ warmup, and batch size $32$. The codebook size $K$ is the only swept hyperparameter (per-bone $K\in\{26,64,128\}$, per-finger $K\in\{64,128,256\}$, hierarchical $(K_1,K_2)=(128,\{8,16,32\})$), selected by matched-rate comparison in the R--D tables; the repair-interface conclusions are checked for stability over $K\in\{64,128,256\}$ (Sec.~\ref{app:interface}). Randomness policy: single-run analyses fix seed $0$ throughout (NumPy \texttt{default\_rng(0)}, \texttt{torch.manual\_seed(0)}); the multi-seed studies above retrain with three distinct integer seeds and report mean$\pm$std; the motion-AE behind AE-FGD is trained once at seed $0$ and shared by all arms of a study.

\paragraph{Quantizers.}
Spherical k-means uses cosine assignment with a mean-then-renormalize update; k-medoids restricts centers to observed directions. Learned VQ/RVQ use a straight-through estimator with codebook EMA; FSQ \citep{mentzer2024fsq}, PQ \citep{jegou2011pq}, and BSQ \citep{zhao2024bsq} follow their standard formulations. For autoregressive forecasting and generation we train a small LLaMA-style transformer \citep{touvron2023llama} over the token grid; the masked decoder follows MoMask/MMM-style confidence decoding \citep{guo2024momask,pinyoanuntapong2024mmm}.

\section{Statistical Significance}
\label{app:signif}
We attach $95\%$ confidence intervals and paired permutation tests to the headline claims, all five rows now from a single protocol and run. The cardinal rule is that the resampling unit equals the split unit---a test sequence or sign sample, never an individual frame, since adjacent frames are near-duplicates and a per-frame bootstrap fabricates tight intervals. For reconstruction and robustness we resample test \emph{sequences}, each weighted by its (frame, bone) count, so the point estimate matches the frame-pooled R--D tables; for classification we resample test samples and average the top-1 hit. Every comparison is paired (the same units scored by both methods), so the permutation test flips the sign of each unit's A$-$B difference ($N=10{,}000$, two-sided, add-one smoothed; seed $0$); CIs are $10{,}000$-replicate percentile bootstraps.

Table~\ref{tab:signif} shows all five headline effects are significant at $p<10^{-4}$ with CIs bounded away from zero---including the deliberate reversal (on noisy ASL handshape the coarse HL-26 \emph{beats} the per-finger$+$whole-hand combination code, $+11.3$ points), which the permutation test confirms is real rather than apparent. For reconstruction we compare at matched code count ($K{=}26$); the adaptive code also uses its budget more fully ($4.38$ vs.\ $3.29$ bits, HL-26's usage being skewed).

\begin{table}[t]
\centering
\footnotesize
\setlength{\tabcolsep}{3pt}
\begin{tabular}{@{}p{0.34\linewidth}ccc@{}}
\toprule
Claim (A $\to$ B) & A & B & $\Delta$ [95\% CI] \\
\midrule
Recon., HL-26 $\to$ DC-A ($K{=}26$), IH ($^\circ$) & $14.71$ & $8.45$ & $-6.26\,[-6.48,-6.03]$ \\
Gesture top-1, HL-26 $\to${+}whole-hand, subj.-disjoint (\%) & $43.0$ & $49.7$ & $+6.7\,[+4.3,+9.0]$ \\
ASL handshape, HL-26 $\to$ DC-F{+}WH \emph{(reversal)} (\%) & $31.4$ & $20.0$ & $-11.3\,[-12.4,-10.3]$ \\
Robust.\ ASL-within, DC-F $\to$ DC-H ($^\circ$) & $14.03$ & $4.29$ & $-9.74\,[-9.84,-9.64]$ \\
Robust.\ IH\,$\to$\,ASL transfer, DC-F $\to$ DC-H ($^\circ$) & $24.68$ & $8.95$ & $-15.73\,[-15.86,-15.60]$ \\
\bottomrule
\end{tabular}
\caption{Five headline effects, single protocol and run (resampling unit $=$ test sequence or sample, \emph{never} a frame; $10{,}000$-replicate bootstrap CIs and paired sign-flip permutation tests, seed $0$); all $p<10^{-4}$. DC-A/F/H $=$ DigitCode-A/F/H, WH $=$ whole-hand token. The gesture row uses the subject-disjoint split, not the identity-leaky one. Rows $3$--$5$ are ASL within-regime relative comparisons; row $3$ is the intended noise reversal (coarse HL-26 beats the finer code under noise).}
\label{tab:signif}
\end{table}

\section{Geometric Bases and Spherical Diagnosis}
\label{app:geom}

\begin{figure}[h]
\centering
\includegraphics[width=0.7\linewidth]{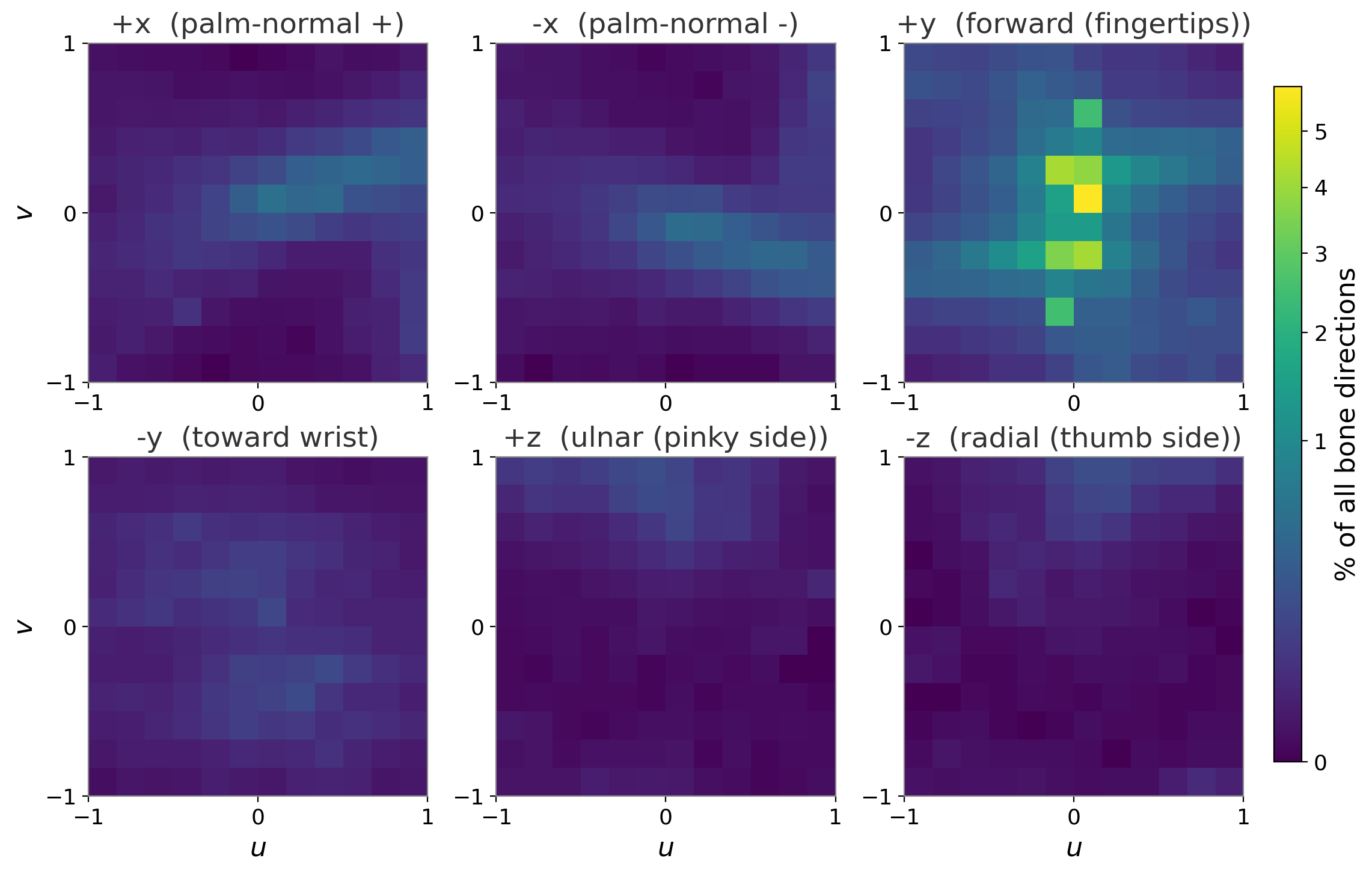}
\caption{Empirical distribution of hand-bone directions on $S^2$ (held-out InterHand2.6M). Directions concentrate heavily on the $+Y$ face, leaving HL-26's uniformly placed codes under-used elsewhere.}
\label{fig:density}
\end{figure}

\begin{figure*}[t]
\centering
\includegraphics[width=0.62\linewidth]{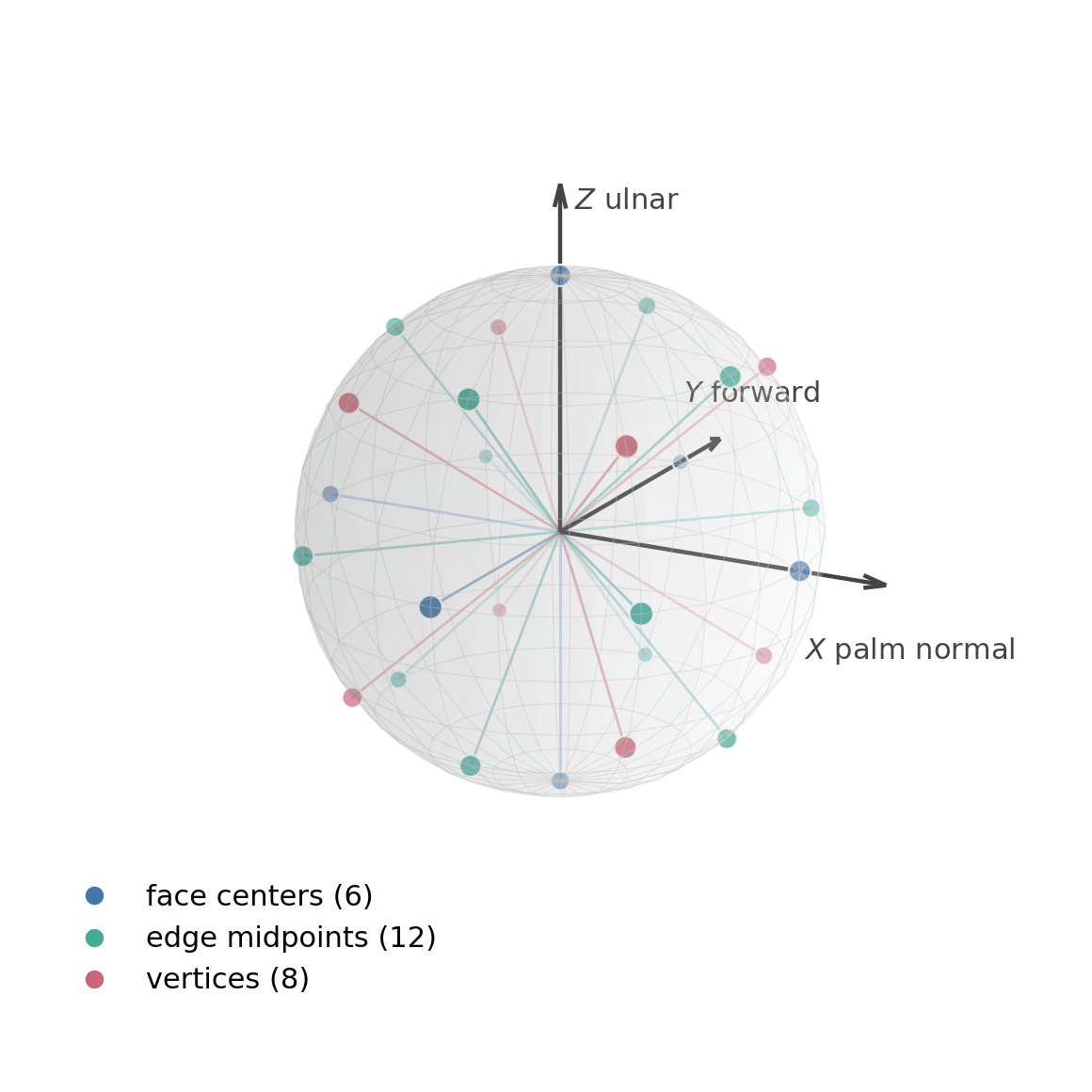}
\caption{The HL-26 alphabet: $26$ directions on a cube surface ($6$ face centers, $12$ edge midpoints, $8$ vertices)---$\{-1,0,1\}^3\setminus\{0\}$ normalized to $S^2$---shown in the hand-canonical frame ($X$ palm normal, $Y$ forward/fingertips, $Z$ ulnar). HL-26 quantizes each bone direction to the nearest of these by argmax cosine.}
\label{fig:hl26codebook}
\end{figure*}

Table~\ref{tab:geom} reports fixed (data-agnostic) codebooks that vary only the geometry; Figure~\ref{fig:hl26codebook} shows the HL-26 alphabet these variants perturb. Densifying or re-coordinatizing a uniform code is \emph{not} the win: a naive uniform spherical grid is worse than HL-26. The empirical direction density (Fig.~\ref{fig:density}) shows why: bone directions concentrate heavily on the $+Y$ face, so a uniform alphabet wastes most of its codes on near-empty regions.

\begin{table}[t]
\centering
\small
\begin{tabular}{lccc}
\toprule
Scheme & $K$ & entr.\ b. & Ang.\ ($^\circ$) \\
\midrule
HL-26 (cube surface) & $26$ & $3.29$ & $14.71$ \\
Spherical $\theta\phi$ uniform $5{\times}8$ & $40$ & $3.64$ & $16.79$ \\
Spherical pole-corrected $8{\times}14$ & $74$ & $4.71$ & $8.63$ \\
Cube-map $4{\times}4$ & $96$ & $4.98$ & $9.38$ \\
Fibonacci near-uniform & $128$ & $5.47$ & $6.96$ \\
\bottomrule
\end{tabular}
\caption{Geometric-basis study (fixed codebooks, held-out InterHand2.6M). A uniform $\theta\phi$ grid is worse than HL-26; pole correction and Fibonacci sampling recover quality but remain above the data-adaptive codes of the main paper.}
\label{tab:geom}
\end{table}

\paragraph{The anisotropy figure, and how it is counted.}
The main paper's $68.1\%$ is measured as follows. Assign every valid bone direction to the nearest of the six cube \emph{faces}---the face whose axis carries the largest $|v_i|$, taken with the sign of that component---and count. Over all $717{,}543$ valid direction vectors of InterHand2.6M ($36$ captures, both hands, per-bone validity mask; \emph{not} restricted to fully valid frames) the six faces receive
\[
\underbrace{68.1}_{+Y}\;\;\underbrace{10.7}_{+X}\;\;\underbrace{10.4}_{-X}\;\;\underbrace{5.4}_{-Y}\;\;\underbrace{2.8}_{+Z}\;\;\underbrace{2.5}_{-Z}\;\;\%,
\]
against $16.7\%$ per face under a uniform distribution: a single face carries more than two-thirds of the mass and three faces carry under $11\%$ between them. Counting instead by \emph{quantized codeword}---the fraction whose HL-26 code is one of the nine alphabet entries with $y{>}0$---gives $82.4\%$; we report the face-assignment number because it measures the data, not the alphabet. Either way the conclusion is the one Fig.~\ref{fig:density} shows: a uniform alphabet spends most of its symbols where there is almost nothing to encode.

\paragraph{Why naive spherical coordinates underperform.}
The naive spherical-grid row fails because of the setup, not because spherical coordinates are intrinsically poor. A naive $(\theta,\phi)$ grid places its pole on the $+Z$ axis---the \emph{sparsest} direction for hand bones---so cells crowd where there is no data ($42$--$50\%$ dead codes at $K{=}128$/$24$). Two fixes recover most of the gap: moving the pole to $+Y$ (the dense ``forward'' axis) improves the error from $15.35^\circ$ to $10.41^\circ$, and switching to equal-area cells improves it further to $10.27^\circ$ (a separate pole-and-cell-shape sweep at higher $K$, not the $K{=}40$ row of Table~\ref{tab:geom}), comparable to Fibonacci sampling. At the low-rate end the adaptive code needs very little budget: per-bone k-means at $K{=}12$ ($3.59$ bits) already reaches $12.7^\circ$, below HL-26 at fewer bits.

\section{Grouping Control: Anatomy or Blocklength?}
\label{app:grouping}

Moving from a per-bone to a per-finger code changes two things at once: the anatomical unit, and the vector-quantization \emph{blocklength} ($3$-D to $12$-D). Classical block quantization \citep{jegou2011pq} gains from blocklength alone for \emph{any} grouping of correlated dimensions, so an anatomical partition might be doing no work beyond making the blocks larger. This control separates the two.

We hold the block size ($5$ groups $\times$ $4$ bones $=12$-D), the codebook size $K$, the fitting pipeline, the split, and the metric fixed, and vary only \emph{which} bones share a group: \emph{within-finger} (DigitCode-F itself); \emph{cross-finger}, joint level $j$ of index/middle/ring/pinky with the thumb kept whole---deliberately \emph{half}-anatomical; \emph{diagonal}, a Latin square in which group $g$ takes joint $j$ from finger $(g{+}j)\bmod 5$, so every group spans four different fingers and four different joint levels; and ten draws of \emph{random} disjoint quartets. A \emph{joint-level} partition ($4$ groups $\times$ $5$ bones) is reported apart because its blocklength differs. The within-finger arm reproduces DigitCode-F's published operating points to three decimals ($7.100^\circ$ at $1.46$ bits/bone for $K{=}64$; $5.504^\circ$ at $1.94$ for $K{=}256$), which anchors the control to the main paper's numbers rather than to a re-implementation.

\begin{table}[t]
\centering
\small
\setlength{\tabcolsep}{4pt}
\resizebox{\columnwidth}{!}{%
\begin{tabular}{@{}lcccc@{}}
\toprule
& \multicolumn{2}{c}{$K{=}64$} & \multicolumn{2}{c}{$K{=}256$} \\
\cmidrule(lr){2-3}\cmidrule(lr){4-5}
Partition & entr.\ b. & Ang.\ ($^\circ$) & entr.\ b. & Ang.\ ($^\circ$) \\
\midrule
Within-finger (DC-F) & $1.46$ & $\mathbf{7.10}$ & $1.94$ & $\mathbf{5.50}$ \\
Cross-finger (thumb whole)  & $1.43$ & $8.91$ & $1.90$ & $7.08$ \\
Random quartets ($10\times$) & $1.45$ & $9.48{\pm}0.23$ & $1.93$ & $7.52{\pm}0.19$ \\
Diagonal (Latin square)     & $1.45$ & $9.65$ & $1.93$ & $7.69$ \\
\midrule
\emph{Joint level} ($15$-D) & $1.14$ & $10.34$ & $1.53$ & $8.40$ \\
\bottomrule
\end{tabular}}
\caption{Grouping control on held-out InterHand2.6M: identical $12$-D blocks, identical $K$, matched rate (all within $0.03$ entropy bits/bone), identical pipeline and split---only the bone-to-group assignment changes. Random quartets are mean$\pm$std over $10$ draws (range $[9.07,9.76]$ at $K{=}64$, $[7.16,7.77]$ at $K{=}256$). The last row uses a different blocklength and is not rate-matched to the others.}
\label{tab:grouping}
\end{table}

The anatomical partition wins by $1.8$--$2.6^\circ$ at $K{=}64$ ($2.4$--$2.5^\circ$ against the random and diagonal arms, $1.8^\circ$ against the half-anatomical cross-finger one) and $1.6$--$2.2^\circ$ at $K{=}256$, and sits below the \emph{best} of ten random draws at both sizes ($9.07^\circ$ and $7.16^\circ$)---roughly ten standard deviations below the random mean. Blocklength cannot be the mechanism: every arm above the rule quantizes the same $12$-D blocks at the same $K$ and within $0.03$ bits of the same rate. What separates them is a dose--response in how much anatomy survives: keeping one finger intact (cross-finger) recovers about a third of the gap to the fully anatomical partition, while the diagonal arm---built to sever every intra-finger chain---is the worst $12$-D arm at both $K$. The blocklength effect is real and is what lifts every $12$-D arm above a per-bone code; the anatomical grouping is what turns it from a modest gain into the frontier. The joint-level arm is not directly comparable (larger blocks, fewer tokens, lower rate), but its ordering is consistent: grouping the same joint across fingers is worse than grouping a finger.

\section{Hierarchical Decomposition}
\label{app:hier}

Table~\ref{tab:hier} details the coarse+residual code: the per-finger coarse layer captures ``which pose'' and the per-bone residual refines ``how much,'' and the two carry near-independent information (coarse$\to$residual MI only $10\%$), so the residual is cheap.

\begin{table}[t]
\centering
\small
\setlength{\tabcolsep}{4pt}
\begin{tabular}{lccc}
\toprule
Scheme & entr.\ b. & fixed b. & Ang.\ ($^\circ$) \\
\midrule
Coarse-only (per-finger)         & $1.70$ & $1.75$ & $6.18$ \\
Coarse $+$ residual ($K_2{=}8$)  & $4.37$ & $4.75$ & $3.26$ \\
Coarse $+$ residual ($K_2{=}16$) & $5.20$ & $5.75$ & $2.55$ \\
Coarse $+$ residual ($K_2{=}32$) & $6.12$ & $6.75$ & $1.86$ \\
\midrule
Flat per-bone (ref.)             & $6.55$ & $7.00$ & $4.07$ \\
Per-bone $2$-stage RVQ (ref.)    & ${\sim}6.6$ & --- & $4.45$ \\
\bottomrule
\end{tabular}
\caption{Hierarchical coarse+residual vs.\ the \emph{stronger} of two per-bone controls (held-out InterHand2.6M, $K{=}128$): flat one-stage k-means ($4.07^\circ$ at $6.55$ entropy b) and two-stage per-bone RVQ ($4.45^\circ$ at ${\sim}6.6$ b---the same residual at the bone unit; Table~\ref{tab:quantfull}). The hierarchy beats both at \emph{lower} rate ($1.86^\circ$ at $6.12$ b; $6.75$ fixed-width), and already reaches $3.26^\circ$ at essentially HL-26's rate ($4.75$ vs.\ $4.70$ fixed bits, $K_2{=}8$). Per-bone codes do eventually pass it, but need $1.5$--$1.8\times$ its rate to do so (flat k-means $1.52^\circ$ at $9.36$ b; RVQ $1.18^\circ$ at $10.75$ b, against the hierarchy's $6.12$ b)---the claim is a better R--D point, not an unconditional one.}
\label{tab:hier}
\end{table}

\paragraph{Unit control: same mechanism, only the coarse grouping changes.}
The comparators above each differ from DigitCode-H in more than the unit---the flat per-bone code differs in depth, the per-bone RVQ in quantizer family---so neither isolates the variable the paper's claim is about. Table~\ref{tab:unitctrl} runs the strict control. It is the \emph{same function}: euclidean $k$-means on a group of bone directions, renormalise, rotate each bone's residual to the canonical pole, one shared spherical $k$-means on the residual. Fitting subset, iteration counts, seed, split, and metric are shared code. The only thing that changes is how the $20$ bones are partitioned for the coarse stage: five groups of four (per finger) or twenty groups of one (per bone). This is the grouping control of Sec.~\ref{app:grouping} applied to the hierarchy instead of to a single layer.

The per-bone coarse stage never reaches the per-finger one anywhere in the sweep. At the nearest matched point the per-bone arm is given \emph{more} rate on both conventions ($7.00$ vs.\ $6.75$ fixed-width, $6.79$ vs.\ $6.12$ entropy) and is still $1.63^\circ$ worse; it needs $+1.25$ bits/bone to come within $0.83^\circ$, and at $+2.25$ bits/bone ($9.00$ fixed-width, a third more rate) it is still behind ($2.06$ vs.\ $1.88^\circ$). We report $1.63^\circ$ as the unit effect in Table~\ref{tab:effect} rather than the larger figures the looser comparators give, since it is the only one of the three that holds mechanism and depth fixed while giving the per-bone control more rate.

\begin{table}[t]
\centering
\small
\setlength{\tabcolsep}{4pt}
\begin{tabular}{@{}llccc@{}}
\toprule
Coarse unit & $(K_1,K_2)$ & fixed b. & entr.\ b. & Ang.\ ($^\circ$) \\
\midrule
per finger (DigitCode-H) & $(128,32)$ & $6.75$ & $6.12$ & $\mathbf{1.88}$ \\
per finger (DigitCode-H) & $(128,16)$ & $5.75$ & $5.18$ & $2.52$ \\
\midrule
per bone (control) & $(8,8)$   & $6.00$ & $5.57$ & $5.21$ \\
per bone (control) & $(32,4)$  & $7.00$ & $6.79$ & $3.51$ \\
per bone (control) & $(16,8)$  & $7.00$ & $6.57$ & $3.60$ \\
per bone (control) & $(8,16)$  & $7.00$ & $6.40$ & $3.85$ \\
per bone (control) & $(16,16)$ & $8.00$ & $7.41$ & $2.71$ \\
per bone (control) & $(32,16)$ & $9.00$ & $8.30$ & $2.06$ \\
\bottomrule
\end{tabular}
\caption{Unit control for the hierarchy (held-out InterHand2.6M). Identical two-stage geometric mechanism, identical fitting code, seed and split; only the coarse stage's bone-to-group partition differs. The per-bone arm does not match the per-finger one at any rate in the sweep, including at $9.00$ fixed-width bits---a third more rate than DigitCode-H spends. The first row reproduces the published $1.86^\circ$ operating point to within $0.02^\circ$ under this pipeline.}
\label{tab:unitctrl}
\end{table}

\paragraph{Spatial, not temporal, decoupling (boundary result).}
We tested whether the two layers form a slow--fast temporal decomposition. They do not: the coarse and residual layers change at almost the same per-frame rate ($47\%$ vs.\ $48\%$). The hierarchy is therefore a \emph{spatial} decoupling (whole-finger pose vs.\ per-bone refinement), not a temporal one---a per-finger code at $K{=}128$ is already fine enough to move every frame.

\section{Relative Encoding and Temporal Structure}
\label{app:rel}

\begin{figure*}[t]
\centering
\includegraphics[width=\linewidth]{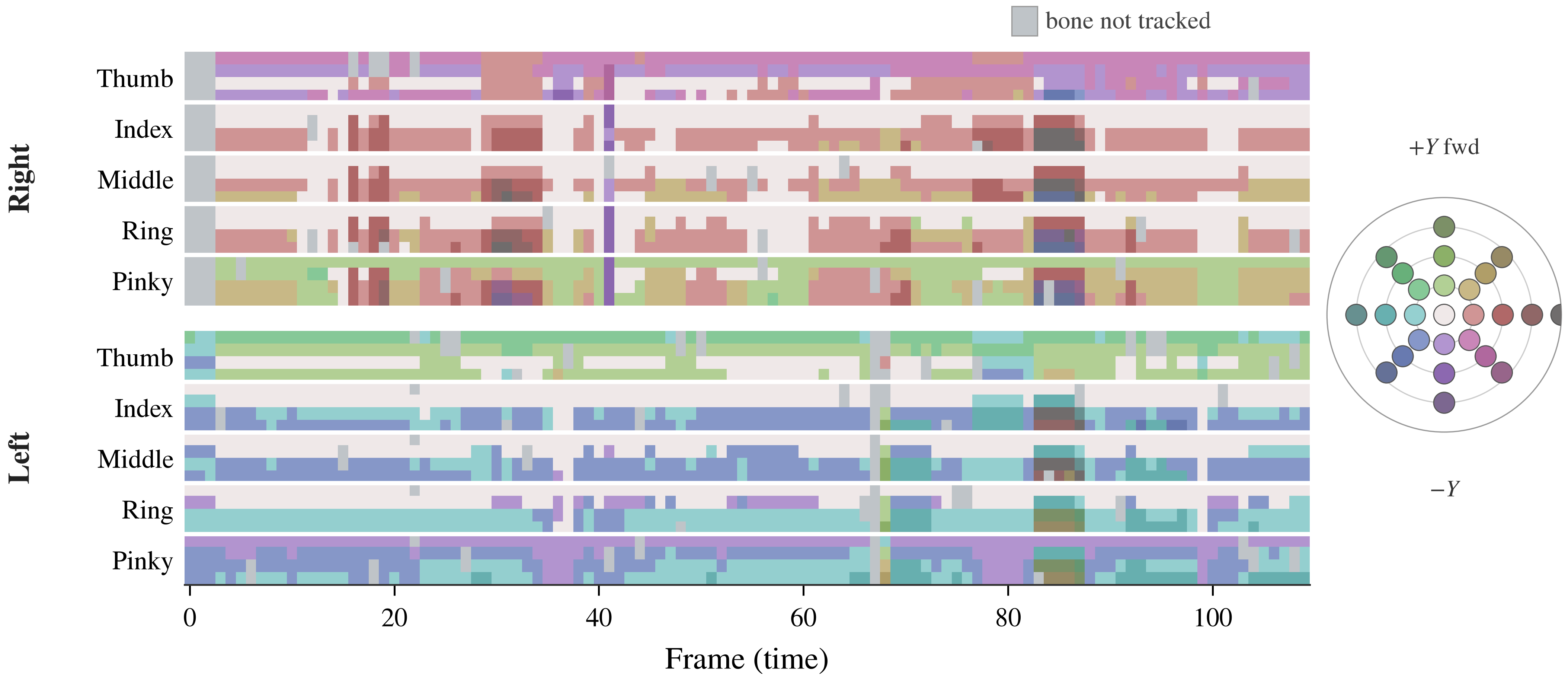}
\caption{The HL-26 token grid as a ``hand pianoroll'': $40$ bone-direction channels (rows; the two hands split by the white divider) over a $110$-frame window (columns). Colour encodes the \emph{direction} a token names rather than its index: lightness falls with the angle from $+Y$ (forward) and hue is the azimuth about that axis, so neighbouring directions receive neighbouring colours and a small pose change reads as a small colour shift. The pale anchor covering most of the grid is the $+Y$ concentration of Sec.~\ref{app:geom}. \emph{Key at right}: the $26$ codes placed at radius $=$ angle from $+Y$, azimuth $=$ azimuth. Long horizontal bands of constant colour---and the near-constant MCP (knuckle) rows---are the temporal and kinematic redundancy a time-aware code can compress.}
\label{fig:pianoroll}
\end{figure*}

\begin{figure*}[t]
\centering
\includegraphics[width=\linewidth]{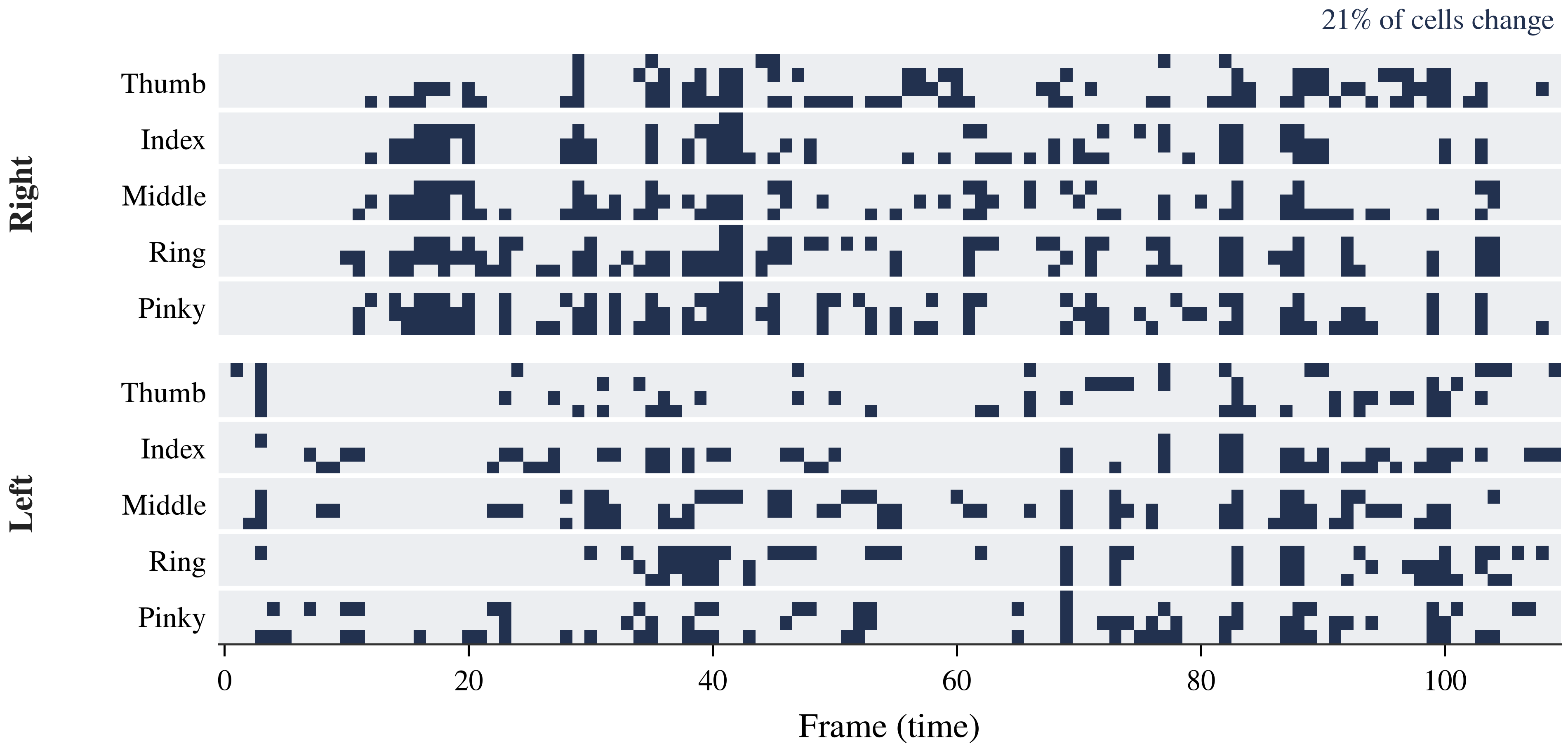}
\caption{Change map of the same window: black marks a bone whose token differs from the previous frame. Only ${\sim}21\%$ of HL-26 cells change (matching the $21.1\%$ dataset-wide rate; finer codes register more change---$34\%$ for DigitCode-A $K{=}64$, $44\%$ per-finger), and the MCP rows are essentially constant---the frame-to-frame redundancy quantified in this section.}
\label{fig:changemap}
\end{figure*}

\paragraph{The grid is visibly redundant.}
Figures~\ref{fig:pianoroll} and~\ref{fig:changemap} render a $110$-frame two-hand window as the $T\times 40$ HL-26 grid and its frame-to-frame change map, making the redundancy quantified below visible at a glance.

\paragraph{Relative encoding.}
Encoding directions relative to a reference concentrates the distribution and shifts the whole R--D curve left (Table~\ref{tab:rel}). The mean angular distance to the $+Y$ axis drops from $43.8^\circ$ (absolute) to $26.8^\circ$ (parent-relative) to $15.4^\circ$ (temporal-relative).

\begin{table}[t]
\centering
\small
\begin{tabular}{lcc}
\toprule
Scheme & entr.\ b. & Ang.\ ($^\circ$) \\
\midrule
Absolute k-means ($K{=}26$)  & $4.38$ & $9.08$ \\
Parent-relative (chain)      & $3.98$ & $6.02$ \\
Temporal-relative            & $2.70$ & $4.70$ \\
Per-finger concat $\to$ delta & $1.72$ & $4.60$ \\
\bottomrule
\end{tabular}
\caption{Relative encodings (held-out InterHand2.6M, entropy rate). Temporal-relative coding roughly halves the rate at lower error; per-finger deltas are the low-rate champion. The absolute $K{=}26$ baseline here ($9.08^\circ$) is from this relative-coding pipeline; the R--D tables report $8.45^\circ$ (frame-pooled) for the same setting.}
\label{tab:rel}
\end{table}

\paragraph{Closed-loop stability.}
Using reconstructed (rather than ground-truth) references at decode time, error growth is bounded: parent-relative chains degrade by ${\sim}{+}1^\circ$ at the chain end, and temporal-relative coding by ${+}1.6$--$2.1^\circ$. Both remain usable.

\paragraph{Temporal-relative coding subsumes compression.}
Two diagnostics show that decorrelation and compression are the same operation here. (i) Multi-step conditional entropy $H(t\mid t{-}k)$ decays slowly for absolute tokens (still low at $k{=}20$), but is essentially flat for $k>1$ under temporal-relative coding---its residual is close to i.i.d. (ii) Off-the-shelf gzip compresses HL-26 token streams to $1.57$ bits/vector, but cannot compress temporal-relative tokens at $3.37$ bits at all. Relatedly, HL-26 changes only $21\%$ of tokens frame-to-frame: a coarse code ``absorbs'' small motion, which helps retrieval and hurts fine forecasting.

\section{Temporal Event Compression}
\label{app:event}
The main paper compresses the frame count by keeping tokens only at keyframes; keeping $45$--$57\%$ of frames costs ${<}0.7^\circ$ over the all-frames floor on HanCo and about $1^\circ$ on the burstier InterHand data, and stacking event with temporal-relative coding reaches $1.81$ bits/direction, the product of the two reductions (a $4.38\times$ better rate$\times$quality index downstream). Table~\ref{tab:event} compares three keyframe rules under slerp decoding at matched keep-rate: uniform subsampling, a drift-threshold event rule, and curvature-aware Douglas--Peucker (RDP). RDP is best at every rate, by ${\sim}1^\circ$ on smooth HanCo and by $3$--$9^\circ$ on the burstier InterHand range-of-motion data.

\begin{table}[t]
\centering
\small
\begin{tabular}{lccc}
\toprule
keep\% & RDP (curv.) & event (drift) & uniform \\
\midrule
\multicolumn{4}{l}{\emph{HanCo} (floor $4.29^\circ$)}\\
$20$ & $\mathbf{7.63}$ & $9.97$ & $8.79$ \\
$30$ & $\mathbf{6.11}$ & $7.95$ & $7.18$ \\
$40$ & $\mathbf{5.24}$ & $6.61$ & $6.29$ \\
$50$ & $\mathbf{4.75}$ & $5.74$ & $5.59$ \\
\multicolumn{4}{l}{\emph{InterHand} (floor $5.58^\circ$)}\\
$20$ & $\mathbf{9.57}$ & $13.00$ & $18.66$ \\
$40$ & $\mathbf{8.19}$ & $10.96$ & $11.44$ \\
$50$ & $\mathbf{6.56}$ & $8.85$ & $9.74$ \\
\bottomrule
\end{tabular}
\caption{Temporal event compression (angular error, $^\circ$; slerp decoding). Curvature-aware keyframes (RDP) dominate at every keep-rate. Keeping ${\sim}45$--$57\%$ of frames is near-lossless.}
\label{tab:event}
\end{table}

\paragraph{Keyframe rule must match the decoder.}
A drift-threshold criterion bounds the zero-order-\emph{hold} error, so under hold decoding it beats uniform (e.g.\ at keep $9\%$ on HanCo, $12.66^\circ$ vs.\ $16.29^\circ$); but slerp decoding needs anchors at curvature extrema, which the drift rule does not provide, so it only ties uniform there. RDP places anchors exactly where slerp error is largest and therefore wins under slerp.

\paragraph{Downstream.}
Under the coding-efficiency protocol standard for event-based symbolic sequences, the event sequence is $2.10\times$ shorter (a $4.38\times$ better coding-efficiency index) at matched distribution fidelity: over $3$ seeds the collapse-sensitive AE-FGD ties uniform-rate modeling ($0.820{\pm}0.108$ vs.\ $0.820{\pm}0.103$). A fairness control decodes the uniform model's own generations through the identical keyframe$+$slerp pipeline; its speed-JS ($0.041{\pm}0.005$) matches the event model's ($0.053{\pm}0.004$), attributing the once-apparent $5\times$ JS gain to the smoothing decoder rather than the tokens. What the event tokens do preserve is speed itself: mean generated speed $7.16$ vs.\ the smoothed control's $6.06$ (real $7.46$), i.e.\ smooth-looking dynamics without the interpolation's systematic slowdown. The claim is efficiency at parity, not better dynamics.

\section{Decoding Schedule: Details}
\label{app:schedule}

The main paper's schedule claim rests on Table~\ref{tab:schedule}: under a matched $12$K-step budget and $3$ seeds on HanCo, the anatomy-grouped delay generates the best motion at every seed under the robust motion-AE latent FGD (protocol and controls under ``Matched-budget control'' below). A broader single-budget sweep under the legacy statistic-FGD (below) then maps the schedule space, from which we read relative ordering only.

\begin{table}[t]
\centering
\small
\begin{tabular}{llcc}
\toprule
Schedule & regime & steps/fr. & AE-FGD\,$\downarrow$ \\
\midrule
flat (full AR) & fine-grained & $20.0$ & $2.43\pm0.25$ \\
parallel & compound & $1.00$ & $2.20\pm0.07$ \\
delay (\textsc{pj-distal}) & delay & $1.08$ & $\mathbf{1.81\pm0.04}$ \\
\bottomrule
\end{tabular}
\caption{Matched-budget decoding schedules on DigitCode-A tokens (leak-free generation; motion-AE latent FGD; HanCo; mean$\pm$std over $3$ seeds). On short, static InterHand clips the ordering reverses.}
\label{tab:schedule}
\end{table}

\paragraph{Why an anatomy-grouped delay.}
The delay grouping is motivated by a conditional-entropy analysis of the DigitCode-A per-bone tokens ($K{=}64$). The intra-finger kinematic chain is strongly directional---parent$\to$child mutual information rises along the chain ($0.16$ for PIP$|$MCP, $0.43$ for DIP$|$PIP, $0.59$ for TIP$|$DIP)---and packing tokens by finger (within-group MI $0.39$) or by joint level ($0.33$) both beat a cross-group baseline ($0.29$). This motivates grouping the delay by anatomy rather than arbitrarily. A broader single-seed sweep (legacy $45$-D statistic FGD, cross-seed std $\pm283$--$424$) confirms only the coarse separation robust to that noise: every anatomy-grouped delay (${\sim}2000$--$2300$) beats parallel ($2572$), random within-frame order ($3261$), and the flat full-AR chain ($6650$), preserved across $K{\in}\{32,64,128\}$. The finer contrasts---chain direction and delay width---fall inside the seed noise, so we do not claim them: the chain direction is adopted on the kinematic-entropy argument above, and the minimal delay magnitude by \emph{cost} ($1.08$ vs.\ $1.6$/$2.1$ steps/frame), not by measured quality.

\paragraph{Why generation, not teacher-forced likelihood.}
A teacher-forced likelihood (bits per bone-token) is \emph{not} a valid metric for cross-frame delay schedules. A delayed bone of frame $t$ is decoded at a step that already conditions on early-delay bones of \emph{future} frames, and because hand motion is smooth those future ground-truth tokens deflate the loss---a delay schedule's bpt is leaked, not earned. The tell is that the zero-delay parallel config (no leak) loses; at the likelihood level the flat full-AR chain is therefore the causal ceiling (leak-free, frame-causal: flat $3.24$ $\approx$ a within-frame reorder $3.28$ $\approx$ random $3.25$, so a within-frame schedule yields no free likelihood). We therefore evaluate schedules only by leak-free generation (FGD), as in the main paper.

\paragraph{Matched-budget control.}
To rule out that flat merely needs more training, we retrain every schedule to the same $12$K-step budget, at $3$ training seeds, and score leak-free generation with the collapse-sensitive motion-AE latent FGD (Sec.~\ref{app:repro}). On HanCo ($n{=}119$ generated clips per configuration) the ordering of Table~\ref{tab:schedule} is identical at every seed. The delay is ${\sim}26\%$ better than a fully trained flat chain and ${\sim}18\%$ better than the equally short parallel schedule, so the gap is neither an under-training artifact nor mere sequence shortening; the two mechanisms separate cleanly as sequence length (the $20\times$-longer flat chain drifts, its generated speed overshooting $17$--$20^\circ$ vs.\ real $7.5$, which also destabilizes the legacy statistic FGD to a cross-seed std of $\pm1709$---why we do not quote the larger legacy ratios) and intra-frame dependency (parallel emits all bones at once and trails the delay). On InterHand2.6M the same matched protocol does \emph{not} reproduce the advantage: evaluation clips are short and largely static ($n{=}8$ per arm), the legacy metric shows no consistent per-seed direction, and the robust metric reverses the ordering (flat $5.72{\pm}0.31$ best, parallel $6.64{\pm}0.26$, delay $6.79{\pm}0.26$). We therefore scope the delay's advantage to long continuous motion, as stated in the main paper. The efficiency gain (${\sim}18\times$ fewer decoding steps) is structural and holds in both regimes.

\section{Full Quantizer Comparison}
\label{app:quant}

Table~\ref{tab:quantfull} expands the main-paper quantizer table with additional operating points and the high-fidelity regime. Every cell is matched by empirical \emph{entropy}, not fixed-width rate, so the main paper's headline per-bone point---HL-26's $14.71^\circ$ at $4.70$ fixed-width bits against DigitCode-A's $8.45^\circ$ at the same $4.70$---does not appear in it verbatim. The corresponding entropy-rate entry is the $K{=}26$ column pair ($8.43$ k-means, $8.53$ VQ, at $4.39$/$4.53$ b). The two are the same operating point under the two conventions, reconciled in Sec.~\ref{app:repro}. Quantizing raw $3$-D directions beats spherical $(\theta,\phi)$ coordinates for every method except PQ, whose per-axis grid fits the factorization; we use raw $3$-D throughout.

\begin{table}[t]
\centering
\small
\begin{tabular}{lcccc}
\toprule
Method & ${\sim}4.4$b & ${\sim}5.6$b & ${\sim}6.6$b & ${\sim}8$b \\
\midrule
k-means      & $8.43$  & $5.64$ & $4.07$ & $2.47$ \\
VQ (learned) & $8.53$  & $5.63$ & $4.08$ & $2.45$ \\
FSQ          & $9.88$  & $8.47$ & --     & -- \\
PQ           & $12.58$ & $9.11$ & $6.25$ & -- \\
RVQ          & $10.02$ & $6.24$ & $4.45$ & $2.80$ \\
BSQ          & $11.65$ & $9.45$ & --     & -- \\
\bottomrule
\end{tabular}
\caption{Full quantizer comparison in raw-3D space (angular error, $^\circ$, matched bits). A single k-means layer matches learned VQ at every rate, the high-fidelity regime included ($2.47$ vs.\ $2.45^\circ$ at ${\sim}8$ b), and both beat the two-stage RVQ there ($2.80^\circ$ at $8.03$ b); FSQ/PQ/BSQ are dominated at equal rate (BSQ reaches $10.33^\circ$ only with a $256$-code book). Rate is matched empirical entropy; the ${\sim}8$b cells sit at $7.99$, $8.30$ and $8.03$ bits respectively.}
\label{tab:quantfull}
\end{table}

\paragraph{The RVQ ladder.}
Residual VQ carries two knobs---per-stage $K$ and depth---so each of its cells above reports the best $(K,\text{stages})$ nearest to that column's rate: $(6,2)$ at $4.33$ b, $(5,3)$ at $5.82$ b, $(16,2)$ at $6.61$ b, $(32,2)$ at $8.03$ b. Depth is not free where bits are scarce: at ${\sim}5.0$ bits a three-stage code is \emph{worse} than a two-stage one ($8.34^\circ$ at $5.04$ b vs.\ $7.51^\circ$ at $5.06$ b), because each extra stage spends a code index before the first one has resolved the direction. This is why the gap to a single k-means layer widens monotonically as rate falls: $0.38$, $0.60$, $1.59^\circ$ at ${\sim}6.6$, ${\sim}5.6$, ${\sim}4.4$ bits.

\paragraph{The ordering does not reverse at the top.}
Extending the single-stage sweep upward closes the other end. Against RVQ's two published high-rate points, k-means gives $2.47^\circ$ at $7.99$ b (RVQ $2.80$ at $8.03$) and $1.08^\circ$ at $10.34$ b (RVQ $1.18$ at $10.75$), with further single-stage points at $1.52^\circ$/$9.36$ b, $0.89^\circ$/$10.88$ b and $0.77^\circ$/$11.28$ b. At every rate we measure, one k-means layer at the bone unit is at least as good as residual depth at the same unit. The main paper states this conservatively, conceding a crossover far above our operating points; at the bone unit there is none. Where depth does pay is on a per-\emph{finger} coarse stage---that is DigitCode-H, and it is the distinction the main paper draws.

\paragraph{What the spherical column is matched on.}
The spherical-coordinate column in the main paper's quantizer table is matched on nominal $K$, not on entropy. A total codebook of $64$ there ($(8,2)$, $5.41$ entropy b) gives $8.71^\circ$, against k-means' $6.59^\circ$ at $5.84$ b.

\paragraph{Why FSQ and BSQ have no high-rate cell.}
Both place cells uniformly, and bone directions are not uniform (Fig.~\ref{fig:density}), so the codes they add land mostly where there is no data and their \emph{achieved} entropy falls ever further below the nominal $\log_2 K$ they are charged. FSQ's per-axis grid runs $2.64$, $3.68$, $4.48$, $5.00$ entropy bits at $L{=}3$--$6$ ($K{=}27$--$216$), a shortfall widening from $2.11$ to $2.75$ bits; BSQ's binary spherical code runs $2.54$, $3.03$, $4.04$, $4.61$, $5.10$ bits at $K{=}8$--$1024$, a shortfall reaching $4.90$. Neither crosses the ${\sim}6.6$-bit column at any codebook size we test---and the shortfall is still \emph{widening} at the largest---so the dashes in Table~\ref{tab:quantfull} record a saturation, not an unrun configuration. This is the quantitative form of the main paper's ${\sim}2.8$ and ${\sim}4.9$ bit figures, and it is why the two families are excluded from the null we claim over \emph{strong} quantizers rather than counted against it.

\paragraph{External full-system baseline: a MoMask-style sequence RVQ-VAE.}
The main paper's quantizer sweep compares \emph{mechanisms}; here we align one full learned \emph{system} on the same grid and split. The model is the MoMask recipe scaled to our data: a temporal convolutional encoder over $W$-frame windows of one hand's $20{\times}3$ direction grid (hidden $256$, three residual blocks, optional stride-$2$ downsampling $d$), a residual-VQ bottleneck ($V$ levels $\times$ $K{=}512$ codes, EMA codebooks with dead-code revival---the same hygiene as the tuned VQ above), and a mirrored decoder with per-bone renormalization, trained for $12$K steps with the standard optimizer settings of the Reproducibility section. The temporal window gives this system context that our frame-local codes never see. Fixed-width rate is $V\log_2 K/(d\cdot 20)$ bits per bone; evaluation is non-overlapping held-out windows ($n{=}40$ at $W{=}32$, $n{=}216$ at $W{=}16$). Table~\ref{tab:extvqvae} reports the sweep with the unit-aligned codes as reference: the system's best configuration ($W{=}16$) draws level with the finger code at low rate ($5.83^\circ$ at $1.80$ bits, between DigitCode-F's $6.18^\circ$ at $1.75$ and $5.50^\circ$ at $2.0$) but stays ${\sim}2\times$ behind the hierarchical code at high rate ($3.90^\circ$ vs.\ $1.86^\circ$), and the genuinely MoMask-like configuration with $4\times$ temporal downsampling is the weakest point. The result is seed-stable ($W{=}32$, $3$ seeds: $7.78\pm0.05^\circ$ at $V{=}4$, $5.47\pm0.03^\circ$ at $V{=}14$). Training error falls below $1^\circ$ while held-out error does not: at this data volume the learned system overfits where the geometric codes, having no encoder to overfit, cannot---the system-level form of the mechanism finding.

We flag the fairness limit rather than leave it implicit. The overfitting gap is evidence about \emph{this} data regime ($717$K vectors, the largest public two-hand source available to us), not about learned tokenizers in general; more hand motion could close it. Two things keep the comparison honest meanwhile: the learned system gets a temporal window our frame-local codes never see, so any context advantage accrues to it; and it is trained with the same optimizer, budget, and codebook hygiene (EMA, dead-code revival) as the tuned VQ, which we report in \emph{tuned} form because an under-tuned VQ collapses and would flatter us. The licensed statement is scoped---at the data volume the hand field currently has, unit alignment buys more than learning does---not unconditional.

\begin{table}[t]
\centering
\small
\setlength{\tabcolsep}{4pt}
\begin{tabular}{lccc}
\toprule
Config ($d$, $V$, $W$) & fixed b. & entr.\ b. & Ang.\ ($^\circ$) \\
\midrule
$d{=}1$, $V{=}2$, $W{=}32$  & $0.90$ & $0.78$ & $10.68$ \\
$d{=}1$, $V{=}4$, $W{=}32$  & $1.80$ & $1.56$ & $7.74$ \\
$d{=}1$, $V{=}8$, $W{=}32$  & $3.60$ & $3.09$ & $6.38$ \\
$d{=}1$, $V{=}14$, $W{=}32$ & $6.30$ & $5.34$ & $5.50$ \\
$d{=}4$, $V{=}8$, $W{=}32$  & $0.90$ & $0.71$ & $10.92$ \\
$d{=}1$, $V{=}4$, $W{=}16$  & $1.80$ & $1.67$ & $5.83$ \\
$d{=}1$, $V{=}14$, $W{=}16$ & $6.30$ & $5.99$ & $3.90$ \\
\midrule
DigitCode-F ($K{=}128$)      & $1.75$ & --     & $6.18$ \\
DigitCode-F ($K{=}256$)      & $2.0$  & $1.94$ & $5.50$ \\
DigitCode-H ($K_1{=}128$, $K_2{=}32$) & $6.75$ & $6.12$ & $1.86$ \\
\bottomrule
\end{tabular}
\caption{MoMask-style sequence RVQ-VAE system aligned on held-out InterHand2.6M (upper block; single run, seed $0$; $V{=}4$/$V{=}14$ at $W{=}32$ are seed-stable at $\pm0.05^\circ$/$\pm0.03^\circ$ over $3$ seeds), with unit-aligned reference points below the rule. The main paper quotes the system's best window ($W{=}16$).}
\label{tab:extvqvae}
\end{table}

\paragraph{Geometric quantization reaches the learned bound.}
Table~\ref{tab:bound} runs a controlled sweep against a \emph{properly tuned} learned VQ (EMA codebook with dead-code revival) at three granularities, $K{=}64$. The two tie everywhere---within $0.2^\circ$ at every granularity, both at ${\approx}100\%$ utilization---because the direction space is low-dimensional (participation ratio $2.8$--$4.2$) and simple, leaving the learned encoder little to add over Lloyd's algorithm. We report the \emph{tuned} VQ deliberately: an under-tuned VQ (no dead-code revival) collapses and loses, which would make geometric quantization \emph{look} superior. The per-bone tie is \emph{seed-stable}: across $3$ VQ seeds, paired per-sequence deltas stay ${\le}0.15^\circ$ and flip sign at $K{=}64$/$128$ (k-means $5.64/4.07^\circ$ vs.\ VQ $5.60{\pm}0.04$/$4.08{\pm}0.03$), the evidence form a claimed match requires. The tie also survives all the way up the rate axis: at ${\sim}8$ entropy bits k-means and VQ differ by $0.02^\circ$ ($2.47$ vs.\ $2.45$), a fourth operating point at which the quantizer rule is null. Residual depth does not extend the per-bone frontier past a single layer at any rate we sweep (Table~\ref{tab:quantfull}); what extends it is changing the unit.

\begin{table}[t]
\centering
\small
\setlength{\tabcolsep}{4pt}
\begin{tabular}{lcccc}
\toprule
 & dim & fixed b. & VQ & k-means \\
\midrule
per-bone   & $3$  & $6.0$ & $5.61$  & $5.58$ \\
per-finger & $12$ & $1.5$ & $10.28$ & $10.25$ \\
per-hand   & $60$ & $0.3$ & $11.84$ & $11.74$ \\
\bottomrule
\end{tabular}
\caption{Geometric k-means vs.\ a tuned learned VQ (EMA $+$ dead-code revival), $K{=}64$, held-out InterHand2.6M (angular error $^\circ$; single run; the main paper's seed-averaged per-bone tie $5.64$ vs.\ $5.60{\pm}0.04$ is from the bake-off pipeline). The two tie at every granularity ($\approx\!100\%$ utilization). This sweep matches codebook \emph{size}, not rate---one token covers $1$/$4$/$20$ bones (b/bone $6.0/1.5/0.3$), so the finer unit wins only by spending $20\times$ the budget; the matched-rate claim is in Tables~\ref{tab:hier},~\ref{tab:effect}, not here. The per-finger $10.25^\circ$ is a \emph{pooled} single-codebook fit, $3.15^\circ$ looser than DigitCode-F's per-finger $7.10^\circ$ (Tables~\ref{tab:robustfull},~\ref{tab:grouping})---pooling discards the per-finger structure, itself the thesis; both arms here use it, so the tie holds but at this looser point.}
\label{tab:bound}
\end{table}

\paragraph{Effect decomposition: the quantizer \emph{rule} is null against its noise floor.}
A non-significant test would be weak evidence for ``the quantizer does not matter'' (absence of evidence is not evidence of absence). Table~\ref{tab:effect} instead calibrates each design knob against the VQ's own run-to-run noise, and---this is the point of the table---every controlled row holds the anatomical unit and the rate fixed while moving exactly one knob. We separate two things that are often conflated: the quantization \emph{rule} (argmax-cosine vs.\ k-means assignment vs.\ FSQ/PQ/BSQ), which is null at fixed unit and rate, and the codebook \emph{content} (a fixed cube vs.\ a data-fit set of codewords), which is not---DigitCode-A keeps HL-26's rule verbatim and changes only its $26$ codewords, worth $6.26^\circ$.

Read row by row: swapping a tuned VQ for training-free k-means moves reconstruction by at most $0.10^\circ$, either sign, the scale of the VQ's own $3$-seed std ($\pm0.04^\circ$); widening the rule to the weaker families (FSQ/PQ/BSQ) does move error, by $3.81^\circ$ at ${\sim}5.6$ bits, so the null we claim is over \emph{strong} rules, not all of them. The alphabet's \emph{content} matters at either end: replacing HL-26's cube with a data-fit codebook at the same $K{=}26$ and the same $4.70$ fixed-width bits is worth $6.26^\circ$, and even against the best fixed geometry we could build (Fibonacci $K{=}128$, $6.96^\circ$ at $5.47$ entropy bits) a data-fit per-bone code at matched rate is $1.32^\circ$ better. Depth at a \emph{fixed} bone unit does not pay: a two-stage per-bone RVQ ($4.45^\circ$ at ${\sim}6.6$ entropy bits) is $0.38^\circ$ \emph{worse} than one-stage k-means ($4.07^\circ$ at $6.55$), and the sign does not change anywhere on the rate axis we sweep, from $4.3$ to $10.9$ bits (Sec.~\ref{app:quant}). The same residual mechanism placed on a per-finger coarse stage instead reaches $1.86^\circ$ at $6.12$ entropy bits. Neither of those comparators isolates the unit, though---one differs in depth, the other in quantizer family---so we ran the contrast the claim actually needs (Table~\ref{tab:unitctrl}), and it gives a \emph{smaller} effect than either: $1.63^\circ$, with the per-bone arm given more rate, not less.

\paragraph{Absolute vs.\ relative scale.}
In absolute degrees the alphabet's content ($6.26^\circ$) exceeds the unit ($1.63^\circ$), but the two contrasts sit at different operating points---content is measured where error is still $14.71^\circ$, the unit where it is already $3.51^\circ$---and angular error floors at zero, so the same fractional gain buys fewer degrees the lower one starts. Relative to the error each removes the two are comparable ($46\%$ vs.\ $43\%$), but the unit is the only knob that improves error and rate together. We report both scales and state the headline claim on the relative one. So both the alphabet and the unit move error at matched rate, and the null is over the quantizer rule, not over code design.

\begin{table}[t]
\centering
\small
\setlength{\tabcolsep}{3pt}
\begin{tabular}{@{}>{\raggedright\arraybackslash}p{0.35\linewidth}>{\raggedright\arraybackslash}p{0.23\linewidth}rr@{}}
\toprule
Knob (what changes) & held fixed & $|\Delta|$ ($^\circ$) & rel.\ \% \\
\midrule
Quantizer \emph{rule}: tuned VQ $\leftrightarrow$ k-means & unit, rate & $\le 0.10$ & $\le 1.7$ \\
\;\emph{VQ run-to-run} ($3$ seeds) & unit, rate, rule & $0.04$ & $0.7$ \\
Quantizer \emph{family}, incl.\ FSQ/PQ/BSQ & unit, rate (${\sim}5.6$b) & $3.81$ & $40$ \\
Alphabet: best fixed geometry $\to$ data-fit & unit (bone), rate (${\sim}5.5$b) & $1.32$ & $19$ \\
Alphabet \emph{content}: fixed cube $\to$ data-fit, $K{=}26$ & unit, $4.70$b, $K$ & $6.26$ & $43$ \\
Depth at fixed bone unit: $1$-stage $\to$ $2$-stage RVQ & unit, rate (${\sim}6.6$b) & $0.38^{\ddagger}$ & $9^{\ddagger}$ \\
Unit of the coarse stage: bone $\to$ finger & depth, mechanism, rate & $1.63$ & $\mathbf{46}$ \\
\midrule
\multicolumn{2}{@{}>{\raggedright\arraybackslash}p{0.60\linewidth}}{\emph{End-to-end total, HL-26 $\to$ DigitCode-H (all knobs; not a decomposition term)}} & \emph{$12.85$} & \emph{$87$} \\
\bottomrule
\end{tabular}
\caption{Effect decomposition on held-out InterHand2.6M (angular error). Every row above the second rule moves exactly one thing, with everything in ``held fixed'' matched; the unit row necessarily changes the anatomical unit and gives the per-bone arm \emph{more} rate rather than an exactly matched one (sources: Tables~\ref{tab:quantfull},~\ref{tab:geom},~\ref{tab:hier},~\ref{tab:signif}). ``rel.\ \%'' is $|\Delta|$ as a fraction of the \emph{higher-error} arm of that same contrast---the comparable scale, since the contrasts sit at different operating points. The quantizer \emph{rule} is null against seed noise ($\pm0.04^\circ$); alphabet \emph{content} and \emph{unit} both move error by degrees, and the two are comparable on the relative scale ($46$ vs.\ $43\%$). The unit row uses the strictest available control (Table~\ref{tab:unitctrl}): the \emph{same} two-stage geometric code with only the coarse grouping changed, and with the per-bone arm given \emph{more} rate. $^{\ddagger}$Depth at a fixed bone unit goes the \emph{wrong} way ($0.38^\circ$/$9\%$ worse). The final row is the end-to-end total (alphabet $+$ unit $+$ depth $+$ rate $4.70\to6.75$ b), \emph{not} a decomposition term.}
\label{tab:effect}
\end{table}

\begin{figure}[h]
\centering
\includegraphics[width=\linewidth]{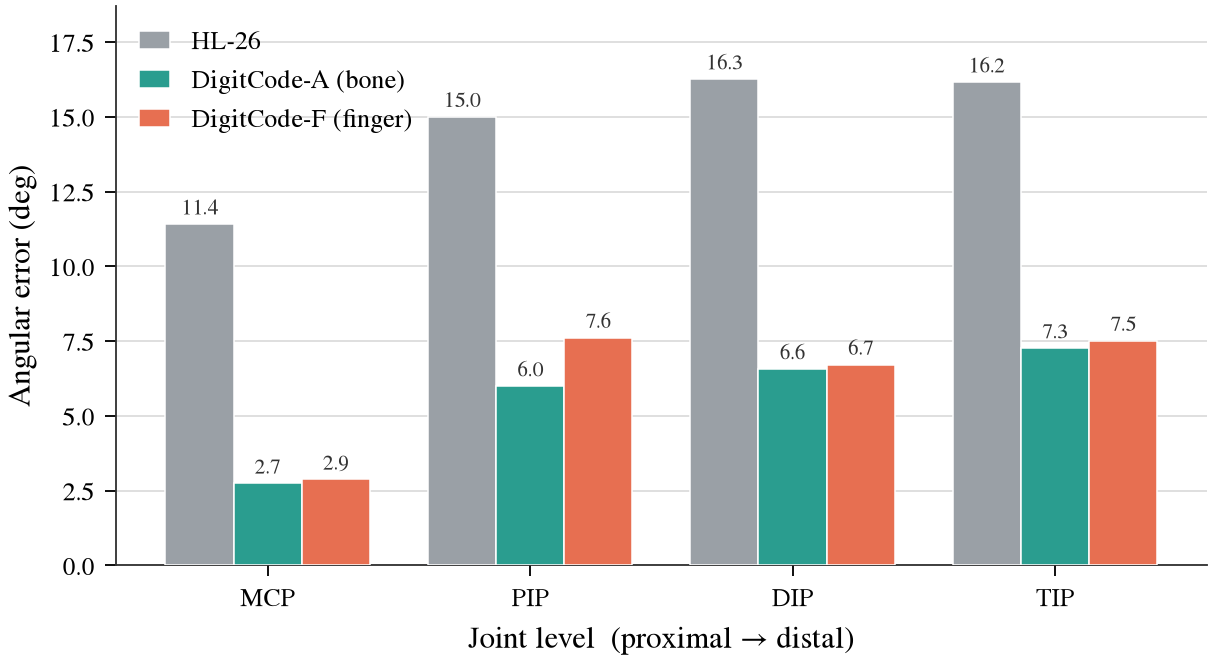}
\caption{Per-bone reconstruction error. DigitCode dominates HL-26 at \emph{every} joint rather than improving an average; the MCP (knuckle) row improves by $75\%$ ($11.41\to2.87^\circ$).}
\label{fig:perbone}
\end{figure}

\paragraph{Per-bone error breakdown.}
Under the adaptive code HL-26 is dominated at every joint rather than on average (Fig.~\ref{fig:perbone}).

\section{Full Downstream Tables}
\label{app:down}

\noindent The main paper's downstream table has six rows and this section carries four of them---forecasting, retrieval, denoising and generation---in full. The remaining two are protocolled where their machinery is defined rather than duplicated here: \emph{single-finger editing} (the exact-zero off-finger column, and the bimanual controls behind it) in Sec.~\ref{app:bimanual}, and \emph{gesture classification} (the $46.8\%$ single-whole-hand-token point, the $49.7\%$ combination, the subject-disjoint split and the feature-dimension control) in Sec.~\ref{app:toplevel}. Retrieval's composed-predicate arm is under ``Symbolic predicate queries'' below.

\begin{figure*}[t]
\centering
\includegraphics[width=0.98\linewidth]{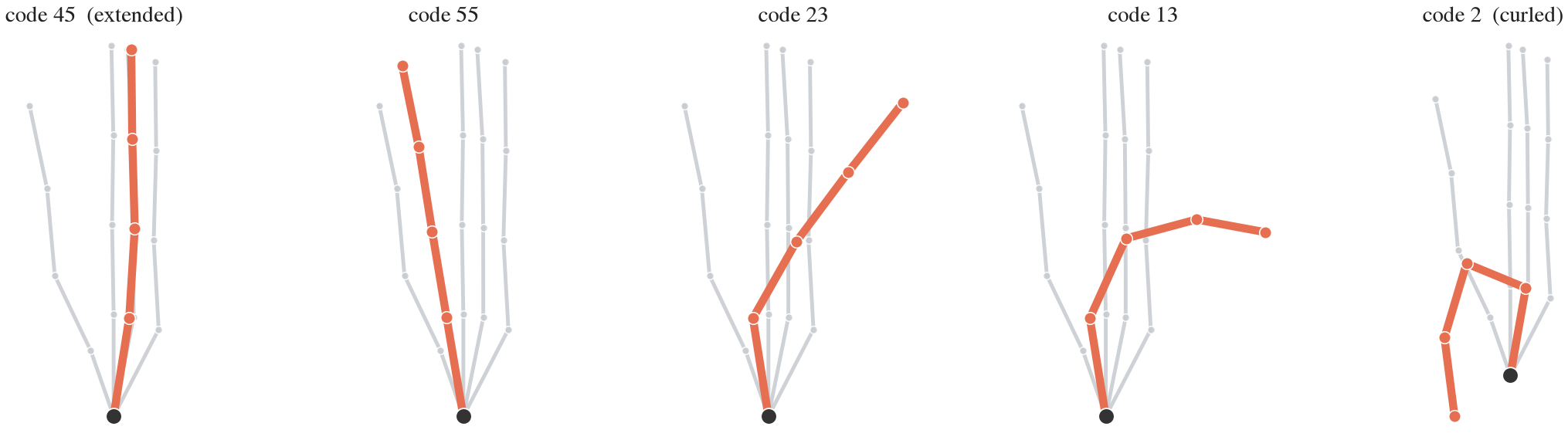}
\caption{DigitCode-F per-finger codebook (index finger): five codes spanning the curl range, each decoded to a whole-finger 3D pose (the varying finger in orange, the rest of the hand in grey). Because a code quantizes a finger jointly, every token is a readable, anatomically plausible configuration---from extended to curled.}
\label{fig:codebook}
\end{figure*}

As in the main paper, DigitCode-A denotes the per-bone data-adaptive code (k-means, $K{=}64$), DigitCode-F the per-finger joint code ($K{=}64$), and DigitCode-H the hierarchical coarse+residual code. Figure~\ref{fig:codebook} decodes five DigitCode-F codes to skeletons, the readability the Method section promises.

\paragraph{Motion forecasting across models.}
Table~\ref{tab:forecast} reports next-pose forecasting for four sequence models plus a \emph{token-copy} baseline. DigitCode-A is the forecasting champion ($-31\%$ vs.\ HL-26); autoregressive modelling beats the token-copy column for it and for HL-26, but the best arm merely matches a tokenizer-independent continuous-pose copy ($12.90$ vs.\ $12.74^\circ$): hand-motion direction is intrinsically hard to predict, and no tokenizer here beats ``predict no motion.''

\begin{table}[t]
\centering
\small
\setlength{\tabcolsep}{4pt}
\begin{tabular}{lccccc}
\toprule
Tokenizer & LLaMA & GRU & LSTM & Trans.\ & copy \\
\midrule
HL-26        & $18.85$ & $28.19$ & $32.99$ & $33.05$ & $22.72$ \\
DigitCode-A   & $\mathbf{12.90}$ & $26.31$ & $30.34$ & $28.21$ & $15.67$ \\
DigitCode-F & $19.04$ & $32.90$ & $38.99$ & $27.49$ & $15.77$ \\
\bottomrule
\end{tabular}
\caption{Forecasting (angular error, $^\circ$, $\downarrow$). Only autoregressive modeling beats the token-copy column; DigitCode-A is the forecasting champion ($-31\%$ vs.\ HL-26) but only ties the tokenizer-independent continuous-pose copy ($12.74^\circ$).}
\label{tab:forecast}
\end{table}

\paragraph{Retrieval.}
Table~\ref{tab:retr} reports training-free sequence retrieval (precision@1, random ${\sim}1.1\%$). Token sequences support exact symbolic matching (bag-of-tokens, $n$-gram) and discrete token-DTW directly, with no metric or threshold to calibrate, unlike metric search over continuous floats.

\begin{table}[t]
\centering
\small
\begin{tabular}{lccc}
\toprule
Tokenizer & bag & $n$-gram & DTW \\
\midrule
HL-26         & $17.0$ & $\mathbf{23.0}$ & $\mathbf{31.2}$ \\
DigitCode-A    & $19.2$ & $19.4$ & $26.6$ \\
DigitCode-F & $\mathbf{25.4}$ & $22.0$ & $28.6$ \\
DigitCode-H & $24.8$ & $17.4$ & $27.0$ \\
\bottomrule
\end{tabular}
\caption{Retrieval precision@1 (\%, $\uparrow$; bold marks the best per matching scheme). Per-finger bag-of-tokens is strongest; the spread is modest (HL-26 is best under $n$-gram and DTW), so retrieval is a usability result rather than a strong discriminator.}
\label{tab:retr}
\end{table}

\paragraph{Symbolic predicate queries.}
The symbols also support \emph{compositional} zero-shot queries. We define five single-hand attributes whose ground truth is a thresholded geometric rule on the continuous pose---that rule is the oracle by construction ($\mathrm{F1}=1.0$)---and compare symbol-set lookup (zero training; enumerate the matching codewords once, then $O(1)$ membership) against a per-attribute logistic detector trained on $14{,}400$ labeled frames (Table~\ref{tab:predq}). On the data-adaptive code, symbol lookup matches or beats the trained detector on \emph{every} attribute at zero training cost, and the gap is largest exactly where HL-26's coarse alphabet is weakest (pinky-up: $0.896$ vs.\ HL-26's $0.641$). On the composed query index-up $\wedge$ thumb-side $\wedge$ $\neg$fist the symbolic query degrades sub-additively ($0.825$) while the trained detectors compound errors ($0.747$): composition is native to symbols (set intersection) and auditable (the matching codewords can be read), the evidence behind the main paper's composed-predicate claim.

\begin{table}[t]
\centering
\small
\begin{tabular}{lccc}
\toprule
Attribute (pos.\ rate) & HL-26 & DigitCode-A & trained \\
\midrule
index-up ($0.32$)    & $0.929$ & $0.924$ & $0.802$ \\
thumb-side ($0.31$)  & $0.838$ & $\mathbf{0.890}$ & $0.866$ \\
pinky-up ($0.24$)    & $0.641$ & $\mathbf{0.896}$ & $0.800$ \\
index-curl ($0.29$)  & $0.844$ & $\mathbf{0.935}$ & $0.919$ \\
fist ($0.25$)        & $0.910$ & $\mathbf{0.978}$ & $0.938$ \\
\midrule
\emph{composite} ($0.11$) & $0.825$ & --- & $0.747$ \\
\bottomrule
\end{tabular}
\caption{Zero-shot symbolic predicate queries (F1$\uparrow$; held-out InterHand2.6M). Symbol-set lookup (HL-26 and DigitCode-A) uses \emph{zero} training and $O(K)$ enumeration vs.\ a logistic detector trained on $14{,}400$ frames. Each attribute's defining geometric rule is the oracle ($\mathrm{F1}{=}1.0$, omitted); bold marks where DigitCode-A beats the trained detector. Composite $=$ index-up $\wedge$ thumb-side $\wedge$ $\neg$fist.}
\label{tab:predq}
\end{table}

\paragraph{Denoising.}
Table~\ref{tab:denoise} reports denoising gain (noisy error minus reconstruction error; positive means the tokenizer cleaned the input) under angular noise $\sigma$. Per-finger joint codes turn net-positive once noise exceeds codebook resolution.

\begin{table}[t]
\centering
\small
\begin{tabular}{lcccc}
\toprule
$\sigma$ & noisy & HL-26 & DigitCode-A & DigitCode-F \\
\midrule
$5^\circ$  & $3.98$  & $-11.2$ & $-2.9$ & $-3.4$ \\
$10^\circ$ & $7.98$  & $-8.7$  & $-1.9$ & $-0.2$ \\
$15^\circ$ & $11.96$ & $-7.2$  & $-1.3$ & $+2.8$ \\
$20^\circ$ & $15.98$ & $-6.1$  & $-0.9$ & $+5.5$ \\
\bottomrule
\end{tabular}
\caption{Denoising gain ($^\circ$, $\uparrow$). At $\sigma{=}20^\circ$ the per-finger reconstruction ($10.51^\circ$) is far cleaner than independent per-bone k-means ($16.87^\circ$) despite a higher clean floor ($7.10$ vs.\ $5.64^\circ$), because joint quantization snaps four bones to one empirical whole-finger prototype. HL-26 is net-negative everywhere.}
\label{tab:denoise}
\end{table}

\paragraph{One-shot cleanup vs.\ alternatives.}
Table~\ref{tab:denoisemano} compares absolute error after cleanup on FreiHAND ($n{=}800$) against a nearest-neighbor projection onto the clean training set and an iterative MANO refit, at ${\sim}280\times$ and ${\sim}440\times$ lower cost respectively.

\begin{table}[t]
\centering
\small
\setlength{\tabcolsep}{4pt}
\begin{tabular}{lcccc}
\toprule
$\sigma$ & noisy & DigitCode-F & nn-clean & MANO re-fit \\
\midrule
$5^\circ$  & $4.01$  & $5.18$ & $\mathbf{2.32}$ & $5.91$ \\
$10^\circ$ & $7.99$  & $6.23$ & $\mathbf{5.08}$ & $7.07$ \\
$15^\circ$ & $11.95$ & $7.42$ & $\mathbf{7.26}$ & $8.51$ \\
$20^\circ$ & $16.08$ & $\mathbf{8.90}$ & $9.42$ & $10.46$ \\
\bottomrule
\end{tabular}
\caption{Absolute error after one-shot cleanup ($^\circ$, $\downarrow$; FreiHAND, $n{=}800$). DigitCode-F beats the iterative MANO re-fit at every $\sigma$ and passes nearest-neighbor at high noise, at a fraction of the cost ($0.018$ vs.\ $4.97$ vs.\ $7.96$\,ms/frame).}
\label{tab:denoisemano}
\end{table}

\paragraph{Distribution-level generation.}
Table~\ref{tab:gen} reports the collapse-sensitive motion-AE latent FGD and the speed-distribution Jensen--Shannon divergence under autoregressive (AR) and masked decoding, at $3$ training seeds. Adaptive tokens generate realistic dynamics; the gap to HL-26 persists under the stronger masked decoder, since the small-motion information is removed at quantization.

\begin{table}[t]
\centering
\small
\setlength{\tabcolsep}{3pt}
\begin{tabular}{lcccc}
\toprule
Tokenizer & AR AE-FGD & AR JS & mask AE-FGD & mask JS \\
\midrule
HL-26       & $4.46{\pm}0.38$ & $0.340$ & $6.75{\pm}0.33$ & $0.366$ \\
DigitCode-A & $\mathbf{3.01{\pm}0.12}$ & $0.152$ & $\mathbf{5.54{\pm}0.32}$ & $0.151$ \\
DigitCode-F & $3.77{\pm}0.23$ & $\mathbf{0.075}$ & $5.85{\pm}0.13$ & $\mathbf{0.083}$ \\
\bottomrule
\end{tabular}
\caption{Distribution-level generation ($\downarrow$; mean$\pm$std over $3$ seeds; AR $n{=}16$ clips/seed, masked $n{=}9$; JS std $\le0.003$ everywhere). Masked AE-FGD is globally higher than AR (different window protocols, fixed $60$-frame windows), so compare within a column, not across. The two metrics order DigitCode-A and -F differently; see below.}
\label{tab:gen}
\end{table}

\paragraph{Why the two metrics order A and F differently.}
Both metrics put HL-26 worst under both decoders, but they split on the two DigitCode variants: AE-FGD prefers per-bone ($3.01$ vs.\ $3.77$), speed-JS prefers per-finger ($0.075$ vs.\ $0.152$). This is not noise (JS std $\le0.003$; the AE-FGD gap is ${\approx}3$ pooled stds) but a difference in what is measured. Speed-JS is a \emph{marginal} over one scalar; AE-FGD is a \emph{joint} distance in a latent encoding pose and dynamics together. At $K{=}64$ the per-finger code registers more frame-to-frame change ($44\%$ vs.\ $34\%$; Fig.~\ref{fig:changemap}), so it resolves small velocities better, while its coarser pose resolution ($7.10^\circ$ vs.\ $5.68^\circ$; Table~\ref{tab:robustfull}) puts its frames further from the real manifold---the two pull opposite ways. We read the ordering off AE-FGD, since a speed marginal cannot detect right speeds with wrong poses (the collapse-sensitivity that retired the statistic-FGD; Metrics), and state the qualification: on the speed marginal alone the finger code generates better. This study also has the smallest $n$ ($16$ AR clips/seed, $9$ masked), where a Fr\'echet covariance is materially biased---which is why only the HL-26 gap, large and sign-consistent at every seed under two decoders, is drawn from it without qualification.

\section{Whole-Hand (Top-Level) Tokens}
\label{app:toplevel}
Beyond the direction codes, we probe two orthogonal token types: a whole-hand shape token and a cross-finger relational token.

\paragraph{Whole-hand shape token.}
A single $k$-means code over the full $60$-D hand (20 bones $\times$ 3) gives one shape token per frame---the most compact, top-level code. On subject-disjoint $137$-class gesture classification (Table~\ref{tab:wholehand}), combining a $128$-entry whole-hand code with DigitCode-F lifts top-1 to $49.7\%$ ($+6.7$pp over HL-26; Table~\ref{tab:signif}) at $1.80$ bits/vector, $55\%$ of HL-26's rate, and DigitCode-F alone matches HL-26 ($43.0\%$) at $1.46$ bits. A \emph{single} whole-hand token is the most rate-efficient identity code by a wide margin: at $K{=}256$ it reaches $46.8\%$ from one token per frame ($0.38$ bits/vector), above the $20$-token per-bone code's $46.2\%$ at $14.6\times$ the rate and above HL-26's $43.0\%$ at $8.6\times$.

Sweeping the whole-hand codebook gives an inverted U---$35.4$, $42.6$, $\mathbf{46.8}$, $44.9$, $37.5$, $27.4\%$ at $K=64$ through $2048$---but a control shows the falling half is \emph{not} the identity task rejecting detail, and we report it rather than read the curve the flattering way. Enlarging $K$ changes three things at once: code resolution, one-hot feature dimensionality, and samples per code. Holding the codebook and the probe fixed and replacing the one-hot feature with the assigned codeword itself (dimensionality fixed at $60$ instead of growing to $2048$) removes the collapse entirely: accuracy then trends \emph{upward} with $K$, $26.7$, $33.9$, $33.2$, $35.8$, $35.5$, $36.5\%$, ending highest where the one-hot feature ended lowest. Test-split codebook occupancy stays at $0.85$--$1.00$ throughout, so dead codes are not the cause either. The high-$K$ collapse is therefore a probe-capacity artifact of the one-hot feature. All arms share one protocol and the subject-disjoint split; the $K\ge512$ points use a GPU k-means with identical semantics, verified to reproduce the $K=64$/$128$/$256$ values exactly.

\paragraph{What the sweep licenses.}
Only the claim the peak supports: at its best operating point a single whole-hand token matches a $20$-token per-bone code at a fraction of the rate. Combining the peak-region code with the finger code is better still ($49.7\%$), so the two are complementary rather than redundant. On noisy ASL handshape classification, however, the whole-hand code \emph{fails} ($10$--$15\%$ vs.\ HL-26's $31.4\%$): a single code over the whole hand is even more noise-sensitive than per-finger, consistent with the data-quality story (Section~\ref{app:regime}).

\begin{table}[t]
\centering
\small
\begin{tabular}{lcc}
\toprule
Tokenizer & entr.\ b. & top-1 (\%) \\
\midrule
HL-26                    & $3.27$ & $43.0$ \\
DigitCode-A ($K{=}26$)   & $4.40$ & $43.0$ \\
DigitCode-A ($K{=}64$)   & $5.55$ & $46.2$ \\
DigitCode-F ($K{=}64$)   & $1.46$ & $43.0$ \\
DigitCode-F ($K{=}128$)  & $1.73$ & $44.6$ \\
whole-hand ($K{=}64$)    & $\mathbf{0.28}$ & $35.4$ \\
whole-hand ($K{=}128$)   & $0.33$ & $42.6$ \\
whole-hand ($K{=}256$)   & $0.38$ & $46.8$ \\
whole-hand ($K{=}512$)   & $0.43$ & $44.9$ \\
whole-hand ($K{=}1024$)  & $0.47$ & $37.5$ \\
whole-hand ($K{=}2048$)  & $0.51$ & $27.4$ \\
DigitCode-F\,+\,whole-hand & $1.80$ & $\mathbf{49.7}$ \\
\bottomrule
\end{tabular}
\caption{Gesture classification (InterHand2.6M, $137$ classes, \emph{subject-disjoint}: official subject annotations, $14$ train / $13$ test subjects, zero overlap; sequence-level majority $1.0\%$). DigitCode-F matches HL-26 at $45\%$ of its rate; the whole-hand combination is best. Rate is empirical entropy bits per bone vector.}
\label{tab:wholehand}
\end{table}

\paragraph{Tokens as classification features (head-to-head).}
Table~\ref{tab:feat} fixes the classifier (pooled mean$+$std features, linear probe) and varies only the representation on the same $137$-class task, so rows are directly comparable. Discrete tokens are the strongest features---a compact per-finger code is best ($43.3\%$ at $1.7$ bits/bone)---and the learned-VQ tie extends downstream: VQ-VAE tokens ($42.0\%$) do not beat DigitCode-A ($42.4\%$). Pooled MANO parameters are competitive ($42.1\%$), so the margin over the best continuous input is modest; the point is that interpretable geometric tokens give up nothing to learned latents. Absolute numbers differ slightly from Table~\ref{tab:wholehand}, whose bag-of-token protocol (and majority baseline) differs; within this table the majority rate is $13.4\%$.

\begin{table}[t]
\centering
\small
\begin{tabular}{llc}
\toprule
Representation & type & top-1 (\%) \\
\midrule
DigitCode-F ($K{=}128$) & token & $\mathbf{43.3}$ \\
DigitCode-A ($K{=}64$) & token & $42.4$ \\
MANO-param (mean$+$std) & continuous & $42.1$ \\
VQ-VAE ($K{=}64$) & token & $42.0$ \\
HL-26 & token & $39.0$ \\
joint-angle (mean$+$std) & continuous & $37.5$ \\
raw-direction (mean$+$std) & continuous & $35.8$ \\
\bottomrule
\end{tabular}
\caption{Representations as classification features under one fixed protocol (linear probe on pooled features; $137$ classes; majority $13.4\%$). Tokens lead; VQ-VAE does not beat geometric DigitCode-A downstream either.}
\label{tab:feat}
\end{table}

\paragraph{Relational token (interpretability).}
Reducing each finger to a three-level curl state (extended/bent/curled) yields a five-symbol relational code per frame ($15$-D, human-readable). It carries real interpretable structure: its normalized mutual information with ASL handshape is $0.301$, and the dominant relational code of each handshape reads off the linguistic shape almost exactly---\texttt{ext/ext/ext/ext/ext} for an open ``5'' hand, \texttt{bent/ext/curl/curl/curl} for a pointing ``1,'' \texttt{bent/ext/ext/curl/curl} for ``V,'' and \texttt{bent/curl/curl/curl/curl} for a closed fist. The relational token's value is interpretability, not raw accuracy (its three levels are coarse), but it answers ``what can a token be read as?'' more directly than any direction code.

\section{Token Editing and Bimanual Preservation}
\label{app:bimanual}

\begin{figure}[h]
\centering
\includegraphics[width=\linewidth]{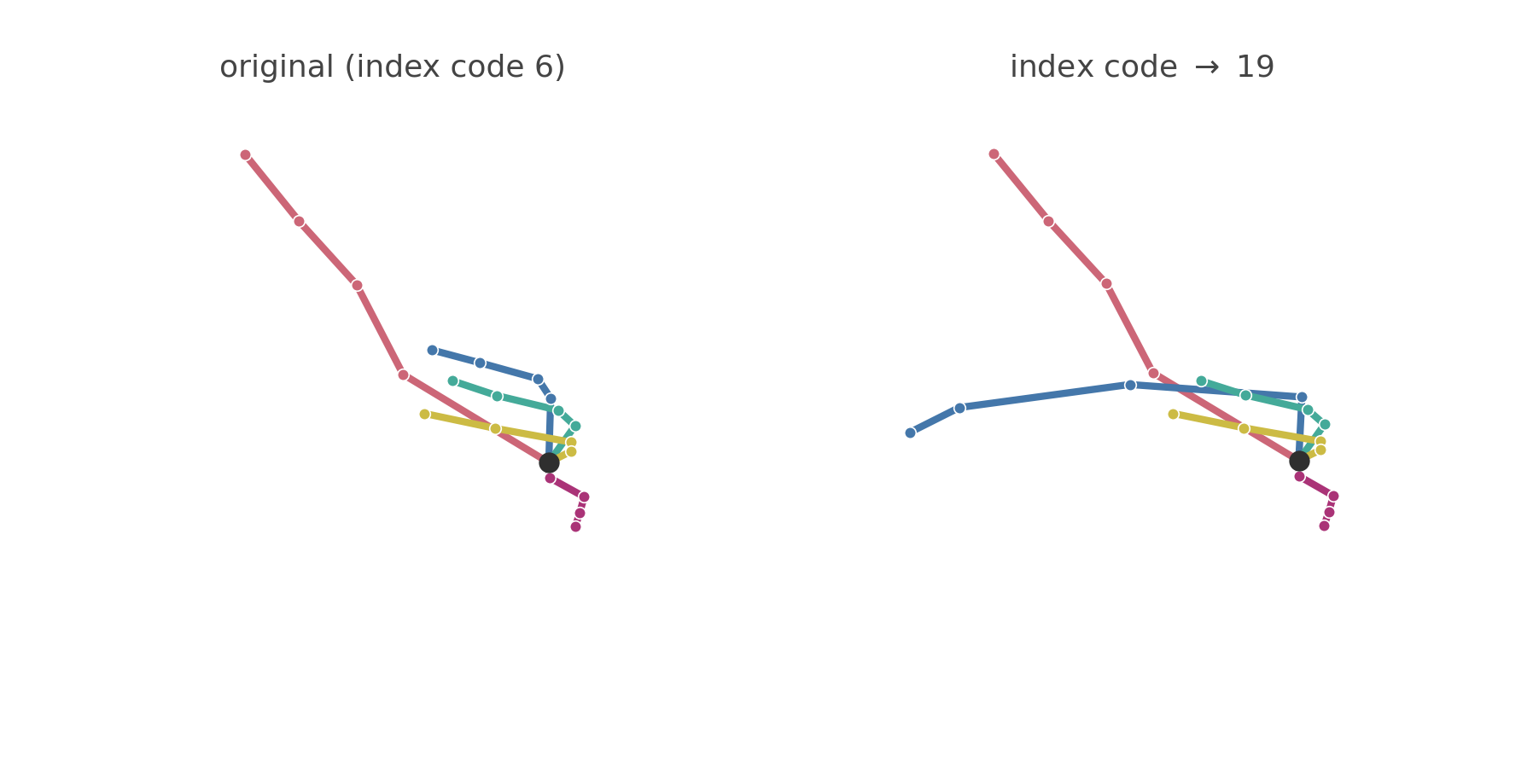}
\caption{Token editing (single hand). Swapping one finger's code rewrites \emph{only} that finger: the edited finger moves $98.3^\circ$ on average, every other finger exactly $0.00^\circ$. In a matched-magnitude MANO head-to-head ($t{=}2$ edits, $400$ clips), an unconstrained raw-MANO edit leaves the flexion envelope on $0.895$ of edits against $0.00017$ for the code swap; over all five edit magnitudes the code swap stays below $0.0005$. The envelope is the $[0.5,99.5]$ percentile range of each finger's three inter-bone flexion angles on the fit split, so this measures ``inside the range real hands occupy,'' not a biomechanical joint-limit model.}
\label{fig:edit}
\end{figure}
On InterHand2.6M two-hand interaction we test a bimanual edit: steer the \emph{right} hand toward a target pose by a symbolic token edit, then measure right-hand fidelity and left-hand disturbance, with $\mathrm{joint}=\mathrm{right\_match}\times\mathrm{left\_preserve}$ ($\tau_{\text{right}}{=}20^\circ$, $\varepsilon_{\text{left}}{=}5^\circ$, sequence-level mean over $18$ captures). We tabulate the \emph{feasible} slice (both hands valid); the \emph{full} slice (right hand valid only) gives the same ordering. Because a factored per-hand code rewrites \emph{only} right-hand tokens, the left hand is byte-for-byte untouched, so its drift is provably $0$ for all factored codes---extending the single-hand locality of Fig.~\ref{fig:edit} to the bimanual case as a structural guarantee, not a learned behavior. Among factored codes the joint score is then driven by right-hand fidelity (a rate--distortion gap): DigitCode-H reproduces the target right hand to $2.2^\circ$ and per-finger to $7.3^\circ$, versus $14.7^\circ$ for the coarse HL-26, so DigitCode-H attains the best joint score (Table~\ref{tab:bimanual}). As a foil we add a non-factored joint code over both hands ($256$ entries on the $40$-bone pose): steering its right half drags the left half by $30.9^\circ$ and collapses preservation to $0.25$. The lesson is to factor the hands---which DigitCode does by construction---rather than that one factored granularity preserves better than another.

\begin{table}[t]
\centering
\footnotesize
\setlength{\tabcolsep}{4pt}
\begin{tabular}{@{}lcccc@{}}
\toprule
Scheme (feasible slice) & r\,err$^\circ$ & l\,drift$^\circ$ & l\_pres & joint \\
\midrule
HL-26 (per-bone)     & $14.71$ & $0.000$ & $1.000$ & $0.788$ \\
DigitCode-F          & $7.30$  & $0.000$ & $1.000$ & $0.966$ \\
DigitCode-H          & $\mathbf{2.18}$ & $0.000$ & $1.000$ & $\mathbf{0.998}$ \\
joint-2-hands (foil) & $10.74$ & $30.92$ & $0.250$ & $0.216$ \\
\bottomrule
\end{tabular}
\caption{Bimanual preservation. Factored per-hand codes leave the non-edited (left) hand exactly fixed (drift $0$); the joint score then follows right-hand fidelity, so DigitCode-H is best. A non-factored both-hands code (foil) leaks the edit into the left hand.}
\label{tab:bimanual}
\end{table}

\section{Cross-Dataset Transfer}
\label{app:cross}

\paragraph{Cross-dataset transfer.}
Table~\ref{tab:crossfull} shows that a per-bone adaptive codebook transfers in both directions and dominates HL-26 ($-61\%$ on FreiHAND). The per-finger \emph{joint} codebook is more data-dependent: a FreiHAND-fit joint code degrades to $22.5^\circ$ on InterHand, because joint codes need diverse training poses. We report this trade-off rather than hide it.

\begin{table}[t]
\centering
\small
\begin{tabular}{lccc}
\toprule
Eval set & HL-26 & DigitCode-A (transfer) & within \\
\midrule
FreiHAND  & $14.43$ & $5.69$ (fit-IH) & $4.26$ \\
InterHand & $14.71$ & $9.96$ (fit-FH) & $5.61$ \\
\bottomrule
\end{tabular}
\caption{Cross-dataset transfer (angular error, $^\circ$). ``within'' fits and tests on the same dataset.}
\label{tab:crossfull}
\end{table}

\paragraph{The other two regimes.}
The main paper reads its transfer claim across all four capture regimes, so the HanCo and lifted-ASL half belongs here with the rest of the transfer evidence (the regimes themselves are set up in Sec.~\ref{app:regime}). The same InterHand-fit per-bone codebook ($K{=}64$), applied to a regime it never saw, gives $6.23^\circ$ on HanCo against HL-26's $14.94^\circ$ ($-58\%$) and $7.90^\circ$ on lifted ASL against $14.59^\circ$ ($-46\%$); with the FreiHAND transfer of Table~\ref{tab:crossfull} ($-61\%$) these are the endpoints of the main paper's $-46$ to $-61\%$ range. (The nearby ASL figures elsewhere are different protocols, not disagreements: $8.14^\circ$ is the robustness study's operating point in Table~\ref{tab:robustfull}, $7.97^\circ$ the refitting study's, both in Sec.~\ref{app:regime}.) The direction distribution is anisotropic in every regime ($+Y$ occupies $57$--$70\%$), which is why one fitted alphabet transfers at all.

\section{Cross-Regime Generalization and Per-Finger Robustness}
\label{app:regime}

\paragraph{Additional regimes.}
We use two openly available datasets that provide 3D joints directly (no MANO fitting), avoiding access and fitting confounds: HanCo \citep{zimmermann2021hanco} ($1{,}455$ free single-hand sequences, $102{,}359$ frames) and ASL-Skeleton3D \citep{amorim2022aslskeleton3d} ($9{,}398$ American Sign Language samples, $28{,}358$ frames, with handshape/gloss/signer labels). ASL-Skeleton3D is a 2D$\to$3D lifting and is therefore noisy; its absolute errors are inflated and we read its numbers as relative comparisons. Each dataset is mapped to a common $21$-joint topology by an identity map.

\paragraph{Distribution and within-regime R--D.}
Across all four regimes the direction distribution is anisotropic ($+Y$ occupies $57$--$70\%$) and HL-26's usage entropy is $2.9$--$3.6$ bits, far below the $\log_2 26 = 4.70$ ceiling, so a fixed code wastes symbols everywhere. Data-adaptive codes win within every regime; on HanCo, k-means-$64$ reaches $4.22^\circ$ versus HL-26's $14.91^\circ$. Temporal-relative coding again lowers both rate and error on HanCo, reproducing the InterHand result.

\paragraph{Transfer.}
An InterHand-fit DigitCode-A codebook, applied to a different regime, beats HL-26 by $-46$ to $-61\%$, confirming the gains are not InterHand-specific.

\paragraph{Refitting vs.\ a frozen parametric prior.}
The main paper's adaptability claim is measured on the lifted ASL lexicon: a DigitCode-A codebook retrained on target skeletons (${\sim}1.6$\,s) reconstructs at $6.75^\circ$, and even transferred without retraining reaches $7.97^\circ$; a frozen MANO-PCA prior sits at $19$--$25^\circ$ depending on the number of retained components, full axis-angle MANO at $9.45^\circ$, and the fixed HL-26 alphabet at $14.6^\circ$. The comparison is about adaptability, not intrinsic accuracy: a codebook refits from any domain's skeleton data in seconds, while a parametric prior would need 3D scans of the target population to re-fit.

\paragraph{Discrimination under noise (reversal).}
On noisy, cross-domain ASL handshape classification the ranking inverts: the coarse HL-26 symbol is the most robust ($31.4\%$ top-1), above data-adaptive codes ($\sim$$24\%$) and the per-finger--based codes (the bootstrap pair tests the per-finger$+$whole-hand combination, $20.0\%$; per-finger alone is $19.3\%$ under the full-set protocol). A coarse alphabet quantizes input noise away, which helps a discriminator even though it hurts reconstruction. Gloss is not recoverable from a single-hand static handshape bag-of-tokens (near majority-class), as expected for a sequence-level label.

\paragraph{Diagnosing per-finger collapse.}
Because DigitCode-F packs a finger's four bones into one $12$-D code, a single corrupted bone can hijack the shared assignment: finger error correlates with the single worst bone at $0.90$--$0.92$, and the fraction of ``hijacked'' fingers rises $7\times$ under noise ($11.0\%$ on ASL vs.\ $1.6\%$ clean). Table~\ref{tab:robustfull} compares two fixes. A robust-assignment variant (I2; choose the centroid by the best three bones) \emph{fails}---it cannot help an off-manifold bone and slightly hurts. Adding an independent per-bone residual on top of the coarse per-finger code (I1, i.e.\ DigitCode-H) decouples the corrupted bone and wins within-ASL and on clean InterHand, beating per-bone DigitCode-A there; under pure IH$\to$ASL transfer DigitCode-A remains marginally better ($8.14$ vs.\ $8.95$).

\begin{table}[t]
\centering
\small
\setlength{\tabcolsep}{4pt}
\begin{tabular}{lccc}
\toprule
Method & ASL within & IH\,$\to$\,ASL & InterHand \\
\midrule
DigitCode-F (per-finger) & $14.03$ & $24.68$ & $7.10$ \\
\;\;I2 robust-assign (fails) & $16.48$ & $26.89$ & $7.65$ \\
\;\;I1 = DigitCode-H        & $\mathbf{4.29}$ & $8.95$ & $\mathbf{2.17}$ \\
DigitCode-A (per-bone)     & $6.81$ & $\mathbf{8.14}$ & $5.68$ \\
\bottomrule
\end{tabular}
\caption{Per-finger robustness diagnosis and fixes (angular error, $^\circ$). \emph{Operating point of this study}, which is coarser than the headline one: per-finger $K{=}64$, hierarchical $K_1{=}64$/$K_2{=}32$ ($5.81$ entropy b), per-bone $K{=}64$. The headline DigitCode-H ($K_1{=}128$/$K_2{=}32$, $6.12$ b) reaches $1.86^\circ$ on InterHand (Table~\ref{tab:hier}); the $2.17^\circ$ here is the same code at half the coarse codebook, not a disagreement. All arms in this table share the coarser point, so the comparison is internally matched. Robust assignment (I2) does not help; an independent per-bone residual (I1, DigitCode-H) repairs the collapse and beats per-bone DigitCode-A within-ASL and on clean InterHand (A is marginally better under pure transfer).}
\label{tab:robustfull}
\end{table}

\section{The Symbolic Repair Interface: Protocol and Full Results}
\label{app:interface}

\paragraph{Protocol.}
Held-out InterHand2.6M; DigitCode-F with $K{=}128$. All conclusions are re-verified at $K\in\{64,128,256\}$ within $\pm0.5^\circ$, with detection AUC rising monotonically ($0.945/0.953/0.960$ off-manifold, $0.808/0.823/0.827$ illegal). We use two corruption regimes: \emph{off-manifold} (each bone of a finger pushed $\sigma{=}45^\circ$; $400$ poses $\times$ $5$ fingers $=2000$ slots) and \emph{anatomically illegal} (PIP/DIP/TIP reflected about the MCP axis into hyperextension, applied only to curled fingers with MCP--TIP angle $>45^\circ$; $n{=}841$). Metrics are corrupted-finger and innocent-finger angular error; plausibility $=$ 1-NN angular distance of the finger's $12$-D direction vector to a $4{,}000$-pose pool of real training fingers (codebook-independent; ground-truth fingers score $8.3^\circ$); and OOL $=$ fraction of fingers whose plausibility exceeds the $99$th percentile of clean fingers ($27.1^\circ$). OOL scores what a repair \emph{emits}, so the unrepaired row carries no entry in either repair table. Because every symbolic arm lands at $5$--$11^\circ$ plausibility across the two regimes, far inside the threshold, OOL is $0$ for all of them by a wide margin. The column's content is that only the MANO refit leaves the manifold at scale ($0.82$--$0.94$; the learned continuous inpaint grazes it at $0.01$ on the illegal regime until the codebook projection zeroes it), and that a codeword-emitting repair cannot leave it at all. Detection thresholds $\tau$ are percentiles of the clean residual distribution.

\paragraph{Detection operating points.}
Off-manifold regime: $\tau{=}90$th pct ($19.4^\circ$) gives recall $0.88$ at $10\%$ false positives; $95$th ($22.7^\circ$): $0.79/0.05$; $99$th ($36.3^\circ$): $0.33/0.01$. On the harder illegal regime the $\tau{=}90$ gate recalls $0.50$; see the repair paragraph below. For the full detector comparison, AUC pairs are reported as off-manifold / illegal: codeword residual $0.953$ / $\mathbf{0.823}$; trained Mahalanobis $\mathbf{0.961}$ / $0.684$; intra-finger incoherence $0.737$ / $0.615$; MANO-refit residual $0.428$ / $0.534$. Read down the pairs by detector rather than across a single regime: the trained detector buys $0.008$ off-manifold and gives back $0.139$ on illegal poses, while the MANO-refit residual is below chance off-manifold and barely above it on illegal poses ($0.428$, $0.534$)---refitting a parameterization that has no notion of a legal set cannot measure distance to one. Only the codeword residual is strong in both columns, and it is the only one that needs no training.

\paragraph{Repair in the illegal regime.}
Table~\ref{tab:repairillegal} complements the main paper's off-manifold table. Snap restores validity but not correctness ($78.9^\circ$, no better than the unrepaired $77.7^\circ$, yet plausibility improves $16.0\to10.0^\circ$): the nearest legal codeword of a reflected finger is a legal but wrong pose. The whole-hand VQ row bypasses the detection gate (it rewrites every finger), so its $29.9^\circ$ is not gate-limited, while the gated arms carry the $\tau{=}90$ recall of $0.50$ on this regime (an undetected finger stays corrupted); on the detected subset the gated arms reach ${\approx}30^\circ$ (infill) and ${\approx}14^\circ$ (continuous inpaint) at $3$--$4\times$ less off-target damage than whole-VQ.

\begin{table}[t]
\centering
\small
\setlength{\tabcolsep}{3pt}
\resizebox{\columnwidth}{!}{%
\begin{tabular}{lcccc}
\toprule
Repair & corr.$^\circ\downarrow$ & innoc.$^\circ\downarrow$ & plaus.$^\circ\downarrow$ & OOL$\downarrow$ \\
\midrule
none                      & $77.7$ & $0$    & $16.0$ & --- \\
snap (nearest code)       & $78.9$ & $2.9$  & $10.0$ & $0.00$ \\
token-infill              & $54.1$ & $5.8$  & $9.1$  & $0.00$ \\
whole-hand VQ (ungated)   & $29.9$ & $20.6$ & $6.2$  & $0.00$ \\
MANO g/l/p & $103$/$104$/$95$ & ${\sim}25$ & $32$--$35$ & $0.82$--$0.88$ \\
cont.\ inpaint (learned)  & $45.9$ & $4.2$ & $11.0$ & $0.01$ \\
\;\;+ codebook projection & $46.1$ & $4.4$  & $10.6$ & $0.00$ \\
\bottomrule
\end{tabular}}
\caption{Repair arms on anatomically illegal corruptions (hyperextension; $\tau{=}90$th pct, $K{=}128$, $n{=}841$; gated arms at recall $0.50$). Snap restores validity but not correctness; the whole-hand VQ shows the cross-finger context signal but is not addressable; MANO arms fail.}
\label{tab:repairillegal}
\end{table}

\paragraph{Threshold sweep (off-manifold).}
Beyond the main table's $\tau{=}90$: at $\tau{=}95$, infill $23.0^\circ$, continuous inpaint $13.7^\circ$, hybrid $13.8^\circ$ (plausibility $8.5^\circ$, OOL $0$); at $\tau{=}99$ (recall $0.33$), $28.8/24.2/24.3^\circ$: once the gate starves, recall dominates every arm. In the illegal regime $\tau{=}99$ leaves nearly everything unrepaired (recall $0.03$), because illegal residuals overlap the clean upper quantiles; $\tau$ should be set by the application's false-positive tolerance, and the $90$th percentile is the sweet spot.

\paragraph{Confidence calibration.}
Binned calibration of the residual on held-out isotropic corruptions ($\sigma\sim U(0,60^\circ)$, disjoint fit/test halves): Spearman(residual, error) $0.836$, expected-error MAE $6.6^\circ$, ECE for $P(\mathrm{err}>15^\circ)$ of $0.017$; thresholds $\tau=15/20/25/30^\circ$ give precision $0.81/0.90/0.95/0.97$ at recall $0.95/0.80/0.62/0.45$. Transferred unchanged to the illegal regime the calibration fails (bias $-54.5^\circ$, Spearman $0.14$): a reflected finger can sit near a legal codeword, so structural illegality is \emph{detectable} (AUC $0.823$) but not \emph{quantifiable} from residual magnitude.

\paragraph{Nameable codes.}
Codes carry zero-training names: semantic naming by majority ASL handshape (signer-disjoint: $166/256$ codes named, purity $0.32$, coverage $0.97$) makes name$\to$code retrieval reach $26.8\%$ vs.\ a $10.1\%$ majority baseline ($44$ classes) by pure table lookup (e.g.\ code $246$ [all-curl] is the ASL ``10''/fist, $232$ [all-ext] is ``5''). The codebook must share the named data's canonical palm frame---fitting in the world frame drops retrieval to $18.4\%$.

\paragraph{Real-estimator study.}
MediaPipe (Tasks API) on the last $2{,}500$ original FreiHAND frames ($2{,}333$ detected); ground truth and estimate are both canonicalized before bone directions are extracted; codebooks and repair models are fit on the remaining ground-truth poses. The noise profile is diffuse and partly systematic: per-finger error mean $28.6^\circ$, median $23.3^\circ$, $90$th percentile $53.4^\circ$, $P(\mathrm{err}>15^\circ)=0.80$. Detection transfers: AUC $0.819$ for the $\mathrm{err}>15^\circ$ label ($0.785$ for $>25^\circ$), precision $0.99$ at recall $0.42$ for $\tau=15^\circ$. Repair reverses: with a $\tau{=}90$ gate flagging $0.93$ of fingers, snap improves $29.0\to21.5^\circ$ (plausibility $2.3^\circ$, OOL $0$) while infill ($32.6^\circ$), continuous inpaint ($36.5^\circ$), and the hybrid ($34.7^\circ$) all fall below no-repair: context conditioning fails when the context itself is noisy and the train (ground-truth) vs.\ test (estimator) distributions shift. This is the basis of the main paper's deployment rule.

\paragraph{Hands from generative video.}
To test the loop on generative content itself, we render $30$ two-second clips with a public text-to-video model (CogVideoX-2b; $10$ hand-centric prompts $\times$ $3$ seeds), sample $270$ frames, and lift hands with the same MediaPipe pipeline ($207$ hands detected). The per-finger codeword residual ($K{=}128$; $\tau$ at the clean InterHand train $95$th percentile) flags $92.3\%$ of generated hands on at least one finger (per-finger rates $0.42$--$0.71$; residual mean $12.1^\circ$), and snapping flagged fingers to the codebook returns them to the real-pose manifold: nearest-neighbor plausibility to a held-out real-pose pool improves from $10.8^\circ$ to $3.0^\circ$ on flagged fingers ($8.2^\circ$ to $4.1^\circ$ over all), with no ground truth at any step. A real-image control (the $2{,}500$ FreiHAND frames above, $2{,}310$ retained after canonicalization) run identically flags $97.1\%$ (plausibility $12.4^\circ\to3.0^\circ$). Two caveats keep this honest: the control does \emph{not} separate generated from real content---both arms sit in the diffuse MediaPipe-noise regime where the clean-data $\tau$ saturates, so the loop is certified consistent across content sources, not specific to generative artifacts---and the plausibility gain under snap is partly by construction. The informative quantities are the flag pattern, the end-to-end operation on generative content, and the absence of any ground-truth requirement.

\begin{table}[t]
\centering
\small
\setlength{\tabcolsep}{4pt}
\begin{tabular}{lcc}
\toprule
 & gen.\ (CogVideoX) & real (FreiHAND) \\
\midrule
hands                          & $207$   & $2{,}310$ \\
flag rate (any finger)         & $0.923$ & $0.971$ \\
residual mean ($^\circ$)       & $12.1$  & $14.6$ \\
plaus.\ before ($^\circ$) & $10.8$ & $12.4$ \\
plaus.\ after snap ($^\circ$) & $3.0$ & $3.0$ \\
\bottomrule
\end{tabular}
\caption{Detect-and-repair on hands lifted from generative video vs.\ a real-image control, identical pipeline ($\tau$ from clean InterHand train; plausibility $=$ NN angular distance to a held-out real-pose pool). Both arms sit in the diffuse estimator-noise regime; see the honesty notes in the text.}
\label{tab:genimg}
\end{table}

\paragraph{Streaming extension.}
Median-filtering the residual over $\pm2$ frames keeps recall $1.00$ while cutting false positives $-31\%$; a temporal infill matches the per-frame repairer's error ($13.8$ vs.\ $13.4^\circ$) but removes its code-switching jitter---the temporal contribution is smoothness, not accuracy.

\section{Embodied Retargeting and Label Screening: Protocol and Per-Hand Tables}
\label{app:embodied}

\paragraph{Protocol.}
Retargeting uses a standard position optimizer \citep{qin2023anyteleop} over six right robot hands spanning the kinematic design space: Allegro and Leap (independent finger chains), Shadow and Schunk SVH (fingers sharing wrist joints), and Ability and Inspire (underactuated, mimic-coupled). Lookup arms and their per-frame reference are compared under the same frozen $6$-DoF base (\emph{goldFB}: frozen-base per-frame optimization), so the comparison is paired and base-matched; a per-hand scale is calibrated once on $30$ reference frames in the same frozen-base regime. Human data are canonical-frame FreiHAND poses; codebooks (DigitCode-F, $K{=}128$ per finger; whole-hand codes at $256$/$640$/$4096$ entries for budget-matched contrasts) are fit on the training partition only. Evaluation uses $n{=}800$ held-out frames, $3$ codebook seeds (std $\le 0.11$\,mm), and frame-level paired bootstrap $95\%$ CIs. Task error is the mean fingertip position error (mm) of the executed robot pose against the retargeting targets.

\paragraph{Compiled lookup across six hands.}
Compiling the per-finger codebook costs $640$ IK solves ($5$ fingers $\times$ $128$ codes, $1.6$\,s, once); streaming retargeting is then $O(1)$ table assembly at ${\sim}0.005$\,ms/frame against $2.4$--$7.5$\,ms/frame for per-frame optimization, a $480$--$1500\times$ amortized speedup that holds on every hand. A per-bone code cannot be compiled at all: bone codes do not compose a finger-level IK target, so the decoded stream still requires per-frame optimization ($2.5$\,ms/frame on Allegro). Table~\ref{tab:robotsweep} reports accuracy. On Allegro the per-finger lookup matches frozen-base per-frame optimization within $0.4$\,mm in this sweep (Table~\ref{tab:robotsweep}; $0.7$\,mm under an earlier calibration); across the other hands a whole-hand table at matched or larger budget is the more accurate compilation unit, consistent with the paper's cross-finger-correlation evidence (real hand poses concentrate on a manifold that a few hundred whole-hand prototypes cover well). This is the granularity-to-task mapping on the robot: the \emph{finger} is the unit that gives addressability (editing one finger re-looks-up only that finger's joints) and the screening interface below, the \emph{whole hand} is the more accurate compilation unit, and the \emph{bone} compiles to nothing.

\begin{table}[t]
\centering
\small
\setlength{\tabcolsep}{4pt}
\begin{tabular}{lcccc}
\toprule
Hand & goldFB & F-lookup & W@$640$ & W@$4096$ \\
\midrule
Allegro  & $35.1$ & $35.5$ & $36.8$ & $35.6$ \\
Shadow   & $27.7$ & $30.3$ & $29.6$ & $29.1$ \\
Leap     & $40.5$ & $45.5$ & $42.1$ & $41.2$ \\
Schunk SVH & $36.9$ & $39.4$ & $37.5$ & $37.1$ \\
Ability  & $49.4$ & $51.3$ & $50.3$ & $49.9$ \\
Inspire  & $39.9$ & $42.1$ & $41.1$ & $40.6$ \\
\bottomrule
\end{tabular}
\caption{Compiled retargeting across six robot hands (task error, mm; $n{=}800$, F-lookup over $3$ codebook seeds, std $\le 0.11$\,mm, paired bootstrap CIs exclude zero). goldFB is frozen-base per-frame optimization; F-lookup the compiled per-finger table; W@$m$ a compiled whole-hand table with $m$ entries.}
\label{tab:robotsweep}
\end{table}

\paragraph{Legality screening: illegal $\Leftrightarrow$ unreachable.}
On FreiHAND evaluation poses with one finger reflected into anatomically illegal hyperextension ($n{=}600$, ground-truth bone lengths, direction-only corruption), the corrupted stream's forward-kinematics residual against its own requested targets rises above the clean floor on all four hands tested with CIs (Allegro/Shadow/Leap/Ability): $+4.6$--$6.1$\,mm, every paired CI excluding zero. Illegal symbols request targets outside the reachable set, so the closed codebook screens labels \emph{before} IK. The zero-training residual detector reaches AUC $0.975$ (recall $0.932$ / precision $0.900$ at $\tau_{90}$) and localizes the bad finger with AUC $0.988$ (clean max residual $5.9^\circ$ vs.\ corrupted $23.7^\circ$). Table~\ref{tab:robotrepair} reports the repair arms on the two fully tabulated hands; the regime rule replicates on all four CI hands: infill restores reach to the clean floor and lowers label error ($\Delta q$ $-25$/$-27\%$, $4/4$ CIs exclude zero), snap restores reachability but not pose (its label-$\Delta q$ CI crosses zero on $3/4$ hands, the same diffuse-vs-structural boundary as Sec.~\ref{app:interface}), and detect-and-drop on a mixed clean/corrupt stream collapses label error ($\Delta q$ $0.033\to0.002$ on Allegro, $0.024\to0.003$ on Shadow) at the cost of discarding $52\%$ of frames---repair vs.\ drop is a data-scarcity decision.

\begin{table}[t]
\centering
\small
\setlength{\tabcolsep}{4pt}
\begin{tabular}{lccc}
\toprule
Arm ($n{=}600$) & reach (Alg/Shd) & label $\Delta q$ & task mm \\
\midrule
clean ($=$ oracle) & $33.2$ / $24.4$ & $0$ & $33.2$ / $24.4$ \\
corrupt            & $38.7$ / $29.0$ & $.067$ / $.049$ & $36.7$ / $26.7$ \\
snap               & $36.5$ / $27.0$ & $.068$ / $.051$ & $36.6$ / $26.6$ \\
infill             & $34.1$ / $24.7$ & $.050$ / $.036$ & $35.0$ / $25.5$ \\
\bottomrule
\end{tabular}
\caption{Illegal-symbol screening and repair on Allegro (Alg) and Shadow (Shd). Columns: FK reach residual (mm), label joint error $\Delta q$, and label task error. The reach gap and the infill gain replicate with CIs on Leap and Ability.}
\label{tab:robotrepair}
\end{table}

\paragraph{Real-estimator labels and curation.}
With MediaPipe estimates from real FreiHAND images ($2{,}333/2{,}500$ detected) retargeted into action labels, snap denoising (detected fingers snapped to the nearest legal code, estimated bone lengths kept) improves label task error $13$--$15\%$ (Allegro $59.1\to51.6$\,mm, Shadow $42.7\to36.1$\,mm; significant on all $4$ CI hands, $-5.3$ to $-7.5$\,mm), matching the diffuse-noise regime rule. The same residual is a monotone curation key: keeping the cleanest $25/50/75/100\%$ of frames gives label-$\Delta q$ $0.101/0.106/0.111/0.127$ on Allegro and $0.060/0.063/0.066/0.083$ on Shadow (Spearman $0.42$/$0.50$), a ground-truth-free data-quality ranking for learning-from-video pipelines.

\paragraph{Compositional out-of-distribution encoding.}
Fit only on FreiHAND and evaluated on ASL handshapes (novel finger combinations), the per-finger encoding degrades less than whole-hand codebooks at equal or $6.4\times$ larger budget (Table~\ref{tab:robotood}): the factored code covers unseen combinations that whole-hand prototypes must have seen. This is a representation-layer property (encoding residual), reported as such; in-distribution the whole-hand code remains the more accurate fit, the same trade-off as Table~\ref{tab:robotsweep}. Two independent implementations with independent sampling agree in direction and magnitude.

\begin{table}[t]
\centering
\setlength{\tabcolsep}{3pt}
\small
\begin{tabular}{lccc}
\toprule
Codebook & in-dist.\ ($^\circ$) & OOD ASL ($^\circ$) & rise \\
\midrule
DigitCode-F ($5{\times}128$) & $3.98$ & $31.2$ & $+27.2$ \\
whole-hand @$640$            & $6.1$  & $36.8$ & $+30.7$ \\
whole-hand @$4096$           & $1.75$ & $34.9$ & $+33.2$ \\
\bottomrule
\end{tabular}
\caption{Encoding residual in-distribution (FreiHAND) vs.\ out-of-distribution (ASL handshapes), codebooks fit on FreiHAND only. The whole-hand code fits the seen distribution best but degrades most on novel finger combinations; the factored per-finger code degrades least and ends lowest. All codes rise steeply on OOD data: the claim is slower degradation from factorization, not absolute robustness.}
\label{tab:robotood}
\end{table}

\section{Consolidated Negative and Boundary Results}
\label{app:neg}

Table~\ref{tab:neg} collects results that did \emph{not} win, with the mechanism in each case. We include them because they delineate the design space and support the central finding that the best representation is task-dependent: several methods that lose on reconstruction encode a useful inductive bias for a specific downstream task, and several that win on compression are poor for generation.

\begin{table}[t]
\centering
\small
\begin{tabular}{>{\raggedright\arraybackslash}p{0.40\linewidth}>{\raggedright\arraybackslash}p{0.47\linewidth}}
\toprule
Result & Mechanism \\
\midrule
Uniform spherical $<$ HL-26 & pole on sparse $+Z$ axis; geometric densification is not the win \\
movMF / weighted k-means & $\kappa$ estimate collapses / only reshapes the tail \\
FSQ / PQ / BSQ dominated & uniform grid / independent axes / uniform octants do not fit the $+Y$ concentration \\
Residual depth at the bone unit & at every rate $4.3$--$10.9$ b a flat layer wins: splitting one bone's budget across stages costs more than staging recovers \\
Synergy PCA ${\sim}10^\circ$ floor & discarded variance is unrecoverable; only suits very low rate \\
Uniform joint-angle bins & a good parameterization $\neq$ a good quantization grid \\
Hierarchy not slow--fast & coarse layer is not slow enough ($47\%$ vs.\ $48\%$ change) \\
Per-finger not tempo-robust & highest DTW cost under tempo warps \\
Per-finger / hierarchical hard to forecast (AR) & high-entropy coarse tokens move too much per frame \\
\bottomrule
\end{tabular}
\caption{Negative and boundary results with mechanisms. Each excludes a trivial explanation or marks a limit of the design space.}
\label{tab:neg}
\end{table}

\end{document}

%% file: Figures/fig_ic_cube.tex
\providecommand{\iccube}{%
  \begin{scope}[line join=round]
    \coordinate (FBL) at (-0.44,-0.40); \coordinate (FBR) at (0.18,-0.40);
    \coordinate (FTR) at (0.18,0.22);   \coordinate (FTL) at (-0.44,0.22);
    \coordinate (BBL) at (-0.18,-0.23);  \coordinate (BBR) at (0.44,-0.23);
    \coordinate (BTR) at (0.44,0.39);    \coordinate (BTL) at (-0.18,0.39);
    \fill[cbone!7]  (FTL)--(FTR)--(BTR)--(BTL)--cycle;   % top
    \fill[cbone!12] (FBR)--(FTR)--(BTR)--(BBR)--cycle;   % right
    \fill[cbone!4]  (FBL)--(FBR)--(FTR)--(FTL)--cycle;   % front
    \draw[cbone!45!black!55,line width=0.5pt] (FBL)--(FBR)--(FTR)--(FTL)--cycle;
    \draw[cbone!45!black!55,line width=0.5pt] (FTR)--(BTR) (FTL)--(BTL) (FBR)--(BBR);
    \draw[cbone!45!black!55,line width=0.5pt] (BTL)--(BTR)--(BBR);
    \foreach \p in {(-0.44,-0.40),(0.18,-0.40),(0.18,0.22),(-0.44,0.22),
                    (0.44,0.39),(-0.18,0.39),(0.44,-0.23),
                    (-0.13,-0.09),(0.0,0.305),(0.31,-0.01),
                    (-0.13,-0.40),(0.18,-0.09),(-0.13,0.22),(-0.44,-0.09),
                    (0.13,0.39),(0.44,0.08),(0.31,0.305)}{
      \fill[cbone!60] \p circle (0.85pt);}
    \coordinate (O) at (0,-0.01);
    \foreach \p in {(0.0,0.305),(0.31,-0.01),(0.18,0.22)}{
      \draw[-{Stealth[length=3.2pt]},cbone!85,line width=0.8pt] (O) -- \p;}
    \fill[black!40] (O) circle (0.9pt);
  \end{scope}
}

%% file: Figures/fig_ic_sphere.tex
\providecommand{\icsphere}{%
  \begin{scope}
    \draw[cbone!45!black!55,line width=0.6pt] (0,0) circle (0.44);
    \draw[cbone!30,line width=0.45pt] (-0.44,0) arc (180:360:0.44 and 0.14);
    \draw[cbone!30,line width=0.45pt,densely dashed] (0.44,0) arc (0:180:0.44 and 0.14);
    \draw[cbone!22,line width=0.45pt] (0,0.44) arc (90:270:0.15 and 0.44);
    \foreach \p in {(-0.05,0.31),(0.05,0.33),(0.00,0.25),(0.11,0.27),(-0.11,0.25),
                    (0.03,0.37),(-0.03,0.20),(0.20,0.13),(0.27,0.07),(0.18,0.04),
                    (0.29,0.17),(-0.24,0.07),(-0.19,-0.03),(0.06,-0.19),(-0.08,-0.14)}{
      \fill[cbone!55] \p circle (1.0pt);}
    \foreach \p in {(0.01,0.29),(0.23,0.10),(-0.21,0.03)}{
      \draw[cbone!85!black,line width=0.9pt,fill=cbone!20] \p circle (2.1pt);}
  \end{scope}
}

%% file: Figures/fig_ic_finger.tex
\providecommand{\fingerpath}{(-0.093,-0.34)
    .. controls (-0.106,-0.07) and (-0.086,0.15) .. (-0.066,0.27)
    .. controls (-0.060,0.35) and (0.060,0.35) .. (0.066,0.27)
    .. controls (0.086,0.15) and (0.106,-0.07) .. (0.093,-0.34)
    .. controls (0.058,-0.40) and (-0.058,-0.40) .. cycle}
\providecommand{\fingershape}[1]{%
  \fill[#1] \fingerpath;
  \draw[cfing!65!black,line width=0.6pt] \fingerpath;
  \foreach \y in {-0.15,0.03,0.19}{                       % three creases -> four bones
    \draw[cfing!65!black,line width=0.4pt] (-0.070,{\y+0.02}) .. controls (0,\y) .. (0.070,{\y+0.02});}
}

\providecommand{\icfingerF}{%
  \begin{scope}[shift={(-0.30,0)}]
    \fingershape{cfing!42}
    \coordinate (mrg) at (0.40,-0.02);
    \foreach \y in {-0.245,-0.06,0.115,0.275}{
      \fill[cfing!72!black] (0.085,\y) circle (0.9pt);
      \draw[cfing!72!black,line width=0.5pt] (0.085,\y) -- (mrg);}
    \fill[cfing!72!black] (mrg) circle (1.2pt);
    \draw[-{Stealth[length=3.6pt]},cfing!75!black,line width=0.9pt] (0.43,-0.02) -- (0.57,-0.02);
    \node[rounded corners=2.5pt,fill=cfing!30,draw=cfing!65!black,line width=0.8pt,
          minimum size=0.32cm,inner sep=0pt] at (0.76,-0.02){};
  \end{scope}
}

\providecommand{\icfingerH}{%
  \begin{scope}[shift={(-0.30,0)}]
    \fingershape{cfing!40}
    \coordinate (O)  at (0.17,-0.10);
    \coordinate (C)  at (0.45,-0.01);   % coarse reconstruction
    \coordinate (Tt) at (0.53,0.19);    % true direction
    \draw[densely dotted,black!40,line width=0.5pt] (O) -- (Tt);
    \draw[-{Stealth[length=4pt]},cfing!70!black,line width=1.2pt] (O) -- (C);   % coarse
    \draw[-{Stealth[length=3.2pt]},cbone,line width=1.1pt] (C) -- (Tt);         % + residual
    \draw[cbone,fill=white,line width=0.8pt] (Tt) circle (1.5pt);
    \fill[cfing!70!black] (O) circle (1.1pt);
  \end{scope}
}

%% file: Figures/fig_overview.tex
% ============================================================
%  Figure 1 — DigitCode overview (teaser). Method-forward.
%  Icons defined in fig_ic_{cube,sphere,finger}.tex, \input in the preamble
%  (main.tex / _preview_overview.tex). This file defines colours and lays
%  out the pipeline. Ground truth for appearance = main.pdf page 3.
% ============================================================

\definecolor{cbone}{RGB}{37,99,235}    % bone   = blue
\definecolor{cfing}{RGB}{13,148,136}   % finger = teal
\definecolor{chand}{RGB}{234,88,12}    % hand   = orange
\definecolor{cboth}{RGB}{22,129,174}   % finger $+$ bone: DigitCode-H carries both

\begin{tikzpicture}[
  font=\small, >=Stealth,
  card/.style={rounded corners=6pt, line width=0.7pt, minimum width=2.4cm,
               minimum height=2.45cm, anchor=center},
  cname/.style={font=\footnotesize\bfseries},
  cdesc/.style={font=\scriptsize, text=black!55, align=center},
  tag/.style ={font=\scriptsize\bfseries, anchor=north east},
  verb/.style ={font=\footnotesize\itshape, text=black!60},
  flow/.style ={-{Stealth[length=6pt,width=5pt]}, line width=1.3pt, black!35},
]

\def\cx{3.5}\def\hw{1.20}
\def\cyi{0.28}        % icon centre
\def\cyname{-0.66}    % name
\def\cydesc{-0.96}    % descriptor
\def\tagx{1.06}\def\tagy{1.08}   % corner metric (north-east anchored)

% ===================== cards =====================
\node[card, fill=black!3, draw=black!25]       (c0) at (0*\cx,0){};
\node[card, fill=cbone!6, draw=cbone!45]       (c1) at (1*\cx,0){};
\node[card, fill=cfing!8, draw=cfing!55!black] (c2) at (2*\cx,0){};
\node[card, fill=cboth!9, draw=cboth!62!black] (c3) at (3*\cx,0){};
\node[card, fill=black!2, draw=black!22]       (c4) at (4*\cx,0){};

% ---- icons ----
\begin{scope}[shift={(0*\cx,\cyi)}]\iccube\end{scope}
\begin{scope}[shift={(1*\cx,\cyi)}]\icsphere\end{scope}
\begin{scope}[shift={(2*\cx,\cyi)}]\icfingerF\end{scope}
\begin{scope}[shift={(3*\cx,\cyi)}]\icfingerH\end{scope}
% ---- token grid (designed): bordered cells, one token column highlighted ----
\begin{scope}[shift={(4*\cx,\cyi)}]
  \draw[rounded corners=3pt, draw=black!18, line width=0.6pt, fill=black!1]
        (-0.60,-0.37) rectangle (0.60,0.41);
  \foreach \j in {0,...,4}{\foreach \i in {0,...,7}{
    \pgfmathtruncatemacro\chg{ifthenelse(\i>(1+\j),1,0)}
    \pgfmathtruncatemacro\sh{18+\j*9+26*\chg}
    \draw[fill=cfing!\sh, draw=white, line width=0.5pt]
      (\i*0.125-0.48,\j*0.125-0.28) rectangle ++(0.108,0.108);}}
  \draw[rounded corners=1.5pt, draw=cfing!75!black, line width=1.0pt]
        (0.005,-0.305) rectangle (0.138,0.345);          % one readable/editable token column
  \draw[<->,black!40,line width=0.5pt] (-0.66,-0.28)--(-0.66,0.33);
  \node[font=\scriptsize,text=black!55,left] at (-0.66,0.02){$T$};
\end{scope}

% ---- names / descriptors / metric tags (borderless, corner) ----
\node[cname]                     at (0*\cx,\cyname){HL-26};
\node[cdesc]                     at (0*\cx,\cydesc){per bone $\cdot$ fixed 26};
\node[tag,text=black!55]         at (0*\cx+\tagx,\tagy){$14.71^\circ$};

\node[cname,text=cbone!75!black] at (1*\cx,\cyname){DigitCode-A};
\node[cdesc]                     at (1*\cx,\cydesc){per bone $\cdot$ adaptive};
\node[tag,text=cbone!85!black]   at (1*\cx+\tagx,\tagy){$8.45^\circ$};

\node[cname,text=cfing!55!black] at (2*\cx,\cyname){DigitCode-F};
\node[cdesc]                     at (2*\cx,\cydesc){per finger $\cdot$ 1 token};
\node[tag,text=cfing!50!black]   at (2*\cx+\tagx,\tagy){$5.50^\circ$};

\node[cname,text=cboth!68!black] at (3*\cx,\cyname){DigitCode-H};
\node[cdesc]                     at (3*\cx,\cydesc){finger $+$ bone resid.};
\node[tag,text=cboth!62!black]   at (3*\cx+\tagx,\tagy){$3.26^\circ$};

\node[cname]  at (4*\cx,\cyname){Token grid};
\node[cdesc]  at (4*\cx,\cydesc){$T\times 40$ readable};

% ===================== flow arrows + operation verbs (above the arrow) ====
\foreach \a/\b/\v in {0/1/fit, 1/2/group, 2/3/layer, 3/4/decode}{
  \draw[flow] ($(\a*\cx+\hw,0)$) -- ($(\b*\cx-\hw,0)$);
  \node[verb] at ($(\a*\cx+0.5*\cx,0.30)$) {\v};
}

\end{tikzpicture}

%% file: digitcode_arxiv.bbl
\begin{thebibliography}{48}
\providecommand{\natexlab}[1]{#1}

\bibitem[{Cha et~al.(2024)Cha, Kim, Yoon, and Baek}]{cha2024text2hoi}
Cha, J.; Kim, J.; Yoon, J.~S.; and Baek, S. 2024.
\newblock Text2HOI: Text-Guided 3D Motion Generation for Hand-Object
  Interaction.
\newblock In \emph{IEEE/CVF Conference on Computer Vision and Pattern
  Recognition (CVPR)}.

\bibitem[{Choensawat, Nakamura, and Hachimura(2015)}]{genlaban2014}
Choensawat, W.; Nakamura, M.; and Hachimura, K. 2015.
\newblock GenLaban: A Tool for Generating Labanotation from Motion Capture
  Data.
\newblock \emph{Multimedia Tools and Applications}, 74: 10823--10846.

\bibitem[{Copet et~al.(2023)Copet, Kreuk, Gat, Remez, Kant, Synnaeve, Adi, and
  D{\'e}fossez}]{copet2023musicgen}
Copet, J.; Kreuk, F.; Gat, I.; Remez, T.; Kant, D.; Synnaeve, G.; Adi, Y.; and
  D{\'e}fossez, A. 2023.
\newblock Simple and Controllable Music Generation.
\newblock In \emph{Advances in Neural Information Processing Systems
  (NeurIPS)}.

\bibitem[{de~Amorim and Zanchettin(2022)}]{amorim2022aslskeleton3d}
de~Amorim, C.~C.; and Zanchettin, C. 2022.
\newblock ASL-Skeleton3D and ASL-Phono: Two Novel Datasets for the American
  Sign Language.
\newblock \emph{arXiv preprint arXiv:2201.02065}.

\bibitem[{Douglas and Peucker(1973)}]{douglas1973rdp}
Douglas, D.~H.; and Peucker, T.~K. 1973.
\newblock Algorithms for the Reduction of the Number of Points Required to
  Represent a Digitized Line or Its Caricature.
\newblock \emph{The Canadian Cartographer}, 10(2): 112--122.

\bibitem[{Guo et~al.(2024)Guo, Mu, Javed, Wang, and Cheng}]{guo2024momask}
Guo, C.; Mu, Y.; Javed, M.~G.; Wang, S.; and Cheng, L. 2024.
\newblock MoMask: Generative Masked Modeling of 3D Human Motions.
\newblock In \emph{IEEE/CVF Conference on Computer Vision and Pattern
  Recognition (CVPR)}.

\bibitem[{Guo et~al.(2022)Guo, Zuo, Wang, and Cheng}]{guo2022tm2t}
Guo, C.; Zuo, X.; Wang, S.; and Cheng, L. 2022.
\newblock TM2T: Stochastic and Tokenized Modeling for the Reciprocal Generation
  of 3D Human Motions and Texts.
\newblock In \emph{European Conference on Computer Vision (ECCV)}.

\bibitem[{Hsiao et~al.(2021)Hsiao, Liu, Yeh, and Yang}]{hsiao2021cp}
Hsiao, W.-Y.; Liu, J.-Y.; Yeh, Y.-C.; and Yang, Y.-H. 2021.
\newblock Compound Word Transformer: Learning to Compose Full-Song Music over
  Dynamic Directed Hypergraphs.
\newblock In \emph{Proceedings of the AAAI Conference on Artificial
  Intelligence}.

\bibitem[{Huang et~al.(2025)Huang, Chen, Lin et~al.}]{huang2025hoigpt}
Huang, M.; Chen, F.-J.; Lin, Y.-H.; et~al. 2025.
\newblock HOIGPT: Learning Long-Sequence Hand-Object Interaction with Language
  Models.
\newblock In \emph{IEEE/CVF Conference on Computer Vision and Pattern
  Recognition (CVPR)}.

\bibitem[{Huang et~al.(2024)Huang, Wan, Yang, Callison-Burch, Yatskar, and
  Liu}]{huang2024como}
Huang, Y.; Wan, W.; Yang, Y.; Callison-Burch, C.; Yatskar, M.; and Liu, L.
  2024.
\newblock CoMo: Controllable Motion Generation through Language Guided Pose
  Code Editing.
\newblock In \emph{European Conference on Computer Vision (ECCV)}.

\bibitem[{Hwang et~al.(2026)Hwang, Jang, Zhou, Wang, Kim, and
  Guo}]{scalemogen2026}
Hwang, I.; Jang, H.; Zhou, B.; Wang, J.; Kim, Y.~M.; and Guo, C. 2026.
\newblock ScaleMoGen: Autoregressive Next-Scale Prediction for Human Motion
  Generation.
\newblock \emph{arXiv preprint arXiv:2605.11704}.

\bibitem[{J{\'e}gou, Douze, and Schmid(2011)}]{jegou2011pq}
J{\'e}gou, H.; Douze, M.; and Schmid, C. 2011.
\newblock Product Quantization for Nearest Neighbor Search.
\newblock \emph{IEEE Transactions on Pattern Analysis and Machine
  Intelligence}, 33(1): 117--128.

\bibitem[{Jiang et~al.(2023)Jiang, Chen, Liu, Yu, Yu, and
  Chen}]{jiang2023motiongpt}
Jiang, B.; Chen, X.; Liu, W.; Yu, J.; Yu, G.; and Chen, T. 2023.
\newblock MotionGPT: Human Motion as a Foreign Language.
\newblock In \emph{Advances in Neural Information Processing Systems
  (NeurIPS)}.

\bibitem[{Jiang et~al.(2024)Jiang, Au, Chen, Xiang, and Chen}]{hkbu_lbnmae}
Jiang, J.; Au, H.~Y.; Chen, J.; Xiang, J.; and Chen, M. 2024.
\newblock Motion Part-Level Interpolation and Manipulation over Automatic
  Symbolic Labanotation Annotation.
\newblock In \emph{International Joint Conference on Neural Networks (IJCNN)}.

\bibitem[{Jiang et~al.(2026)Jiang, Au, Xiang, and Chen}]{jiang2026lamogen}
Jiang, J.; Au, H.~Y.; Xiang, J.; and Chen, J. 2026.
\newblock LaMoGen: Language to Motion Generation Through LLM-Guided Symbolic
  Inference.
\newblock In \emph{IEEE/CVF Conference on Computer Vision and Pattern
  Recognition (CVPR)}.

\bibitem[{Li et~al.(2024)Li, Yang, Xing, Yu, and Zhang}]{li2024hl}
Li, L.; Yang, W.; Xing, J.; Yu, X.; and Zhang, X.-P. 2024.
\newblock Translating Motion to Notation: Hand Labanotation for Intuitive and
  Comprehensive Hand Movement Documentation.
\newblock In \emph{Proceedings of the 32nd ACM International Conference on
  Multimedia (ACM MM)}.

\bibitem[{Li et~al.(2026)Li, Li, Hou, Chen, Chang, Liu, and Shan}]{anymo2026}
Li, Y.; Li, Z.; Hou, R.; Chen, Y.; Chang, H.; Liu, H.; and Shan, S. 2026.
\newblock AnyMo: Scaling Any-Modality Conditional Motion Generation with Masked
  Modeling.
\newblock \emph{arXiv preprint arXiv:2605.29488}.

\bibitem[{Liu et~al.(2023)Liu, Zhou, Yang, Gupta, and Wang}]{liu2024contactgen}
Liu, S.; Zhou, Y.; Yang, J.; Gupta, S.; and Wang, S. 2023.
\newblock ContactGen: Generative Contact Modeling for Grasp Generation.
\newblock In \emph{IEEE/CVF International Conference on Computer Vision
  (ICCV)}.

\bibitem[{Lloyd(1982)}]{lloyd1982kmeans}
Lloyd, S.~P. 1982.
\newblock Least Squares Quantization in PCM.
\newblock \emph{IEEE Transactions on Information Theory}, 28(2): 129--137.

\bibitem[{Lu et~al.(2024)Lu, Xu, Zhang, Wang, and Tao}]{lu2024handrefiner}
Lu, W.; Xu, Y.; Zhang, J.; Wang, C.; and Tao, D. 2024.
\newblock HandRefiner: Refining Malformed Hands in Generated Images by
  Diffusion-Based Conditional Inpainting.
\newblock In \emph{Proceedings of the 32nd ACM International Conference on
  Multimedia (ACM MM)}.

\bibitem[{Luo et~al.(2025)Luo, Feng, Zhang, Zheng, Wang, Yuan, Liu, Xu, Jin,
  and Lu}]{luo2025beingh0}
Luo, H.; Feng, Y.; Zhang, W.; Zheng, S.; Wang, Y.; Yuan, H.; Liu, J.; Xu, C.;
  Jin, Q.; and Lu, Z. 2025.
\newblock Being-H0: Vision-Language-Action Pretraining from Large-Scale Human
  Videos.
\newblock \emph{arXiv preprint arXiv:2507.15597}.

\bibitem[{Luo, Yu, and Wang(2023)}]{daskel}
Luo, S.; Yu, B.; and Wang, Z. 2023.
\newblock DASKEL: An Interactive Choreographic System with
  Labanotation-Skeleton Translation.
\newblock In \emph{Pacific Graphics Short Papers}. The Eurographics
  Association.

\bibitem[{Mentzer et~al.(2024)Mentzer, Minnen, Agustsson, and
  Tschannen}]{mentzer2024fsq}
Mentzer, F.; Minnen, D.; Agustsson, E.; and Tschannen, M. 2024.
\newblock Finite Scalar Quantization: VQ-VAE Made Simple.
\newblock In \emph{International Conference on Learning Representations
  (ICLR)}.

\bibitem[{Moon et~al.(2020)Moon, Yu, Wen, Shiratori, and
  Lee}]{moon2020interhand}
Moon, G.; Yu, S.-I.; Wen, H.; Shiratori, T.; and Lee, K.~M. 2020.
\newblock InterHand2.6M: A Dataset and Baseline for 3D Interacting Hand Pose
  Estimation from a Single RGB Image.
\newblock In \emph{European Conference on Computer Vision (ECCV)}.

\bibitem[{Pavlakos et~al.(2024)Pavlakos, Shan, Radosavovic, Kanazawa, Fouhey,
  and Malik}]{pavlakos2024hamer}
Pavlakos, G.; Shan, D.; Radosavovic, I.; Kanazawa, A.; Fouhey, D.; and Malik,
  J. 2024.
\newblock Reconstructing Hands in 3D with Transformers.
\newblock In \emph{IEEE/CVF Conference on Computer Vision and Pattern
  Recognition (CVPR)}.

\bibitem[{Pinyoanuntapong et~al.(2024)Pinyoanuntapong, Wang, Lee, and
  Chen}]{pinyoanuntapong2024mmm}
Pinyoanuntapong, E.; Wang, P.; Lee, M.; and Chen, C. 2024.
\newblock MMM: Generative Masked Motion Model.
\newblock In \emph{IEEE/CVF Conference on Computer Vision and Pattern
  Recognition (CVPR)}.

\bibitem[{Prillwitz et~al.(1989)Prillwitz, Leven, Zienert, Hanke, and
  Henning}]{prillwitz1989hamnosys}
Prillwitz, S.; Leven, R.; Zienert, H.; Hanke, T.; and Henning, J. 1989.
\newblock \emph{{HamNoSys} Version 2.0: {Hamburg} Notation System for Sign
  Languages: An Introductory Guide}.
\newblock Hamburg: Signum Press.

\bibitem[{Qian et~al.(2026)Qian, Gu, Zhao, and Wang}]{qian2026beat}
Qian, L.; Gu, H.; Zhao, J.; and Wang, Z. 2026.
\newblock BEAT: Tokenizing and Generating Symbolic Music by Uniform Temporal
  Steps.
\newblock In \emph{International Conference on Machine Learning (ICML)}.

\bibitem[{Qin et~al.(2022)Qin, Wu, Liu, Jiang, Yang, Fu, and
  Wang}]{qin2022dexmv}
Qin, Y.; Wu, Y.-H.; Liu, S.; Jiang, H.; Yang, R.; Fu, Y.; and Wang, X. 2022.
\newblock DexMV: Imitation Learning for Dexterous Manipulation from Human
  Videos.
\newblock In \emph{European Conference on Computer Vision (ECCV)}.

\bibitem[{Qin et~al.(2023)Qin, Yang, Huang, Van~Wyk, Su, Wang, Chao, and
  Fox}]{qin2023anyteleop}
Qin, Y.; Yang, W.; Huang, B.; Van~Wyk, K.; Su, H.; Wang, X.; Chao, Y.-W.; and
  Fox, D. 2023.
\newblock AnyTeleop: A General Vision-Based Dexterous Robot Arm-Hand
  Teleoperation System.
\newblock In \emph{Robotics: Science and Systems (RSS)}.

\bibitem[{Romero, Tzionas, and Black(2017)}]{romero2017mano}
Romero, J.; Tzionas, D.; and Black, M.~J. 2017.
\newblock Embodied Hands: Modeling and Capturing Hands and Bodies Together.
\newblock \emph{ACM Transactions on Graphics (Proc. SIGGRAPH Asia)}, 36(6).

\bibitem[{Saunders, Camgoz, and Bowden(2020)}]{saunders2020progressive}
Saunders, B.; Camgoz, N.~C.; and Bowden, R. 2020.
\newblock Progressive Transformers for End-to-End Sign Language Production.
\newblock In \emph{European Conference on Computer Vision (ECCV)}.

\bibitem[{Shaw, Bahl, and Pathak(2023)}]{shaw2023videodex}
Shaw, K.; Bahl, S.; and Pathak, D. 2023.
\newblock VideoDex: Learning Dexterity from Internet Videos.
\newblock In \emph{Conference on Robot Learning (CoRL)}.

\bibitem[{Shoemake(1985)}]{shoemake1985slerp}
Shoemake, K. 1985.
\newblock Animating Rotation with Quaternion Curves.
\newblock In \emph{Proceedings of SIGGRAPH (ACM Computer Graphics)}, volume~19,
  245--254.

\bibitem[{Symeonidis-Herzig et~al.(2026)Symeonidis-Herzig, Low,
  Mercanoglu~Sincan, and Bowden}]{m3t2026}
Symeonidis-Herzig, A.; Low, J.; Mercanoglu~Sincan, O.; and Bowden, R. 2026.
\newblock M3T: Discrete Multi-Modal Motion Tokens for Sign Language Production.
\newblock \emph{arXiv preprint arXiv:2603.23617}.

\bibitem[{Touvron et~al.(2023)Touvron, Lavril, Izacard
  et~al.}]{touvron2023llama}
Touvron, H.; Lavril, T.; Izacard, G.; et~al. 2023.
\newblock LLaMA: Open and Efficient Foundation Language Models.
\newblock \emph{arXiv preprint arXiv:2302.13971}.

\bibitem[{van~den Oord, Vinyals, and Kavukcuoglu(2017)}]{vandenoord2017vqvae}
van~den Oord, A.; Vinyals, O.; and Kavukcuoglu, K. 2017.
\newblock Neural Discrete Representation Learning.
\newblock In \emph{Advances in Neural Information Processing Systems
  (NeurIPS)}.

\bibitem[{Wang et~al.(2025)}]{wang2025realishuman}
Wang, B.; et~al. 2025.
\newblock RealisHuman: A Two-Stage Approach for Refining Malformed Human Parts
  in Generated Images.
\newblock In \emph{Proceedings of the AAAI Conference on Artificial
  Intelligence}.

\bibitem[{Wang, Tan, and Yang(2025)}]{wang2025timeshifted}
Wang, T.-K.; Tan, C.-P.; and Yang, Y.-H. 2025.
\newblock Time-Shifted Token Scheduling for Symbolic Music Generation.
\newblock \emph{arXiv preprint arXiv:2509.23749}.

\bibitem[{Xie et~al.(2024)Xie, Zhang, Peng, Tang, Du, and Li}]{xie2024g2pddm}
Xie, P.; Zhang, Q.; Peng, T.; Tang, H.; Du, Y.; and Li, Z. 2024.
\newblock G2P-DDM: Generating Sign Pose Sequence from Gloss Sequence with
  Discrete Diffusion Model.
\newblock In \emph{Proceedings of the AAAI Conference on Artificial
  Intelligence}.

\bibitem[{Yuan et~al.(2024)Yuan, Yu, He, Zhang et~al.}]{yuan2024mogents}
Yuan, W.; Yu, Y.; He, J.; Zhang, Y.; et~al. 2024.
\newblock MoGenTS: Motion Generation based on Spatial-Temporal Joint Modeling.
\newblock In \emph{Advances in Neural Information Processing Systems
  (NeurIPS)}.

\bibitem[{Zeghidour et~al.(2021)Zeghidour, Luebs, Omran, Skoglund, and
  Tagliasacchi}]{zeghidour2021soundstream}
Zeghidour, N.; Luebs, A.; Omran, A.; Skoglund, J.; and Tagliasacchi, M. 2021.
\newblock SoundStream: An End-to-End Neural Audio Codec.
\newblock \emph{IEEE/ACM Transactions on Audio, Speech, and Language
  Processing}.

\bibitem[{Zhang et~al.(2026)Zhang, Li, Fan, Xu, and
  Tang}]{laban_transformer_lstm}
Zhang, H.; Li, Y.; Fan, K.; Xu, T.; and Tang, H. 2026.
\newblock Automatic Generation of Labanotation Based on a Hybrid
  Transformer--LSTM Network with Multi-Scale Spatio-Temporal Features.
\newblock \emph{Scientific Reports}, 16: 18244.

\bibitem[{Zhang et~al.(2023)Zhang, Zhang, Cun, Huang, Zhang, Zhao, Lu, and
  Shen}]{zhang2023t2mgpt}
Zhang, J.; Zhang, Y.; Cun, X.; Huang, S.; Zhang, Y.; Zhao, H.; Lu, H.; and
  Shen, X. 2023.
\newblock T2M-GPT: Generating Human Motion from Textual Descriptions with
  Discrete Representations.
\newblock In \emph{IEEE/CVF Conference on Computer Vision and Pattern
  Recognition (CVPR)}.

\bibitem[{Zhao, Xiong, and Kr{\"a}henb{\"u}hl(2024)}]{zhao2024bsq}
Zhao, Y.; Xiong, Y.; and Kr{\"a}henb{\"u}hl, P. 2024.
\newblock Image and Video Tokenization with Binary Spherical Quantization.
\newblock \emph{arXiv preprint arXiv:2406.07548}.

\bibitem[{Zimmermann, Argus, and Brox(2021)}]{zimmermann2021hanco}
Zimmermann, C.; Argus, M.; and Brox, T. 2021.
\newblock Contrastive Representation Learning for Hand Shape Estimation.
\newblock In \emph{German Conference on Pattern Recognition (GCPR)}.

\bibitem[{Zimmermann et~al.(2019)Zimmermann, Ceylan, Yang, Russell, Argus, and
  Brox}]{zimmermann2019freihand}
Zimmermann, C.; Ceylan, D.; Yang, J.; Russell, B.; Argus, M.; and Brox, T.
  2019.
\newblock FreiHAND: A Dataset for Markerless Capture of Hand Pose and Shape
  from Single RGB Images.
\newblock In \emph{IEEE/CVF International Conference on Computer Vision
  (ICCV)}.

\bibitem[{Zuo et~al.(2025)Zuo, Potamias, Ververas, Deng, and
  Zafeiriou}]{signsastokens2024}
Zuo, R.; Potamias, R.~A.; Ververas, E.; Deng, J.; and Zafeiriou, S. 2025.
\newblock Signs as Tokens: A Retrieval-Enhanced Multilingual Sign Language
  Generator.
\newblock In \emph{IEEE/CVF International Conference on Computer Vision
  (ICCV)}.

\end{thebibliography}
